\documentclass[journal,10pt]{IEEEtran}

\usepackage[utf8]{inputenc} % allow utf-8 input
\usepackage[T1]{fontenc}    % use 8-bit T1 fonts
\usepackage{hyperref}       % hyperlinks
\usepackage{url}            % simple URL typesetting
\usepackage{booktabs}       % professional-quality tables
\usepackage{amsfonts}       % blackboard math symbols
\usepackage{nicefrac}       % compact symbols for 1/2, etc.
\usepackage{microtype}      % microtypography
\usepackage{cite}
\usepackage{amsmath,amssymb,amsthm}
\usepackage{algorithm, algorithmic}
\usepackage{url}
\usepackage{pgf}
\usepackage{tikz}
\usetikzlibrary{arrows,automata}
\usepackage{subfigure}
\usepackage{multirow}
\usetikzlibrary{spy}
\usepackage{pgfplots}
\usepackage{enumitem}
\pgfplotsset{width=7cm,compat=1.8}
\makeatletter
\def\parsecomma#1,#2\endparsecomma{\def\page@x{#1}\def\page@y{#2}}
\tikzdeclarecoordinatesystem{page}{
    \parsecomma#1\endparsecomma
    \pgfpointanchor{current page}{north east}
    \pgf@xc=\pgf@x%
    \pgf@yc=\pgf@y%
    \pgfpointanchor{current page}{south west}
    \pgf@xb=\pgf@x%
    \pgf@yb=\pgf@y%
    \pgfmathparse{(\pgf@xc-\pgf@xb)/2.*\page@x+(\pgf@xc+\pgf@xb)/2.}
    \expandafter\pgf@x\expandafter=\pgfmathresult pt
    \pgfmathparse{(\pgf@yc-\pgf@yb)/2.*\page@y+(\pgf@yc+\pgf@yb)/2.}
    \expandafter\pgf@y\expandafter=\pgfmathresult pt
}
\makeatother

\newcommand\zoombox[2][]{
    \begin{scope}[zoombox paths]
        \pgfmathsetmacro\xpos{
            (\columncount-1)*(\imagewidth / \pgfkeysvalueof{/tikz/zoomboxarray columns} + \pgfkeysvalueof{/tikz/zoomboxarray inner gap} / \pgfkeysvalueof{/tikz/zoomboxarray columns} ) + \pgflinewidth
        }
        \pgfmathsetmacro\ypos{
            (\rowcount-1)*( \imageheight / \pgfkeysvalueof{/tikz/zoomboxarray rows} + \pgfkeysvalueof{/tikz/zoomboxarray inner gap} / \pgfkeysvalueof{/tikz/zoomboxarray rows} ) + 0.5*\pgflinewidth
        }
        \edef\dospy{\noexpand\spy [
            #1,
            zoombox paths/.append style={
                black and white pattern=\patternnumber
            },
            every spy on node/.append style={#1},
            x=\imagewidth,
            y=\imageheight
        ] on (#2) in node [anchor=north west] at ($(zoomboxes container.north west)+(\xpos pt,-\ypos pt)$);}
        \dospy
        \pgfmathtruncatemacro\pgfmathresult{ifthenelse(\columncount==\pgfkeysvalueof{/tikz/zoomboxarray columns},\rowcount+1,\rowcount)}
        \global\let\rowcount=\pgfmathresult
        \pgfmathtruncatemacro\pgfmathresult{ifthenelse(\columncount==\pgfkeysvalueof{/tikz/zoomboxarray columns},1,\columncount+1)}
        \global\let\columncount=\pgfmathresult
        \ifblackandwhitecycle
            \pgfmathtruncatemacro{\newpatternnumber}{\patternnumber+1}
            \global\edef\patternnumber{\newpatternnumber}
        \fi
    \end{scope}
}

\renewcommand{\qedsymbol}{$\blacksquare$}
\newcommand{\Op}[1]{\operatorname{\mathcal{#1}}}

\newcommand{\x}{\boldsymbol{x}}
\newcommand{\y}{\boldsymbol{y}}
\newcommand{\z}{\boldsymbol{z}}
\newcommand{\e}{\boldsymbol{e}}

\newcommand{\X}{\mathbb{X}}
\newcommand{\Y}{\mathbb{Y}}
\newcommand{\V}{\mathbb{V}}

\newcommand{\R}{\mathcal{R}}

\newcommand{\W}{\mathcal{W}}

\newcommand{\B}{\mathcal{B}}
\newcommand{\U}{\mathcal{U}}

\newtheorem{lemma}{Lemma}
\newtheorem{prop}{Proposition}

\DeclareMathOperator*{\argmin}{arg\,min}

\ifCLASSINFOpdf
\else
\fi
\begin{document}
\title{Learned convex regularizers for inverse problems}
\author{Subhadip Mukherjee$^1$,  S\"oren Dittmer$^2$, Zakhar Shumaylov$^1$, Sebastian Lunz$^1$, Ozan \"Oktem$^3$, and Carola-Bibiane Sch\"onlieb$^1$ % <-this % stops a space
\thanks{$1$: Department of Applied Mathematics and Theoretical Physics, University of Cambridge, UK; emails: \{sm2467, zs334, sl767, cbs31\}@cam.ac.uk. $2$: Center for Industrial Mathematics, University of Bremen, Germany; email: sdittmer@math.uni-bremen.de. $3$: Department of Mathematics, KTH -- Royal Institute of Technology, Sweden; email: ozan@kth.se.}
}

\markboth{}
{Shell \MakeLowercase{\textit{et al.}}: Bare Demo of IEEEtran.cls for Journals}
\maketitle
\begin{abstract}
We consider the variational reconstruction framework for inverse problems and propose to learn a data-adaptive input-convex neural network (ICNN) as the regularization functional. The ICNN-based convex regularizer is trained adversarially to discern ground-truth images from unregularized reconstructions. Convexity of the regularizer is desirable since (i) one can establish analytical convergence guarantees for the corresponding variational reconstruction problem and (ii) devise efficient and provable algorithms for reconstruction. In particular, we show that the optimal solution to the variational problem converges to the ground-truth if the penalty parameter decays sub-linearly with respect to the norm of the noise. Further, we prove the existence of a sub-gradient-based algorithm that leads to a monotonically decreasing error in the parameter space with iterations. To demonstrate the performance of our approach for solving inverse problems, we consider the tasks of deblurring natural images and reconstructing images in computed tomography (CT), and show that the proposed convex regularizer is at least competitive with and sometimes superior to state-of-the-art data-driven techniques for inverse problems.
\end{abstract}
\section{Introduction}
\IEEEPARstart{I}{nverse} problems arise in numerous scientific applications, e.g., in virtually every modern medical imaging modality, wherein the key objective is to estimate some parameters of interest based on an indirect and possibly noisy measurement. 

An inverse problem is said to be ill-posed if it has no or multiple solutions, or if its solution is not continuous in the measurement. 
Many traditional approaches attempt to alleviate the issue of ill-posedness by involving hand-crafted prior information on possible reconstructions. 

While such analytical priors usually lead to provable properties, they fall short in terms of data-adaptability; i.e., it is impossible to define formally, which, of all possible images, are \textit{natural images}. In recent years, several solutions to this problem emerged with the rise of deep learning methods~\cite{data_driven_inv_prob}. While the deep learning-based techniques often produce reconstructions of astonishingly high quality, they typically lack many, if not all, of the provable properties that the traditional variational approaches offer. In this work, we propose an approach that integrates deep learning into the classical regularization theory by learning a strongly convex regularizer to incorporate prior information into a variational reconstruction setting. Before explaining our contributions in more detail, we provide a brief overview of related works.
\subsection{Related works}
Fully data-driven approaches for inverse problems aim to either map the measurement directly to the model parameter \cite{automap}, or remove artifacts from an analytical reconstruction method \cite{postprocessing_cnn} by training an over-parametrized neural network using pairs of appropriate input and target images. Such end-to-end fully trained methods are data-intensive and might generalize poorly if they are trained on limited amount of examples.

Another recent data-driven approach for inverse problems consists in \textit{unrolling} of iterative model-based methods \cite{jonas_learned_iterative,lpd_tmi,kobler2017variational,meinhardt2017learning}. Iterative unrolling techniques are supervised and incorporate the forward operator in the learning model to achieve data-efficiency (thereby generalizing well from moderate amount of data), setting them apart from fully-trained \cite{automap} or post-processing-based approaches \cite{postprocessing_cnn}.

The idea of learning a sparsity-promoting data-driven regularizer has been a prominent research direction in recent years, the origin of which can be traced back to the pioneering work by Aharon et al. \cite{elad_ksvd1}. The key concept in \cite{elad_ksvd1} was to learn a sparsifying synthesis dictionary leading to a parsimonious signal representation in terms of only a few dictionary elements. Sparsity in the learned basis can be promoted by penalizing the $\ell_1$-norm of the representation coefficients, resulting in a convex variational problem for a fixed pre-trained dictionary. The linear synthesis sparsity model in \cite{elad_ksvd1} was subsequently extended to analysis- and transform-based linear sparsity models \cite{elad_ksvd1_analysis,sparse_t}. The surge of deep learning research led to a refinement of the classical sparsity models and attempts were made to draw connections between convolutional neural networks (CNNs) and multi-layer convolutional sparse modeling (see \cite{mlcsc_elad} and references therein).

One of the recent approaches for solving inverse problems is based on the idea of using a denoiser inside an algorithm for minimizing the variational objective \cite{romano2017RED,chan2016plug}. The regularization by denoising (RED) algorithm proposed in \cite{romano2017RED} belongs to this class, wherein one constructs an explicit regularizer from an image denoiser by penalizing the inner product of the image with its denoising residual. Contrary to the claim in \cite{romano2017RED}, it was shown in \cite[Sec. III.G]{red_schniter} that such a construction may fail to produce a convex regularizer even when the Jacobian matrix of the denoiser is symmetric and its eigenvalues lie in $[0,1]$. However, it is argued in \cite{red_schniter} that the RED algorithm in \cite{romano2017RED} does not necessarily minimize the variational loss arising from the RED-regularizer. Therefore, instead of relying on convexity of the regularizer, \cite{red_schniter} develops a new framework based on \textit{score-matching by denoising} (SMD) for the convergence analysis of the RED algorithm. The \textit{regularization by artifact-removal} (RARE) approach proposed in \cite{rare_deep_prior} aims to learn an artifact-removing CNN as the denoiser in the RED framework by using pairs of undersampled measurements. Although it might be possible, in principle, to learn the artifact-removing CNN in such a way that the resulting RED regularizer is convex, the constraints that one has to impose on the denoising CNN during training to ensure convexity of the regularizer are not straightforward to derive, especially in view of the analysis presented in \cite{red_schniter}. A fixed-point convergence analysis of  proximal gradient and alternating directions method of multipliers (ADMM) was developed in \cite{online_pnp_tci} by replacing the proximal operator with a plug-and-play (PnP) denoising operator of the form $\mathcal{D}=\gamma\,\mathcal{I}+(1-\gamma)\,\mathcal{D}_1$, where $\gamma\in(0,1)$, $\mathcal{I}$ denotes the identity operator, and $\mathcal{D}_1$ is non-expansive. PnP methods are equivalent to regularized image reconstruction in spirit, since the PnP denoiser corresponds to an implicit prior.

The line of research that we build on directly uses a neural network to parametrize the regularization functional, allowing to reconstruct from the observed data by solving a variational optimization problem \cite{ar_nips,nett_paper, ulyanov2018deepImagePrior,kobler2020total}. More specifically, our work relies upon the philosophy of learning an adversarial regularizer (AR) introduced in \cite{ar_nips}, which is parametrized using a neural network. Aside from \cite{ar_nips}, the idea of using a trained neural network as a regularizer was considered in \cite{nett_paper} (referred to as network Tikhonov (NETT)) and more recently in \cite{kobler2020total} (referred to as total deep variation (TDV)). The key difference between the AR, and the TDV and NETT approaches lies in the training protocol for the regularizer. While the AR is trained with an objective to discriminate desired images from noisy ones, the NETT regularizer rests upon an encoder-decoder setup and the TDV approach trains the regularizer by differentiating through the minimization of the variational problem, similar to the unrolling schemes discussed before \cite{jonas_learned_iterative}.
%\textcolor{red}{Say here somewhere that AR is unsupervised.}
In all cases (AR, NETT, and TDV), gradient-descent is used for solving the variational problem for reconstruction and no convergence guarantees that match the classical results for convex regularizers can be derived in these settings. Further, since a penalty term may be interpreted as the inclusion of prior information, the idea of incorporating the prior knowledge by restricting the reconstruction to lie in the range of a generative model proposed in~\cite{peng2019auto} is also closely related to our work. Under smoothness conditions, this approach can be shown to be equivalent to learning a penalty function via a Lagrangian argument~\cite{dittmer2019regularization}.
% \Sebastian{instead of 'no convergence guarantees' should we add here 'no convergence guarantees that match the classical results for convex regularization functionals' so people cannot argue that we ignore Thm 3.1 in NETT?} for it can be given. 

In this work, we leverage strong convexity by generalizing the parametrization of input-convex neural networks (ICNNs) proposed in~\cite{amos2017input} while designing the regularizer. This not only allows us to show well-posedness and strong convergence of the variational reconstruction but also guides the development of a convergent sub-gradient descent algorithm for reconstruction. We state the specific contributions in the following subsection. 
\subsection{Specific contributions}
This work builds upon \cite{ar_nips} that introduces the adversarial regularizer (AR) framework. The idea in AR is to replace a hand-crafted regularizer with a learned one, parametrized by a deep neural network. The parametric regularizer is first trained to separate ground-truth images from images containing artifacts. The loss function for training seeks to maximize the output of AR for noisy images, while minimizing it for the ground-truth images. When the regularizer is constrained to be 1-Lipschitz, the optimal training loss corresponds to the Wasserstein distance \cite{wgan_main} between the distributions of the ground-truth and that of the noisy images. One does not need paired images to approximate the training objective in AR, which makes the framework unsupervised in theory. Subsequently, the trained AR is deployed in a variational scheme for solving an ill-posed inverse problem. 

To the best of our knowledge, this work makes the first attempt to enforce convexity on the learned regularizer by restricting the architecture in order to combine deep learning with convex regularization theory. This helps us establish stronger convergence results (proposition \ref{prop_conv_prop_var_opt}) and derive a precise stability estimate (Proposition \ref{prop_stability_prop}) as opposed to the AR framework. Further, by exploiting convexity, we prove the existence of a convergent sub-gradient algorithm for minimizing the variational objective (Lemma \ref{subgrad_convergence_lemma}).

By utilizing ICNNs~\cite{amos2017input}, one can rigorously study regularizing properties of such variational schemes even when the regularizer is learned. 
The convergence guarantees and ICNN parametrization developed by us are applicable in the general setting where the underlying parameters and the measurement belong to Hilbert spaces. The resulting data-driven adversarial convex regularizer (ACR) offers reconstruction performance that is competitive with similar learned methods while being provably convergent simultaneously. In specific applications, we found that a convex regularizer even outperforms its non-convex variant, especially when the training dataset is small, and the forward operator is severely ill-conditioned.
% \textcolor{red}{Ozan: We should at some point address in what sense our approach is ``kernel aware'', i.e., in what sense we avoid developing develop instabilities in the sense of Theorem 3.1 in \href{https://arxiv.org/abs/2001.01258}{https://arxiv.org/abs/2001.01258}
% }
\section{Background on inverse problems}
We denote function spaces by blackboard-bold letters (e.g. $\X$) and functionals (mapping function spaces to the real line) by calligraphic letters (e.g. $\R$). Generic elements of a space are denoted by boldface lowercase letters (e.g. $\x\in\X$). Simple uppercase letters are used to denote random variables, e.g., a generic $\X$-valued random variable is denoted as $X$, with its distribution being denoted as $\pi_X$.
\subsection{Classical formulation}
Inverse problems deal with reconstructing unknown model parameter $\x^*\in \X$ from the indirect measurement
\begin{equation}
    \y^{\delta} = \Op{A}({\x}^*)+ \e \in\Y,
    \label{inv_prob_data}
\end{equation}
where $\Op{A}:\X\rightarrow \Y$ is the forward operator and $\e\in \Y$, $\left\|\e\right\|_2\leq \delta$, denotes measurement noise. Here $\X$ and $\Y$ are Hilbert spaces containing possible model parameters and data, respectively. 

In the context of medical imaging, e.g., computed tomography (CT), the model parameter ${\x}^* \in \X$ is the image of the interior structure one seeks to recover. The measurement $\y^{\delta} \in \Y$ (\textit{data}) represents indirect observations of $\x^*$. 

In the classical function-analytic formulation, ${\x}^*$ is modeled as deterministic and the standard practice is to approximate it from $\y^{\delta}$ by solving a variational reconstruction problem:
\begin{equation}
    \underset{\x \in \X}{\min}\text{\,}\mathcal{L}_{\Y}\left(\y^{\delta},\Op{A}(\x)\right)+\alpha \R(\x).
    \label{var_recon}
\end{equation}
The loss functional $\mathcal{L}_{\Y}:\Y \times \Y \rightarrow{\mathbb{R}}$ provides a measure of data-fidelity and is typically chosen based on the statistical properties of the measurement noise $\e$, whereas the regularization functional $\R:\X \rightarrow{\mathbb{R}}$ penalizes undesirable images. The penalty parameter $\alpha>0$ trades-off data-fidelity with the regularization penalty and is chosen depending on the noise strength. In the subsequent part, we consider the squared-$\ell_2$ loss to measure data-fidelity, i.e., $\mathcal{L}_{\Y}(\y_1,\y_2)=\|\y_1-\y_2\|_{\Y}^2$, unless otherwise specified, and correspondingly, \eqref{var_recon} reduces to
\begin{equation}
    \underset{\x \in \X}{\min}\text{\,}\|\y^{\delta}-\Op{A}(\x)\|_{\Y}^2+\alpha \R(\x).
    \label{var_recon_L2}
\end{equation}
\subsection{Statistical formulation}
In the statistical formulation of inverse problems, one models the data as a single sample $\y^{\delta}$ of the $\Y$-valued random variable  
\begin{equation}
    Y = \Op{A}(X)+ \e,
    \label{inv_prob_data_stat}
\end{equation}
and aims to estimate the posterior distribution of $X$ conditioned on $Y=\y^{\delta}$, denoted as $\pi_{\text{post}}(X=\x|Y=\y^{\delta})$. Using the Bayes rule, the posterior distribution can be expressed in terms of the data-likelihood and the prior:
\begin{equation}
    \pi_{\text{post}}(X=\x|Y=\y^{\delta}) = \frac{\pi_{\text{data}}(Y=\y^{\delta}|X=\x)\pi_{X}(X=\x)}{Z(\y^{\delta})},
    \label{bayes}
\end{equation}
where $Z(\y^{\delta})$ is a normalizing constant independent of $\x$. While the data-likelihood is known in most inverse problems, the prior, which encodes a-priori belief about $\x$, is typically unknown. In the sequel, we write $\pi_{\text{post}}(X=\x|Y=\y^{\delta})$ as $\pi_{\text{post}}(\x|\y^{\delta})$, and likewise for the other probability measures in \eqref{bayes} for simplicity. An approximation of the true image is typically obtained by summarizing the posterior distribution into a point-estimate, such as the mean. Among many choices available for extracting a point-estimate from the posterior, a particularly popular one is to compute the \textit{mode}, leading to the so-called \textit{maximum a-posteriori probability} (MAP) estimate: 
\begin{equation}
    \underset{\x \in \X}{\min}\text{\,}-\log \pi_{\text{data}}(\y^{\delta}|\x)-\log \pi_{X}(\x).
    \label{map_recon}
\end{equation}
For a Gibbs-type prior $\pi_{X}(\x)\propto \exp\left(-\lambda \R(\x)\right)$, the MAP estimation problem \eqref{map_recon} is essentially equivalent to the variational reconstruction framework \eqref{var_recon} in the classical setting.
\section{Theoretical results}
In this section, we will first prove that the minimizer of the variational loss resulting from a strongly convex regularizer converges to the ground-truth. Subsequently, we develop a parametrization strategy using a neural network that achieves the properties required for convergence. We then explain the training protocol and describe the steps involved in learning the convex regularizer in a data-driven manner.
\subsection{Properties of strongly-convex regularizers}
First, we demonstrate that the variational problem corresponding to a strongly-convex regularizer is well-posed, and derive the resulting convergence and stability estimates. The analysis serves as a motivation for the parametrization of the regularizer discussed in the next section. In particular, we begin by investigating the analytical properties of a regularization functional of the form
\begin{equation}
    \R(\x) = \R'(\x) + \rho_0 \left\|\x\right\|_{\X}^2,
    \label{cvx_r_def1}
\end{equation}
where $\Vert \,\cdot\, \Vert_{\X}$ is the norm induced by the inner product structure of $\X$ and $\R':\X\rightarrow\mathbb{R}$ is 1-Lipschitz and convex in $\x$. 
No smoothness assumption is made on $\R'(\x)$. The corresponding reconstruction problem consists in minimizing the variational objective $J_{\alpha}\left(\x;\y^{\delta}\right)$ with respect to $\x$, where     
\begin{equation}
    J_{\alpha}\left(\x;\y\right) := \|\y-\Op{A}(\x)\|_{\Y}^2+\alpha \left(\R'(\x) + \rho_0 \left\|\x\right\|_{\X}^2\right).
    \label{var_loss_maindef1}
\end{equation}
The forward operator is assumed to be bounded and linear, so its operator norm is bounded, i.e., 
\[ \beta_1 := \displaystyle\underset{\x\in \X}{\sup}\text{\,}\frac{\left\|\Op{A}(\x) \right\|_{\Y}}{\left\|\x \right\|_{\X}}< \infty.
\]
In the following, we formally show the well-posedness of the variational problem corresponding to the objective in \eqref{var_loss_maindef1}. Specifically, we show that the variational objective $J_{\alpha}\left(\x;\y\right)$ has a unique minimizer $\hat{\x}_{\alpha}\left(\y\right)$, for any $\y$ and $\alpha>0$, varying continuously in the data $\y$. Further, when the noise level $\delta\rightarrow 0$, the minimizer of the variational loss $J_{\alpha}(\x;\y^{\delta})$ approaches the $\R$-minimizing solution $\x^{\dagger}$ given by
\begin{equation}
    \x^{\dagger} \in \underset{\x}{\argmin}\text{\,}\R(\x) \text{\,\,subject to\,\,}\Op{A}(\x)=\y^{0},
    \label{r_min_sol}
\end{equation}
where $\y^{0}$ is the clean data. Convergence to $\x^{\dagger}$ defined in \eqref{r_min_sol} holds provided that the regularization penalty $\alpha(\delta)$ is chosen appropriately as a function of $\delta$. These results can be derived as special cases from the general convex regularization theory \cite[Theorems 3.22, 3.23, and 3.26; Proposition 3.32]{scherzer2009variational}. Here, we present their proofs, which, by directly using strong convexity, are simpler than their more general counterparts. That is why we include them here to make the theoretical treatment self-contained. An interested reader must refer to \cite[Chapter 3]{scherzer2009variational} for more general proofs that do not require strong-convexity.

\begin{prop} (Existence and uniqueness)
\label{existence_uniqueness_prop}
$J_{\alpha}(\x; \y)$ is strongly convex in $\x$ with parameter $2\alpha \rho_0$ and has a unique minimizer $\hat{\x}_{\alpha}\left(\y\right)$ for every $\y$ and $\alpha>0$. We also have 
\begin{equation}
   J_{\alpha}(\x; \y)\geq J_{\alpha}\left(\hat{\x}_{\alpha}\left(\y\right);\y\right)+\alpha \rho_0\left\|\x- \hat{\x}_{\alpha}\left(\y\right)\right\|_{\X}^2,
   \label{strong_cvx_mainprop}
\end{equation}
for any $\x\in\X$.
\end{prop}
\noindent\textbf{Proof}: It follows from the definition of strong convexity that $h_{\mu}(\x)=h(\x)+\frac{\mu}{2}\left\|\x\right\|_{\X}^2$ is $\mu$-strongly convex when $h$ is convex. Since $\|\y-\Op{A}(\x)\|_{\Y}^2+\alpha\text{\,}\R'(\x)$ is convex when $\Op{A}$ is linear, it follows that $J_{\alpha}(\x; \y)$ is $2\alpha\rho_0$-strongly convex in $\x$, and consequently, for any $\x,\boldsymbol{v}\in\X$, we have that
\begin{equation}
    J_{\alpha}(\x; \y)\geq J_{\alpha}(\boldsymbol{v}; \y) + \langle(\x-\boldsymbol{v}),\boldsymbol{g}_{\boldsymbol v}\rangle +\alpha\rho_0\left\|\x-\boldsymbol{v}\right\|_{\X}^2,
    \label{strong_cvx_prop1}
\end{equation}
for all $\boldsymbol{g}_{\boldsymbol v}\in\partial J_{\alpha}(\boldsymbol{v};\y)$. In particular, when $\boldsymbol{v}=\hat{\x}$ is a minimizer of $J_{\alpha}(\cdot; \y)$, we have $\boldsymbol{0}\in \partial J_{\alpha}(\boldsymbol{v};\y)$, and therefore \eqref{strong_cvx_prop1} leads to 
\begin{equation}
    J_{\alpha}(\x; \y)\geq J_{\alpha}(\hat{\x}; \y)+\alpha\rho_0\left\|\x-\hat{\x}\right\|_{\X}^2.
    \label{strong_cvx_proof_unique1}
\end{equation}
\eqref{strong_cvx_proof_unique1} also ascertains that if there are two minimizers $\hat{\x}_1$ and $\hat{\x}_2$, one must have $\hat{\x}_1=\hat{\x}_2$, thereby guaranteeing uniqueness. The unique minimizer, denoted as $\hat{\x}_{\alpha}\left(\y\right)$, satisfies \eqref{strong_cvx_mainprop}.\hfill \qedsymbol

\begin{prop} (Stability) The optimal solution $\hat{\x}_{\alpha}\left(\y\right)$ is continuous in $\y$.
\label{prop_stability_prop}
\end{prop}
\noindent\textbf{Proof}: Denote a perturbation of magnitude $\delta_1$ on $\y$ as $$\y^{\delta_1}=\y+\boldsymbol{\delta}_1, \text{\,\,with\,\,}\left\|\boldsymbol{\delta}_1\right\|_{\Y}\leq \delta_1.$$
\noindent Define for any $\delta_1>0$ $$p_{\delta_1} := J_{\alpha}\left( \hat{\x}_{\alpha}\left(\y^{\delta_1}\right);\y\right) - J_{\alpha}\left( \hat{\x}_{\alpha}\left(\y^{\delta_1}\right);\y^{\delta_1}\right).$$ 
\noindent Clearly, $\underset{\delta_1 \rightarrow 0}{\lim}\text{\,}p_{\delta_1}=0$ since $J_{\alpha}(\x;\y)$ is continuous in $\y$ for any $\x\in\X$. Further, $p_{\delta_1}$ can be expressed as 
\begin{eqnarray}
p_{\delta_1} &=& \left[J_{\alpha}\left( \hat{\x}_{\alpha}\left(\y^{\delta_1}\right);\y\right) - J_{\alpha}\left( \hat{\x}_{\alpha}(\y);\y\right)\right] \nonumber\\&+& \left[J_{\alpha}\left( \hat{\x}_{\alpha}(\y);\y\right) - J_{\alpha}\left( \hat{\x}_{\alpha}(\y); \y^{\delta_1}\right)\right]\nonumber\\&+& \left[J_{\alpha}\left( \hat{\x}_{\alpha}(\y); \y^{\delta_1}\right) - J_{\alpha}\left( \hat{\x}_{\alpha}\left(\y^{\delta_1}\right);\y^{\delta_1}\right)\right].
\label{p_e_eq}
\end{eqnarray}
For convenience, denote the terms within square brackets in \eqref{p_e_eq} as $t_1$, $t_2$, and $t_3$, respectively. By Proposition \ref{existence_uniqueness_prop}, $$\displaystyle t_1,t_3\geq \alpha \rho_0 \left\|\x_{\alpha}\left(\y^{\delta_1}\right)-\x_{\alpha}(\y)\right\|_{\X}^2,$$ and by continuity of $J_{\alpha}(\cdot;\y)$ in $\y$, we have $\underset{\delta_1 \rightarrow 0}{\lim}\text{\,}t_2=0$. Therefore $\underset{\delta_1 \rightarrow 0}{\lim}\text{\,}\left(p_{\delta_1}-t_2\right)=0$, and \eqref{p_e_eq} implies that 
\begin{equation}
    p_{\delta_1} -t_2= t_1+t_3 \geq 2\alpha \rho_0 \left\|\hat{\x}_{\alpha}\left(\y^{\delta_1}\right)-\hat{\x}_{\alpha}(\y)\right\|_{\X}^2.
    \label{stability_step1}
\end{equation}
Further, we have that
\small
\begin{eqnarray}
    p_{\delta_1}-t_2 &=& \left\|\Op{A}\left(\hat{\x}_{\alpha}\left(\y^{\delta_1}\right)\right)-\y \right\|_{\Y}^2- \left\|\Op{A}\left(\hat{\x}_{\alpha}\left(\y^{\delta_1}\right)\right)-\y^{\delta_1}\right\|_{\Y}^2 \nonumber\\&+&\left\|\Op{A}\left(\hat{\x}_{\alpha}(\y)\right)-\y^{\delta_1} \right\|_{\Y}^2-\left\|\Op{A}\left(\hat{\x}_{\alpha}(\y)\right)-\y \right\|_{\Y}^2.
    % &=&2\boldsymbol\epsilon_y^\top\left(\Op{A}\left(\hat{\x}(\y+\boldsymbol\epsilon_y)\right)+\y\right)-\left\|\boldsymbol\epsilon_y\right\|_2^2 \nonumber\\&-& 2\boldsymbol\epsilon_y^\top\left(\Op{A}\left(\hat{\x}(\y)\right)+\y\right)+\left\|\boldsymbol\epsilon_y\right\|_2^2\nonumber\\
    % &=&2\boldsymbol\epsilon_y^\top \left(\Op{A}\left(\hat{\x}(\y+\boldsymbol\epsilon_y)-\hat{\x}(\y)\right) \right)\nonumber\\&\leq& 2\epsilon \beta_1 \left\|\hat{\x}(\y+\boldsymbol\epsilon_y)-\hat{\x}(\y)\right\|_2.
    \label{stability_est1}
\end{eqnarray}
\normalsize
Now, substituting $\y^{\delta_1}=\y+\boldsymbol{\delta}_1$ in \eqref{stability_est1} and expanding further,
\begin{eqnarray}
    p_{\delta_1}-t_2 &=& 2\langle \boldsymbol{\delta}_1, \Op{A}\left(\hat{\x}_{\alpha}\left(\y^{\delta_1}\right)\right)\rangle-\left\|\boldsymbol{\delta}_1\right\|_{\Y}^2\nonumber\\
    &-&2\langle \boldsymbol{\delta}_1, \Op{A}\left(\hat{\x}_{\alpha}\left(\y\right)\right)\rangle+\left\|\boldsymbol{\delta}_1\right\|_{\Y}^2\nonumber\\
    &=&2\langle \boldsymbol{\delta}_1, \Op{A}\left(\hat{\x}_{\alpha}\left(\y^{\delta_1}\right)-\hat{\x}_{\alpha}\left(\y\right)\right)\rangle\nonumber\\
    &\leq&2\beta_1\delta_1\left\|\hat{\x}_{\alpha}\left(\y^{\delta_1}\right)-\hat{\x}_{\alpha}(\y)\right\|_{\X},
    \label{stability_step1_2}
\end{eqnarray}
where the last inequality in \eqref{stability_step1_2} is due to Cauchy-Schwarz. Finally, combining \eqref{stability_step1} with \eqref{stability_step1_2}, we have
\begin{equation}
  \left\|\hat{\x}_{\alpha}\left(\y^{\delta_1}\right)-\hat{\x}_{\alpha}(\y)\right\|_{\X} \leq \frac{\beta_1\delta_1}{\alpha\rho_0}.
   \label{stability_est}
\end{equation}
\eqref{stability_est} indicates that $\underset{\delta_1\rightarrow 0}{\lim}\text{\,}\left\|\hat{\x}_{\alpha}\left(\y^{\delta_1}\right)-\hat{\x}_{\alpha}(\y)\right\|_{\X}=0$, confirming that $\hat{\x}_{\alpha}(\y)$ is continuous in $\y$ for a fixed $\alpha$. \hfill \qedsymbol \\
Unlike Theorem 3 in \cite{ar_nips}, Proposition 2 establishes convergence of $\hat{\x}_{\alpha}\left(\y^{(k)}\right)=\underset{\x}{\argmin}\,J_{\alpha}(\x;\y^{(k)})$ to $\hat{\x}_{\alpha}\left(\y\right)$ with respect to the norm topology $\left\|\cdot\right\|_{\X}$ on $\X$ when $\left\|\y^{(k)}-\y\right\|_{\Y}\rightarrow 0$, and \eqref{stability_est} yields  a stability estimate, in addition.
\begin{prop} (Convergence)
For $\delta\rightarrow 0$ and $\alpha(\delta) \rightarrow 0$ such that $\displaystyle\frac{\delta}{\alpha(\delta)}\rightarrow 0$, we have that $\hat{\x}_{\alpha}\left(\y^{\delta}\right)$ converges to the $\R$-minimizing solution $\x^{\dagger}$ given in \eqref{r_min_sol}.
% \begin{equation}
%     \x^{\dagger} = \underset{\x}{\argmin}\text{\,}\R(\x) \text{\,\,subject to\,\,}\Op{A}(\x)=\y^{0},
%     \label{r_min_sol}
% \end{equation}
% where $\y^{0}$ is the clean data .
\label{prop_conv_prop_var_opt}
\end{prop}
\noindent\textbf{Proof}: 
By \eqref{stability_est}, we have that
\begin{equation}
    \left\|\hat{\x}_{\alpha}\left(\y^{\delta}\right)-\hat{\x}_{\alpha}\left(\y^{0}\right) \right\|_{\X}\leq \frac{\beta_1\delta}{\alpha \rho_0}.
    \label{conv_step1}
\end{equation}
Since $\Op{A}$ is linear and $\R$ is strongly-convex, the solution $\x^{\dagger}$ to \eqref{r_min_sol} is unique and it can alternatively be expressed as
\begin{eqnarray}
    \x^{\dagger} &=& \underset{\alpha\rightarrow 0+}{\lim}\left(\underset{\x}{\argmin}\text{\,}\R(\x)+\frac{1}{\alpha}\left\|\Op{A}(\x)-\y^{0} \right\|_{\Y}^2\right)\nonumber\\
    &=& \underset{\alpha\rightarrow 0+}{\lim}\left(\underset{\x}{\argmin}\text{\,}\alpha\R(\x)+\left\|\Op{A}(\x)-\y^{0} \right\|_{\Y}^2\right)\nonumber\\
    &=&\underset{\alpha\rightarrow 0+}{\lim}\text{\,\,}\hat{\x}_{\alpha}\left(\y^{0}\right).
    \label{x_alpha_conv}
\end{eqnarray}
Let $\epsilon(\alpha)=\left\|\hat{\x}_{\alpha}\left(\y^{0}\right) - \x^{\dagger}\right\|_{\X}$. Then, by \eqref{x_alpha_conv}, $\underset{\alpha\rightarrow 0}{\lim}\text{\,}\epsilon(\alpha)=0$. Thus, combining \eqref{conv_step1} and \eqref{x_alpha_conv}, and using the triangle inequality, one can argue that
\begin{eqnarray}
    \left\|\hat{\x}_{\alpha}\left(\y^{\delta}\right) - \x^{\dagger}\right\|_{\X}&\leq& \left\|\hat{\x}_{\alpha}\left(\y^{\delta}\right) - \hat{\x}_{\alpha}\left(\y^{0}\right)\right\|_{\X}\nonumber\\
    &+& \left\|\hat{\x}_{\alpha}\left(\y^{0}\right) - \x^{\dagger}\right\|_{\X}\leq\frac{\beta_1\delta}{\alpha \rho_0} + \epsilon(\alpha).\nonumber
\end{eqnarray}
Now, if $\underset{\delta\rightarrow 0}{\lim}\text{\,}\alpha(\delta)\rightarrow 0$ and $\underset{\delta\rightarrow 0}{\lim}\text{\,}\frac{\delta}{\alpha(\delta)}\rightarrow 0$, the inequality above implies that $\underset{\delta\rightarrow 0}{\lim}\text{\,}\left\|\hat{\x}_{\alpha}\left(\y^{\delta}\right) - \x^{\dagger}\right\|_{\X}= 0$.\hfill\qedsymbol
\begin{figure}
\subfigure{
\includegraphics[width=3.40in]{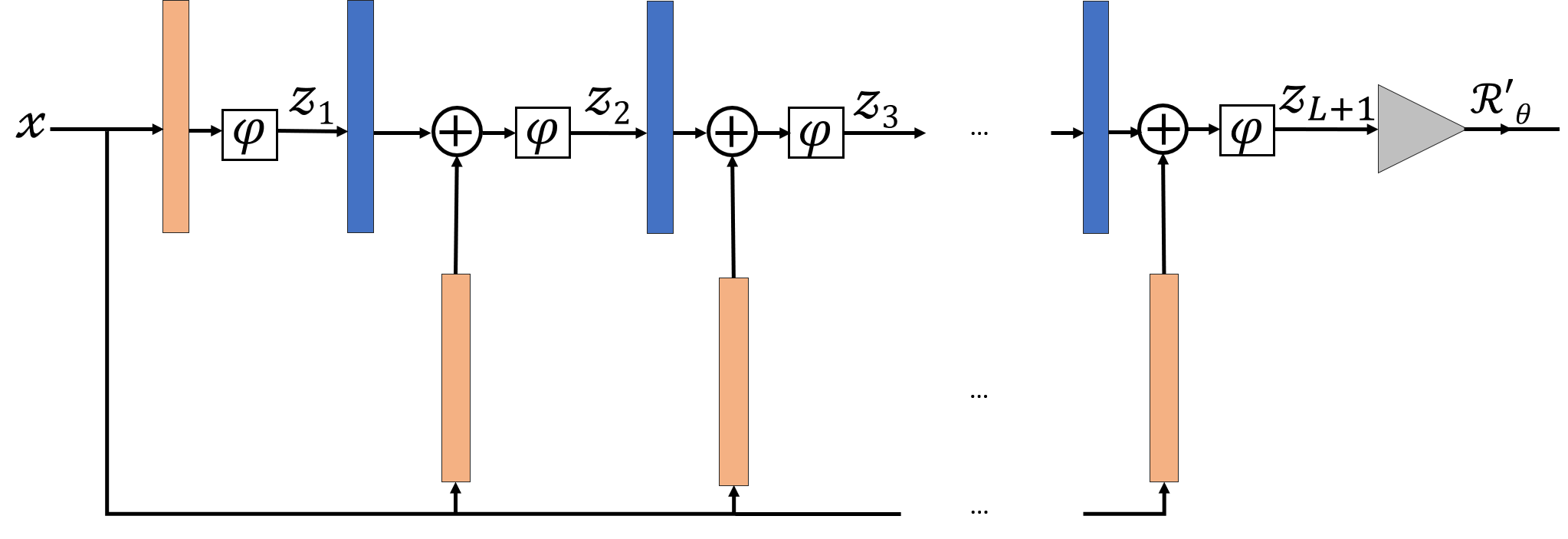}}
\caption{Architecture of the convex regularizer. The blue rectangles indicate the layers $\left\{\mathcal{B}_i\right\}_{i=1}^{L}$ that are allowed to contain only non-negative weights for convexity of the output, while the rectangles in orange denote layers with weights and biases $\left\{\mathcal{W}_i,\boldsymbol{b}_i\right\}_{i=0}^{L}$ which are allowed to take any real value. The triangle in the end represents a global average-pooling layer. The activations $\varphi_i$ in all layers are chosen to be leaky-ReLU and denoted by $\varphi$.}
\label{net_arch_fig}
\end{figure}
%%%%%%%%%
\subsection{Input-convex neural networks}
\label{icnn_sec}
This section introduces ICNNs in the infinite dimensional setting, which boils down to the construction in \cite{amos2017input} in the finite dimensional case. In particular, we need to implement the convex functional $\R'$ in \eqref{cvx_r_def1} on $\mathbb{L}^2$ spaces, i.e, $\X=\mathbb{L}^2$.
Discretizing this parameterized operator coincides with the convex network construction presented in~\cite{amos2017input}. More precisely, define the activation-spaces of our network to be $\V_i := \mathbb{L}^2\left([0,1]^{n_i}\right)$ for $i=0,\cdots, L-1$, with $\V_0=\X$ and $\V_L=\mathbb{R}$. We assume the input $\x$ of our network to be in $\V_0$ and set $\boldsymbol{0}=:\z_0\in \V_0$. We then define the output of each layer $i=0,\cdots, L$ to be
\begin{equation}
    \z_{i+1}(\x) = \varphi_i\left(\B_i\left( \z_i(\x)\right) + \W_i(\x) + \boldsymbol{b}_i\right).
    \label{construct_icnn_eq}
\end{equation}
Here $\B_i:\V_i\to \V_{i+1}$ and $\W_i:\V_0\to \V_{i+1}$ are bounded integral transforms and we assume the kernels of the $\B_i$'s to be pointwise non-negative. The activations $\varphi_i:\V_{i+1}\to \V_{i+1}$ are given by the pointwise application of a convex monotone function from $\mathbb{R}$ to $\mathbb{R}$. Further let the biases, $\boldsymbol{b}_i$, be in $\V_{i+1}$. The overall output of the network $\R':\V_0\to\mathbb{R}$ is obtained by applying a global-average pooling operator $\mathcal{H}_{\text{avg}}$ on $\z_{L+1}$, i.e., $\R'(\x) := \mathcal{H}_{\text{avg}}(\z_{L+1}(\x))$. In practice, $\left\{\B_i,\W_i\right\}$'s represent convolutional layers. The combined set of parameters $\left\{\B_i,\W_i,\boldsymbol{b}_i\right\}_{i=0}^{L}$ of the convolutional layers in $\R'$ is denoted by the shorthand notation $\theta$ and the corresponding parametric regularizer is written as $\R'_{\theta}$. A schematic diagram of the architecture of $\R'_{\theta}$ is shown in Figure \ref{net_arch_fig}. \\
The convexity of $\R'_{\theta}(\x)$ in $\x$ can be argued using a recursive logic which uses the following two facts \cite{boyd2004convex}:
\begin{itemize}[leftmargin=*]
    \item Non-negative combination of finitely many convex functions is another convex function, and
    \item the composition $\psi_1\circ \psi_2:\mathbb{V}\rightarrow \mathbb{R}$ of two functions $\psi_1$ and $\psi_2$ is convex when $\psi_2$ is convex and $\psi_1$ is convex and monotonically non-decreasing.
\end{itemize} 
We note that $\z_{1}(\x) = \varphi_0\left(\W_0(\x) + \boldsymbol{b}_0\right)$ is convex in $\x$, since $\z_1$ is the composition of a convex and monotonically non-decreasing activation $\varphi_0$ (leaky-ReLU, in particular) with an affine (and hence convex) function of $\x$. Further, when $\z_i$ is convex and the weights in $\mathcal{B}_i$ are non-negative, the argument of $\varphi_i$ in \eqref{construct_icnn_eq} is convex in $\x$, and so is $\z_{i+1}$, since $\varphi_i$ is convex and monotonically non-decreasing (also chosen to be leaky-ReLU activation). Consequently, $\z_{L+1}$ is convex in $\x$ and so is the final output $\R'_{\theta}$, since the average-pooling operator $\mathcal{H}_{\text{avg}}$ maintains convexity. To preserve convexity during training, one has to ensure that the weights in $\mathcal{B}_i$ remain non-negative throughout training. This is achieved by applying a zero-clipping operation on the negative weights following each weight update.

The specific choice of the hyper-parameters in the architecture in the discretized implementation, such as the number of layers, kernel size, etc., is mentioned in the context of the specific experiments in Sec.~\ref{experiment_sec}.  

\begin{algorithm}[t]
\caption{Training the ACR via \eqref{r_opt}.}
\begin{algorithmic}
\STATE {\bf  1.} {\bf Input:} gradient penalty $\lambda_{\text{gp}}$, initial value of the network parameters $\theta^{(0)}=\left\{\B_i\geq 0, \W_i, \boldsymbol{b}_i\right\}$, mini-batch size $n_{b}$, parameters $\left(\eta,\beta_1,\beta_2\right)$ for the Adam optimizer.
\STATE {\bf  2.} {\bf for mini-batches $m=1,2,\cdots$, do (until convergence)}: 
\begin{itemize}[leftmargin=*]
\item Sample $\x_j\sim\pi_X$, $\y_j\sim \pi_Y$, and $\epsilon_j\sim \text{uniform}\,[0,1]$; for $1 \leq j \leq n_b$. Compute $\x^{(\epsilon)}_j=\epsilon_j \x_j+\left(1-\epsilon_j\right)\Op{A}^{\dagger}\y_j$.
\item Compute the training loss for the $m^{\text{th}}$ mini-batch:
\begin{eqnarray*}
   \mathcal{L}\left(\theta\right)&:=&\frac{1}{n_b}\sum_{j=1}^{n_b}\R_{\theta}(\x_j)-\frac{1}{n_b}\sum_{j=1}^{n_b}\R_{\theta}(\Op{A}^{\dagger}\y_j) \\
   &+& \lambda_{\text{gp}}\cdot\frac{1}{n_b}\sum_{j=1}^{n_b}\left(\left\|\nabla \R_{\theta}\left(\x^{(\epsilon)}_j\right)\right\|_2-1\right)^2.
\end{eqnarray*}
\item Update $\theta^{(m)}=\text{Adam}_{\eta,\beta_1,\beta_2}\left(\theta^{(m-1)},\nabla_{\theta}\mathcal{L}\left(\theta^{(m-1)}\right)\right)$.
\item Zero-clip the negative weights in $\B_i$ to preserve convexity.
\end{itemize}
\STATE {\bf  3.} {\bf Output:} Parameter $\theta$ of the trained ACR.
\end{algorithmic}
\label{algo_acr_train}
\end{algorithm}
%%%%%%%%%%%%%%%%%%%%%%%%
%%%%%%%%%
\begin{figure}
\centering
% \subfigure{
% \includegraphics[width=2.8in]{figures/paired_vs_unpaired_train_combined.png}}
\subfigure[paired]{
\includegraphics[width=1.60in]{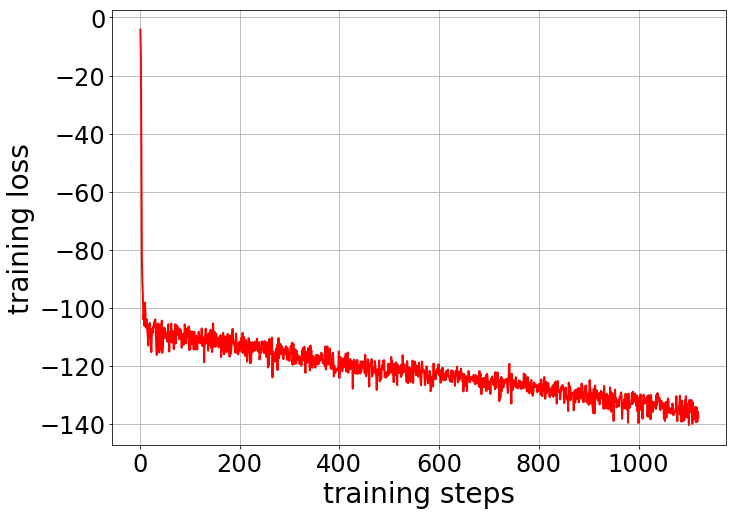}}
\subfigure[unpaired]{
\includegraphics[width=1.60in]{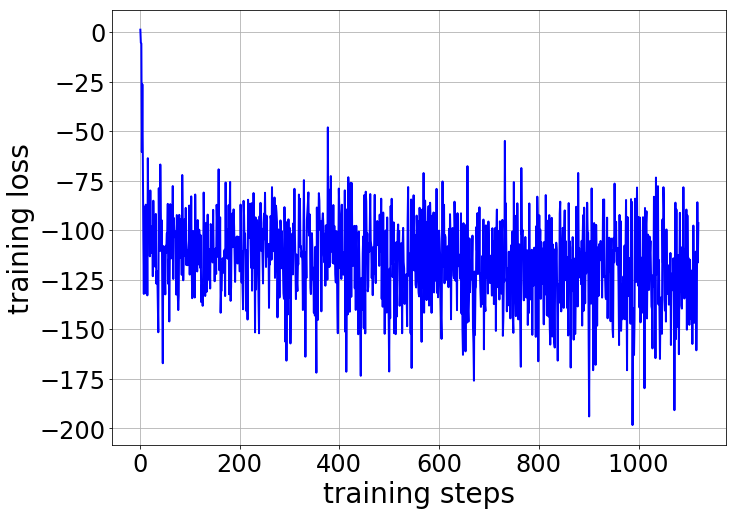}}
\subfigure[paired: 31.9044 dB, 0.8743]{
\includegraphics[width=1.65in]{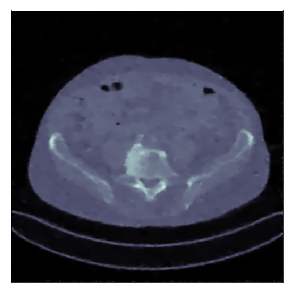}}
\subfigure[unpaired: 31.8323 dB, 0.8722]{
\includegraphics[width=1.65in]{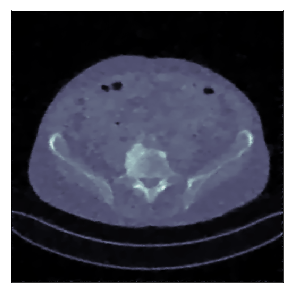}}
\caption{Training ACR on paired vs. unpaired data for sparse-view CT, where each training step corresponds to update over 20 mini-batches. Training on paired data is more stable. The reconstruction produced by ACR trained on paired data is nearly indistinguishable from ACR trained on unpaired data, both qualitatively and in terms of objective measures of reconstruction quality, such as the peak signal-to-noise ratio (PSNR) (measured in dB) and the structual similarity index (SSIM).}
\label{paired_vs_unpaired_fig}
\end{figure}
%%%%%%%%%
\subsection{Learning objective and protocol}
\label{learning_protocol_sec}
The training objective and protocol adopted for ACR follow the same philosophy introduced in \cite{ar_nips} for training AR. The regularizer is trained with the objective of favoring solutions that are similar to the ground-truth images in the training dataset and penalizing reconstructions with artifacts. Consequently, it should produce a small output when a true image is given as input and a large output when it is presented with an unregularized reconstruction. The distribution of unregularized reconstructions is denoted as $\Op{A}^\dagger_{\#}\pi_{Y}$, the push-forward of the measurement distribution $\pi_Y$ by the pseudo-inverse of the forward operator. Naturally, during training, one seeks to minimize the average output of the regularizer $\R_{\theta}$ when the input is sampled from the true image distribution $\pi_X$, while simultaneously maximizing the average output of $\R_{\theta}$ when the input is an unregularized reconstruction. This is equivalent to minimizing the difference of the average output of $\R_{\theta}$ over the true image distribution $\pi_X$ and the distribution $\Op{A}^\dagger_{\#}\pi_{Y}$ of unregularized solutions and can be posed as
\begin{eqnarray}
\theta^*&=&\underset{\theta}{\argmin}\left(\mathbb{E}_{\pi_X}\left[\R_{\theta}\left(X\right)\right]-\mathbb{E}_{\Op{A}^\dagger_{\#}\pi_{Y}}\left[\R_{\theta}\left(X\right)\right]\right)\nonumber\\&&\text{\,\,subject to\,\,}\R_{\theta}\in 1-\text{Lipschitz}.
\label{r_opt}
\end{eqnarray}
The 1-Lipschitz condition in \eqref{r_opt} encourages the output of $\R_{\theta}$ to transition smoothly with respect to the input, thus making the corresponding variational loss stable. The 1-Lipschitz constraint is enforced by adding a gradient-penalty term \cite{wgan_gp} of the form 
\begin{equation}
    \mathcal{L}_{\text{gp}}=\lambda_{\text{gp}}\text{\,}\mathbb{E}_{\pi_{X^{(\epsilon)}}}\left[\left(\left\|\nabla \R_{\theta}\left(X^{(\epsilon)}\right)\right\|_2-1\right)^2\right],
    \label{grad_penalty}
\end{equation}
to the training objective in \eqref{r_opt}. Here, $X^{(\epsilon)}$ is uniformly sampled on the line-segment between the images $X$ and $\Op{A}^{\dagger}Y$. The actual steps involved in approximating the training objective in \eqref{r_opt} and imposing the gradient penalty \eqref{grad_penalty} for training $\R_{\theta}$ are listed in Algorithm \ref{algo_acr_train}. 

The training protocol is unsupervised in principle, since it suffices to have $N$ i.i.d. samples $\left\{\x_i\right\}_{i=1}^{N}\in \X$ and $\left\{\y_i\right\}_{i=1}^{N}\in\Y$ drawn from the marginal distributions $\pi_X$ and $\pi_Y$, respectively, for approximating the loss in \eqref{r_opt}. Nevertheless, in practice, one has two options when it comes to estimating the loss in \eqref{r_opt} by using empirical average over mini-batches: (i) using paired samples $\{\x_i,\Op{A}^{\dagger}\y_i\}$, where $\x_i\sim\pi_X$ and
\begin{equation*}
    \y_i\sim\pi_{\text{data}}(\y_i|\x_i)=\pi_{\text{noise}}\left(\y_i-\Op{A}(\x_i)\right); \text{\,\,and}
    \label{sample_y_eq}
\end{equation*}
(ii) using unpaired samples $\{\x_i,\Op{A}^{\dagger}\y_{(i)}\}$, where the samples $\left\{\y_{(i)}\right\}_{i=1}^{N}$ are obtained by applying a random permutation on $\y_i\sim \pi_{\text{data}}(\y_i|\x_i)$. The actual training Process would be supervised or unsupervised depending on whether one chooses the first or the second option, respectively. Notably, both options lead to identical outcomes in the limit of infinite amount of training data. However, for limited amount of data, we observe that supervised training leads to a more graceful evolution of the training loss with respect to training steps, while unsupervised training leads to considerably more fluctuations in the training loss. A comparison of the supervised and unsupervised approaches for training the ACR is shown in Fig. \ref{paired_vs_unpaired_fig} for sparse-view CT reconstruction (see Sec. \ref{sv_ct_sec} for details). As the training objective evolves in a more stable manner when paired samples are used for approximation, we adopt a supervised learning process in the face of limited training data (similarly for AR as well to ensure a fair comparison of their performance). Nevertheless, the reconstruction metrics obtained using ACRs trained on paired and unpaired examples turn out to be nearly identical (see Figures \ref{paired_vs_unpaired_fig}(c) and \ref{paired_vs_unpaired_fig}(d)). A similar behavior during training and reconstruction was observed for the other two inverse problems considered in Section \ref{experiment_sec}, namely limited-view CT reconstruction and image deblurring.
\section{Convergence of the sub-gradient algorithm}
Although the regularizer is strongly-convex, it seems exceedingly difficult to come up with a closed-form expression for its proximal operator~\cite{parikh2014proximal}, thus preventing the use of proximal gradient-descent. We, however, show that a sub-gradient descent algorithm converges to the minimizer of the variational loss.  We begin by rewriting the variational loss defined in \eqref{var_loss_maindef1} as
\begin{equation}
    J_{\alpha}(\x;\y)=f_{\alpha}(\x;y)+\alpha \text{\,}g(\x),
    \label{smooth_plus_nonsmooth}
\end{equation}
where $f_{\alpha}(\x;\y) = \left\|\Op{A}(\x)-\y \right\|_{\Y}^2+\alpha\rho_0\left\|\x\right\|_{\X}^2$ is smooth, strongly-convex, and differentiable, while $g(\x)=\R'(\x)$ is convex and 1-Lipschitz. Note that $g$ need not be differentiable. For brevity of notation, we drop the subscript $\alpha$ in the loss functions and denote the Lipschitz constant of $g$ by $L_g$ for generality in the remainder of this section (the gradient penalty in Algorithm \ref{algo_acr_train} enforces $L_g$ to be close to one). Since the forward operator is linear, $\nabla f(\cdot;\y)$ is Lipschitz-continuous, with a Lipschitz constant which we denote as $L_{\nabla f}$.

The convergence of gradient-descent for convex optimization has been widely studied. To put our analysis into perspective and to outline the points of difference of our analysis with the classical convex optimization literature, we recall the following well-known convergence results for gradient-descent applied to $\underset{\x}{\min}\,J(\x)$:
\begin{itemize}[leftmargin=*]
    \item When $J(\x)$ is convex and has an $L_{\nabla J}$-Lipschitz gradient, gradient-descent with step-size $\eta_k\leq \frac{2}{L_{\nabla J}}$ generates a sequence of iterates $\x_k$ satisfying $J(\x_k)-J(\hat{\x})\leq \mathcal{O}\left(\frac{1}{k}\right)$ \cite[Thm. 2.1.13]{nesterov_tutorial_cvx}.
    \item When $J(\x)$ is $\mu_J$-strongly convex, with an $L_{\nabla J}$-Lipschitz gradient, one can show a linear convergence rate 
    \begin{equation*}
        \left\|\x_k-\hat{\x}\right\|_2^2 \leq \left(\frac{q_0-1}{q_0+1}\right)^k\,\left\|\x_0-\hat{\x}\right\|_2^2,
    \end{equation*}
    for $\eta_k\leq \frac{2}{\mu_J+L_{\nabla J}}$, where $q_0=\frac{L_{\nabla J}}{\mu_J}$ \cite[Thm. 2.1.14]{nesterov_tutorial_cvx}.
    \item When $J$ is convex, non-smooth, the sub-gradients of $J$ are bounded by $L_J$ (i.e., $J$ is $L_J$-Lipschitz), the step-sizes satisfy $\sum_{k=1}^{\infty}\eta_k^2<\infty$, and $\sum_{k=1}^{\infty}\eta_k=\infty$, the following convergence result holds for the sequence of iterates generated by the sub-gradient update rule (see \cite[Sec. 3]{subgrad_convergence} for a proof):
    \begin{equation*}
        \underset{k\rightarrow \infty}{\lim}\,\,J_{\text{best}}^{(k)}=J(\hat{\x}), \text{\,where\,}J_{\text{best}}^{(k)}=\underset{0\leq i\leq k}{\min}\,J(\x_i).
    \end{equation*}
\end{itemize} 
The structure of our variational loss in \eqref{smooth_plus_nonsmooth} is non-standard, in the sense that it does not fit into any of these classical results stated above. Therefore, we perform an independent analysis of the convergence of sub-gradient updates starting from the first principle, using the properties of $J_{\alpha}$ in \eqref{smooth_plus_nonsmooth}. More precisely, we establish a sub-linear rate of convergence of $\left\|\x_k-\hat{\x}\right\|_2^2$ to $0$, where $\left\{\x_k\right\}_{k=1}^{\infty}$ is the sequence of updates generated by the sub-gradient algorithm. 

\indent The following lemma ascertains the existence of a step-size parameter that leads to a convergent sub-gradient algorithm. Nevertheless, in practice, we found that an appropriately chosen constant step-size works reasonably well. 
% Notably, the standard sublinear convergence of subgradient updates \cite[Sec. 3]{subgrad_convergence} is not applicable in our case since the variational objective $J(\cdot;\y)$ is not Lipschitz-continuous. 
\begin{lemma}
\label{subgrad_convergence_lemma} (Convergence of sub-gradient updates)
Starting from any initial estimate $\x_0$, there exist optimal step-sizes $$\displaystyle\eta_k^*=2\alpha\rho_0\frac{\left\|\x_{k}- \hat{\x}\right\|_{\X}^2}{\left\|\boldsymbol{z}_k\right\|_{\X}^2}$$ such that the updates  $\x_{k+1}=\x_{k}-\eta_{k}^*\z_k$, where $\z_k=\nabla f(\x_k;\y)+\boldsymbol{u}_k$, with $\boldsymbol{u}_k\in \partial \left(\alpha g(\x_k)\right)$, converge to the minimizer $\hat{\x}$ of $J_{\alpha}(\x;\cdot)$ defined in \eqref{smooth_plus_nonsmooth}, i.e., $\underset{k \rightarrow \infty}{\lim}\text{\,}\x_k=\hat{\x}$ in $\X$ with respect to the norm topology.
\end{lemma}
\noindent\textbf{Proof}: Let $e_k=\left\|\x_k-\hat{\x} \right\|_{\X}^2$ be the squared estimation error in the $k^{\text{th}}$ iteration, and let $\mu= 2\alpha \rho_0$. Omitting the argument $\y$ in $J(\x;\y)$ for simplicity and using $\mu$-strong convexity of $J(\x)$, we have
\begin{equation}
    J(\x_k)\leq J(\hat{\x})+\left\langle\boldsymbol{z}_k,\x_k-\hat{\x}\right\rangle-\frac{\mu}{2} \left\|\x_k- \hat{\x}\right\|_{\X}^2.
    \label{ub_Jk}
\end{equation}
From Proposition \ref{existence_uniqueness_prop}, we get $J(\x_k)- J(\hat{\x})\geq \frac{\mu}{2}\left\|\x_k- \hat{\x}\right\|_{\X}^2$, which, combined with \eqref{ub_Jk}, leads to $$\left\langle\boldsymbol{z}_k,\x_k-\hat{\x}\right\rangle\geq \mu\left\|\x_k- \hat{\x}\right\|_{\X}^2.$$ 
Therefore, we have the following bound on $e_{k+1}$:
\begin{eqnarray}
    e_{k+1}&=&\left\|\x_{k+1}- \hat{\x}\right\|_{\X}^2 = \left\|\x_{k}- \eta_k \boldsymbol{z}_k- \hat{\x}\right\|_{\X}^2\nonumber\\ 
    &=& \left\|\x_{k}- \hat{\x}\right\|_{\X}^2-2\eta_k\left\langle\boldsymbol{z}_k,\x_k-\hat{\x}\right\rangle+\eta_k^2\left\|
    \boldsymbol{z}_k\right\|_{\X}^2\nonumber\\ 
    &\leq& (1-2\mu\eta_k)e_k+\eta_k^2\left\|
    \boldsymbol{z}_k\right\|_{\X}^2. 
    \label{eta_bdd}
\end{eqnarray}
The bound in \eqref{eta_bdd} becomes the tightest when $$\eta_k=\eta_k^*=\frac{\mu e_k}{\left\|\boldsymbol{z}_k\right\|_{\X}^2}=\mu\frac{\left\|\x_{k}- \hat{\x}\right\|_{\X}^2}{\left\|\boldsymbol{z}_k\right\|_{\X}^2},$$ leading to $\displaystyle e_{k+1}\leq e_k-\frac{\mu^2 e_k^2}{\left\|
    \boldsymbol{z}_k\right\|_{\X}^2}$. This indicates that the estimation error decreases monotonically with iteration. Further, $\left\|
    \boldsymbol{z}_k\right\|_{\X}^2$ can be bounded as
\begin{eqnarray}
    \left\|\boldsymbol{z}_k\right\|_{\X}^2 &=&\left\|\nabla f(\x_k)+\boldsymbol{u}_k\right\|_{\X}^2 \leq 2 \left\| \nabla f(\x_k)\right\|_{\X}^2+ 2\left\|\boldsymbol{u}_k\right\|_{\X}^2 \nonumber\\ &\leq& 2 \left\| \nabla f(\x_k)\right\|_{\X}^2 + 2\alpha^2L_g^2,
    \label{zk_bdd1}
\end{eqnarray}
as $g$ is $L_g$-Lipschitz. Since $\hat{\x}$ is the unique minimizer of $J(\x)$, $\exists\text{\,}\hat{\boldsymbol{u}}\in \partial \left(\alpha g(\hat{\x})\right)$ such that $\nabla f(\hat{\x})+\hat{\boldsymbol{u}}=0$. Using the triangle inequality and Lipschitz-continuity of $\nabla f(\x)$, 
\begin{eqnarray}
    \left\|\nabla f(\x_k) \right\|_{\X} - \left\| \hat{\boldsymbol{u}}\right\|_{\X} &\leq& \left\| \nabla f(\x_k)+\hat{\boldsymbol{u}}\right\|_{\X} \nonumber\\&=&\left\| \nabla f(\x_k)-\nabla f(\hat{\x})\right\|_{\X} \nonumber\\ &\leq& L_{\nabla f} \left\| \x_k-\hat{\x}\right\|_{\X},
    \label{grad_f_bdd1}
\end{eqnarray}
leading to $\left\|\nabla f(\x_k) \right\|_{\X}\leq L_{\nabla f} \sqrt{e_k}+ \alpha L_g$. Plugging this in \eqref{zk_bdd1} and substituting $\left\|\boldsymbol{z}_k\right\|_{\X}^2$ in \eqref{eta_bdd} with its resulting upper bound, we arrive at
\begin{equation}
    e_{k+1}\leq e_k-\frac{\mu^2 e_k^2}{2 L_{\nabla f}^2 e_k+4\alpha L_g L_{\nabla f}\sqrt{e_k}+4\alpha^2L_g^2}.
    \label{final_bdd2}
\end{equation}
Finally, since $0\leq e_{k+1}<e_k$, $\underset{k \rightarrow \infty}{\lim}\text{\,}e_k$ exists. Denote the limit by $e$. Taking limit as $k\rightarrow\infty$ on both sides of \eqref{final_bdd2}, we get
\begin{equation}
    e\leq e-\frac{\mu^2 e}{2 L_{\nabla f}^2 e+4\alpha L_g L_{\nabla f}\sqrt{e}+4\alpha^2L_g^2},
    \label{final_bdd2_1}
\end{equation}
implying that $e=0$; therefore proving $\underset{k \rightarrow \infty}{\lim}\text{\,}\x_k=\hat{\x}$. In other words, the sequence $\left\{\x_k\right\}_{k=1}^{\infty}$ converges to the unique optimal solution $\hat{\x}$ to the variational problem. \hfill \qedsymbol
%%%%
\begin{figure*}[t]
	\centering
	\subfigure[\small{ground-truth}]{
		\includegraphics[width=2.0in]{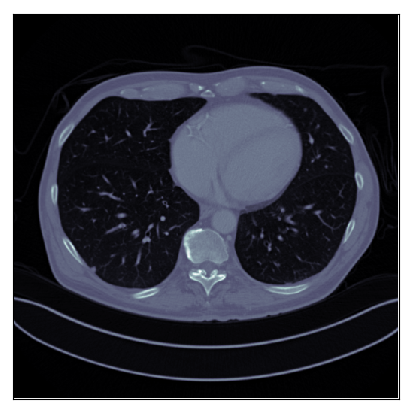}}
	\subfigure[\small{FBP: 21.6262 dB, 0.2435}]{
		\includegraphics[height=2.0in]{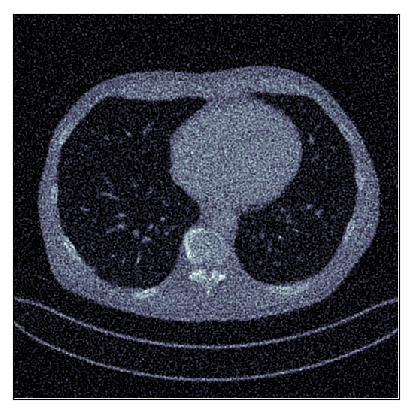}}
	\subfigure[\small{TV: 29.2506 dB, 0.7905}]{
		\includegraphics[width=2.00in]{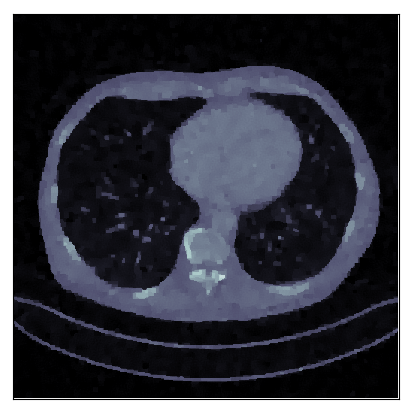}}\\
	\subfigure[\small{LPD: 33.6218 dB, 0.8871}]{
		\includegraphics[width=2.0in]{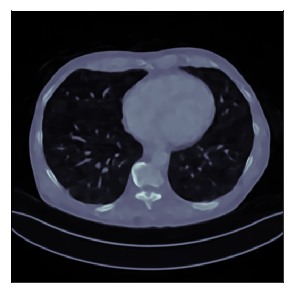}}
	\subfigure[\small{AR: 31.8257 dB, 0.8445}]{
		\includegraphics[width=2.0in]{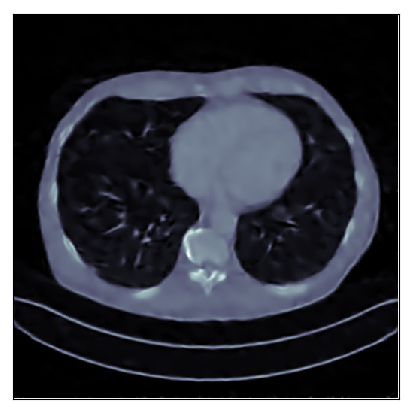}}
	\subfigure[\small{ACR: 30.0016 dB, 0.8246}]{
		\includegraphics[width=2.0in]{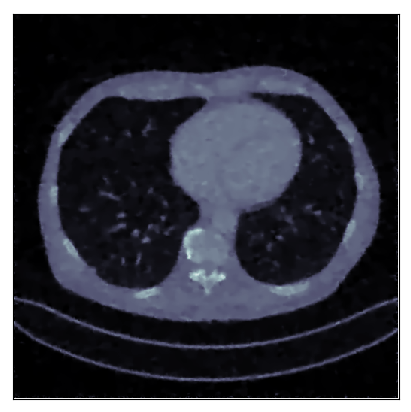}}
	\caption{\small{Comparison of different reconstruction methods for sparse-view CT. The learned ACR approach outperforms the most widely used TV-based convex regularization. AR turns out to be better than ACR in this case, pointing to the limited expressive power of a convex regularizer. LPD trained on supervised data leads to the best performance.}}
	%\vspace{-0.1in}
	\label{ct_image_figure}
\end{figure*}
%%%%
\section{Numerical results}
\label{experiment_sec}
For performance evaluation and comparison with the state-of-the-art, we consider three applications, namely, CT reconstruction with (i) sparse-view and (ii) limited-angle projection, and (iii) natural image deblurring. For the CT experiments, human abdominal CT scans for 10 patients provided by Mayo Clinic for the low-dose CT grand challenge \cite{mayo_ct_challenge} are used. We simulate the projection data by using the ODL library \cite{odl} with a GPU-accelerated \textit{astra} back-end. Our training dataset for the CT experiments consists of a total of 2250 2D slices, each of dimension $512 \times 512$, corresponding to 9 patients. The remaining 128 slices corresponding to one patient are used to evaluate the reconstruction performance. The deblurring experiments are conducted on the \textit{STL-10} dataset~\cite{coates2011analysis}. In all cases, we measure the performance in terms of the peak signal-to-noise ratio (PSNR) and the structural similarity index (SSIM) \cite{ssim_paper_2004}. The PSNR (in dB) is calculated as
\begin{equation*}
    \text{PSNR}=10\,\log_{10}\frac{\x^{*2}_{\max}}{\frac{1}{P^2}\sum_{i,j=1}^{P}\left(\x^{*}_{ij}-\hat{\x}_{ij}\right)^2},
\end{equation*}
where $\x^{*}_{\max}=1.0$ is the maximum allowable intensity in our experiments (since the intensity values are normalized to $[0,1]$). Here, $\x^*$ and $\hat{\x}$ denote the ground-truth and the reconstructed images (of size $P\times P$), respectively. In all settings, we compare our results also with total-variation (TV) reconstructions, computed by employing the ADMM-based solver in ODL. We choose the penalty parameter of each variational method to maximize the mean PSNR of its reconstructions. Besides the parametrization of ACR elucidated in Sec.~\ref{icnn_sec}, we compare the performance of three simple variants of it. In the first variant, we incorporate an additional regularization term
\begin{equation}
    \R'_{\text{sfb}}(\x)  = \|\U\x\|_1,
    \label{sfb}
\end{equation}
where $\U$ denotes a two-dimensional convolutional layer, into the regularizer. When the filters in $\U$ have bounded norms, this additional term is also convex and Lipschitz-continuous, like $\R'(\x)$. We refer to this term as \textit{sparsifying filter-bank} (SFB) penalty, since the $\ell_1$ penalty on $\U\x$ is tantamount to seeking sparsity of $\x$ in the learned filters $\U$. In other words, learning the filters $\U$ in the SFB penalty is equivalent to learning a sparsifying convolutional transform proposed in \cite{sfb_bresler,conv_transform_learning_angshul}, albeit with a completely different training strategy. While the key characteristic of the adversarial learning strategy adopted in this paper is to tell apart true images from the noisy ones, traditional approaches for learning SFB-like priors entail minimizing an objective that comprises data representation error, sparsity prior, and suitable regularization terms on the transform to avoid degenerate solutions \cite{sparse_t,conv_transform_learning_angshul}. The additional SFB term is motivated by the widely used wavelet-sparsity prior in image processing and essentially gives ACR a chance to learn a similar prior, and possibly a better one, during training. The other two variants correspond to only the ACR without any SFB term and the SFB term alone as a prior. We compare the performance of all three variants of ACR, namely with (i) $\R'(\x)$ and (ii) $\R'(\x)+\R'_{\text{sfb}}(\x)$ as the Lipschitz convex part, and (iii) $\R'_{\text{sfb}}(\x)  = \|\U\x\|_1$ as the standalone Lipschitz-convex part to investigate if a simpler and more commonly used convex model indeed works on par or better than the ICNN model in Sec.~\ref{icnn_sec}. While learning the SFB regularizer standalone or in combination with $\R'(\x)$, we add an $\ell_2$ penalty term on the SFB weights to control the Lipschitz constant of the resulting regularizer. For all of these variants of ACR, the same training strategy is adopted, as described in Sec.~\ref{learning_protocol_sec}.\\
\noindent The individual details of the experimental setups for all three applications are described in the following.
% Additionally to comparing our ACR with other methods, we also compare a to highly simplified version of it which we call sparsifying filter bank (SFB). It is trained analogously to the ACR, see Equation~\ref{cvx_r_def1}, but the architecture of $\R'$ is simply given by
%%%%
\begin{table}[t]
\centering
\begin{tabular}{c| c| c| c}
\textbf{methods} &  \textbf{optimizer}  & \textbf{$\alpha$} & \textbf{\# iterations} \\[1.2ex]
\hline
AR & gradient-descent, $\eta=0.5$ & 0.1 & 600  \\[1.2ex]
\hline
ACR & gradient-descent, $\eta=0.8$ & 0.05 & 400  \\[1.2ex]
\hline
\end{tabular}
%} 
\vspace{0.1in}
\caption{\small{The optimizer setting for the variational reconstruction problems corresponding to AR and ACR.}}
\label{var_recon_setting_table} % is used to refer this table in the text 
\end{table}
%%%%
%%%%%%%%%%%%%
\begin{table}[t]
\centering
\begin{tabular}{l| c| c| r|r}
\textbf{methods} &  \textbf{PSNR (dB)}  & \textbf{SSIM} & \textbf{\# param.}& \textbf{time (ms)} \\[1.2ex]
\hline
FBP & 21.2866  & 0.2043 & 1 & 14.0 \\[1.2ex]

TV & 30.3476  & 0.8110 & 1 & 21\,315.0\\[1.2ex]

LPD &  35.1519 & 0.9073  & 1\,138\,720& 184.4   \\[1.2ex] %34.3590, 0.8889 with 3x3 filters, #params 411\,680
FBP + U-Net & 31.8008  &  0.7585 & 7\,215\,233 & 18.6\\[1.2ex]

AR & 33.6207  & 0.8750 & 19\,347\,890 & 41\,058.0\\[1.2ex]

ACR (no SFB) &  31.2822 & 0.8468 & 590\,928 & 242\,414.3\\[1.2ex]

ACR (with SFB) &  31.4555 & 0.8644 & 606\,609 & 245\,366.5\\[1.2ex]

SFB + $\rho_0\left\|\x\right\|_2^2$ &  26.8001 & 0.5678 & 15\,681 & 86\,864.3\\[1.2ex]

\hline
\end{tabular}
%} 
\vspace{0.1in}
\caption{\small{Average PSNR, SSIM, and reconstruction time over test data for different reconstruction methods for sparse-view CT. The number of free parameters in different reconstruction techniques, which is indicative of the complexity of the model, is also indicated.}}
\label{sparse_ct_table} % is used to refer this table in the text 
\end{table}
%%%%%%%%%%%%%%%%%%%%%%%%%%
\subsection{Reconstruction in sparse-view CT}
\label{sv_ct_sec}
The sparse-view projection data is simulated in ODL using a parallel-beam acquisition geometry with 200 angles and 400 rays/angle. White Gaussian noise with $\sigma=2.0$ is added to the projection data to simulate noisy measurement. The unregularized reconstructions $\Op{A}^{\dagger}\y_i$ are taken as the output of the classical filtered back-projection method. The proposed ACR approach is compared with two model-based techniques, namely (i) FBP and (ii) TV regularization; and three data-driven methods, namely (i) the learned primal-dual (LPD) method proposed in \cite{lpd_tmi}, (ii) the adversarial regularization (AR) approach introduced in \cite{ar_nips}, and the U-Net-based post-processing of FBP reconstruction \cite{postprocessing_cnn}. The LPD method is trained on pairs of target image and projection data, while AR and the proposed ACR method are trained on ground-truth images and the corresponding FBP reconstructions. As explained before (Sec. \ref{learning_protocol_sec}), using paired data for training ACR leads to a stable decay of the training loss as compared to the case where unpaired examples are used.\\
%\noindent The TV solution is computed by employing the \textit{alternating directions method of multipliers} (ADMM)-based solver in ODL with penalty parameter $\alpha_{\text{tv}}=10.0$, which was selected empirically to maximize the PSNR for TV reconstruction. The ADMM algorithm was run for 2000 iterations starting with the FBP reconstruction as the initial estimate.
For LPD and AR, we develop \textit{PyTorch}~\cite{paszke2017automatic} implementations\footnote{The ACR implementation is available upon request.} based on their publicly available \textit{TensorFlow} codes\footnote{LPD: \url{https://github.com/adler-j/learned_primal_dual}, AR: \url{https://github.com/lunz-s/DeepAdverserialRegulariser}}. The LPD architecture was adapted to sparse-view projections by increasing the filter-size to $5\times 5$ and stacking 20 primal-dual layers (instead of $3\times 3$ filters and 10 layers, respectively, originally proposed in \cite{lpd_tmi} for densely sampled projections) to increase its overall receptive field. This gives LPD a fair chance to reconstruct features that extend over a larger region in the image, in the face of sparse angular sampling. The optimizer setting used for AR and ACR while solving the variational problem is presented in Table \ref{var_recon_setting_table}. The hyper-parameters in Table \ref{var_recon_setting_table} are chosen empirically to maximize the reconstruction PSNR for our experiments.\\
\indent The ACR architecture is constructed as described in Sec.~\ref{icnn_sec} with $L=10$ layers. $\B_i$ and $\W_i$ are convolutional layers with $5\times 5$ kernels applied with a stride of 1 and consisting of 48 output channels. The layers $\B_i$ are restricted to have non-negative weights to preserve convexity, while no such restriction is needed on $\W_i$'s. The activation at all layers except the final layer is chosen as the leaky-ReLU function with a negative slope of 0.2. The activation of the final layer is the identity function, and we apply a global average pooling on the output feature map to convert it to a scalar. The SFB term is composed of 10 2D convolution layers, each consisting of $7\times 7$ filters with 32 output channels. The \textit{Adam} optimizer~\cite{kingma2014adam} is used for training the network, with a learning rate of $2\times 10^{-5}$ and $(\beta_1,\beta_2)=(0.5, 0.99)$. The gradient penalty term in \eqref{grad_penalty} is chosen to be 5.0. The penalty parameter $\rho_0$ corresponding to the squared-$\ell_2$ term in ACR is initialized at $\rho_0=\log\left(1+\exp(-9.0)\right)$ and then learned from training data. ACR and AR were trained for 5 and 10 epochs, respectively. We found that the performance of AR during reconstruction tends to deteriorate if the network is over-trained, while for ACR, we did not find any noticeable improvement or deterioration due to over-training, suggesting that AR is more susceptible to overfitting as compared to ACR.\\
\indent The performance of ACR and the competing model- and data-based techniques, averaged over the test data, is reported in Table~\ref{sparse_ct_table}. Representative reconstructed images for different methods are shown in Fig.~\ref{ct_image_figure}. The average PSNR and SSIM of the reconstructed images indicate that the proposed ACR method leads to better reconstruction than  TV (approximately 1 dB higher PSNR), which is by far the most popular analytical convex regularizer with well-studied convergence properties. Including the SFB term in conjunction with $\R'$ leads to slight improvement in performance, although the SFB term alone turns out to be significantly inferior to TV. This indicates the need for a convex regularizer, like ours, that goes beyond linear sparsity-based convex prior models and learns more intricate structures specific to the application at hand. 

The convex regularizers arising from linear sparsifying dictionaries (somewhat akin to the SFB-like priors in spirit) lead to a convex variational problem that can be formulated as a quadratic program (for the squared-$\ell_2$ data-fidelity measure) and solved efficiently. Nevertheless, convex priors emerging from linear sparsity might be limited in their ability to capture reasonably complicated signal priors. On the other extreme of the spectrum, a convex prior built by utilizing a parametric ICNN is substantially more expressive, although the corresponding variational objectives are relatively harder to minimize as compared to the linear sparsity priors. In particular, it was shown in \cite[Theorem 1]{icnn_opt_transport} that ICNNs with non-negative weights and ReLU activations are universal approximators in the class of Lipschitz convex functions over a compact domain. Therefore, the proposed ACR regularizers subsume the class of $\ell_1$ sparsity based convex priors and are capable of representing more complicated image priors. The performance metrics reported in Table \ref{sparse_ct_table} indicate the performance-vs.-complexity trade-off in a quantitative manner. While the proposed ACR prior outperforms the SFB prior alone by a significant margin, it has approximately 40 times as many parameters as compared to SFB and the reconstruction time is nearly three-fold higher. Thus, the proposed convex prior results in a variational problem that entails somewhat higher complexity in comparison with TV or the SFB prior, but one can still use a provably convergent sub-gradient method for optimization (as shown in Lemma \ref{subgrad_convergence_lemma}). Unsurprisingly, the end-to-end trained networks on supervised data turn out to be more efficient in terms of reconstruction time as opposed to the methods that require solving a variational reconstruction problem in high dimension.

For the sparse-view CT experiment, the classical AR approach yields reconstruction that is superior to ACR, although our experiments in the sequel reveal that this improvement is not consistent across applications. As seen from the average PSNR and SSIM on test images reported in Table \ref{sparse_ct_table}, the performance of ACR is comparable (slightly worse in terms of PSNR, but better in SSIM) with a U-Net-based post-processing network \cite{postprocessing_cnn} trained with FBPs as the input and the ground-truth images as the target. This indicates that the proposed ACR approach, along with offering the flexibility of unsupervised training, performs approximately on par with a fully data-driven supervised method. As one would expect, the LPD framework trained using paired examples performs the best among all the methods we compare, since it incorporates the acquisition physics into the architecture.
% Although the SFB model encompasses the TV regularizer, we believe that the unsupervised training strategy on limited amount of data is the reason behind its worse performance as compared to TV.
%%%%

% \textcolor{red}{Ozan: Note that one should perhaps slightly adapt the LPD architecture to the sparse view setting to optimise its performance. Our current set-up is parallel beam CT with 200 angles and 400 rays/angle. The original LPD architecture was designed for fan-beam CT with 1000 source positions and 1000 rays/source position. Given the number of layers and size of the convolution kernels used in the CNNs encoding the primal and dual maps in that architecture, the resulting DNN will essentially have a field of view corresponding to 30 pixels. This could be too little for the sparse view setting we have here, i.e., it would make sense to increase the field of view for the CNNs in LPD to reflect the sparse view and/or limited angle settings.}
%%%%
\begin{figure*}[h]
	\centering
	\subfigure[\small{ground-truth}]{
		\includegraphics[width=2.0in]{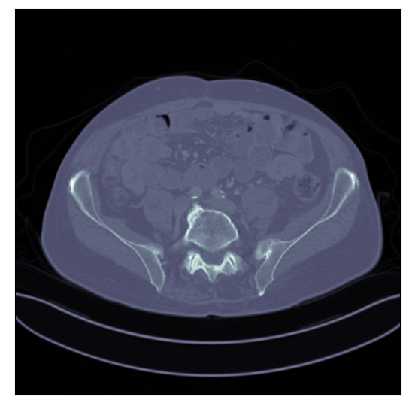}}
	\subfigure[\small{FBP: 21.6122 dB, 0.1696}]{
		\includegraphics[height=2.0in]{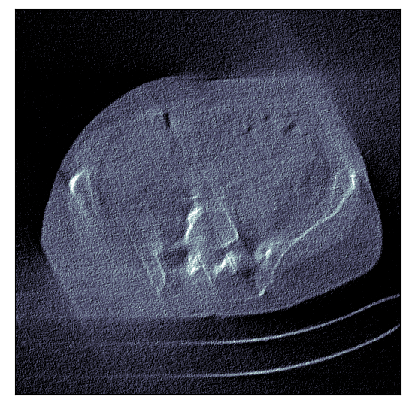}}
	\subfigure[\small{TV: 25.7417 dB, 0.7968}]{
		\includegraphics[width=2.00in]{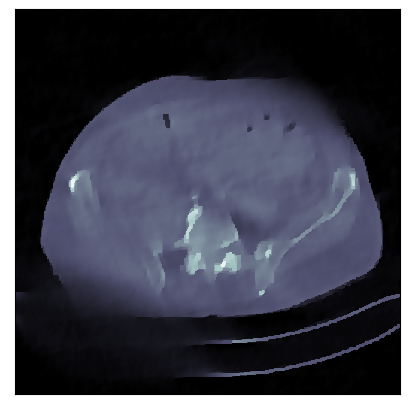}}\\
	\subfigure[\small{LPD: 29.5083 dB, 0.8466}]{
		\includegraphics[width=2.0in]{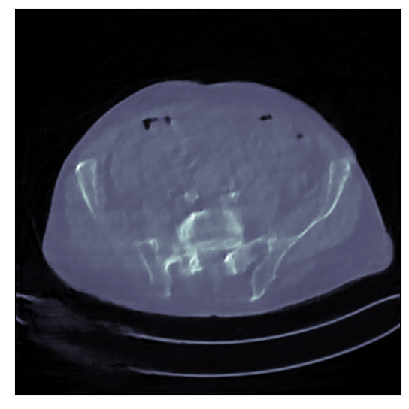}}
	\subfigure[\small{AR: 26.8305 dB, 0.7137}]{
		\includegraphics[width=2.0in]{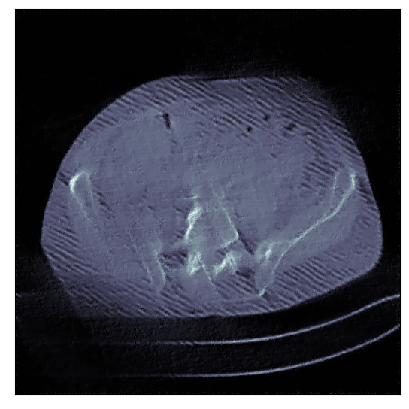}}
	\subfigure[\small{ACR: 27.9763 dB, 0.8428}]{
		\includegraphics[width=2.0in]{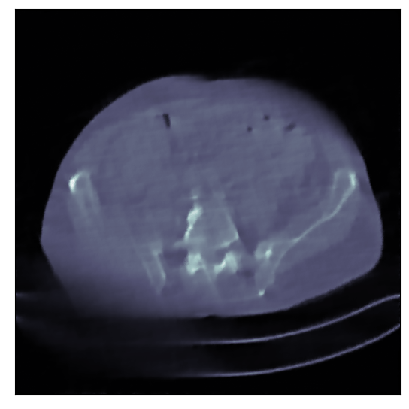}}
	\caption{\small{Reconstructed images obtained using different methods, along with the associated PSNR and SSIM, for limited-angle CT. In this case, ACR outperforms TV and AR in terms of reconstruction quality.}}
	\label{lim_ct_image_figure}
\end{figure*}
%%%%%%%%%%%%%%%%%%%%
\subsection{Reconstruction in limited-angle CT}
CT reconstruction from limited-angle projection data, where no measurement is available in a particular angular region, is considered to be a particularly challenging ill-posed inverse problem. Limited-angle projections arise primarily because of limited scan-time or restricted scanner movement in certain applications. Due to the lack of projection data in an angular region, the reconstruction performance depends critically on the image prior. Akin to the sparse-view CT experiment, the projection data for the limited-angle CT experiment is simulated using ODL with a parallel-beam acquisition geometry. The projection data is corrupted using white Gaussian noise with $\sigma = 3.2$ and reconstructed with 350 angles, 700 rays/angle, with a missing angular wedge of 60$^{\circ}$. In this experiment, we compare our method with two model-based (FBP and TV) and three data-driven approaches (LPD, AR, and post-processing U-Net) as before. We found that unlike sparse-view CT, the LPD architecture had to be modified for limited-angle CT by reducing the overall number of layers and the number of feature channels in the convolutional layers, and by adding batch normalization after each convolutional layer in the primal and dual spaces. These modifications resulted in an LPD network with fewer learnable parameters and helped avoid overfitting while training on considerably ill-posed measurements. The LPD model was also trained with an increased batch size of 15 to ensure stability in terms of convergence during training.\\
% The U-net denoiser is trained in a supervised manner, with the FBP reconstruction as the input and the true image as the target, similar to the post-processing approach followed in \cite{postprocessing_cnn}.We found that unlike sparse-view CT, such a post-processing network works better than the LPD method for limited-angle CT.\\  
\indent The ACR architecture is constructed with $L=5$ layers, and $\B_i$ and $\W_i$ are all chosen as convolutional layers with $5\times 5$ kernels and 16 channels. Reducing the total number of layers and the number of feature channels in each layer of ACR as compared to the sparse-view reconstruction experiment helped avoid overfitting in the face of severely ill-posed data. The \textit{RMSprop} optimizer (following the recommendation in \cite{ar_nips}, since \textit{Adam} gave no improvement in convergence during training, unlike sparse-view CT) with a learning rate of $5\times 10^{-5}$ is used for training. The gradient penalty term in \eqref{grad_penalty} is chosen to be 10.0 and the penalty parameter $\rho_0$ is initialized the same way as in the sparse-view CT experiment.
The convolutional layers for the SFB penalty is taken as $5\times 5$ filters with 32 output channels.\\
\indent The reconstruction is performed by solving the variational problem via gradient-descent for 2000 iterations with a step size of $10^{-5}$. We observed that the reconstruction performance of AR can deteriorate if an early stopping is not applied. So, for a fair comparison, we report the highest PSNR achieved by AR during reconstruction. Such an early stopping was not needed for ACR. This phenomenon indicates a notable algorithmic advantage of a convex regularizer over a non-convex one.\\  
% For both ACR and AR, The step with the highest PSNR value is chosen as the reconstruction. In the case of ACR, this corresponds to the last iteration, whereas when using AR, early stopping must be utilised in order to achieve best performance. 
\indent The average PSNR and SSIM are reported in Table~\ref{limited_ct_table} for various competing methods. To facilitate visual comparison, an example of the reconstruction quality for a representative test image is shown in Fig.~\ref{lim_ct_image_figure}. From the average PSNR and SSIM reported in Table~\ref{limited_ct_table} we can see that ACR outperforms both model-based approaches and AR. As opposed to sparse-view CT, AR under-performs significantly in the limited-angle setting and produces streak-like artifacts that cover the whole image and are particularly concentrated near the missing angular region. The reason behind this phenomenon is that AR favors oscillations in the direction of the blurring artifacts in the FBP reconstruction. Since the ground-truth images exhibit more high frequency components pointing in this spatial direction as compared to FBP reconstructions, this becomes a natural feature to use for telling apart the two distributions.
% parallel to the missing wedge, as the FBP reconstructions are particularly blurry in those regions.
Furthermore, the data fidelity term does not contain any information on these regions, and, therefore, it cannot steer AR away from over-regularizing. As a result, AR attempts to maximize the variation 
%  orthogonal to 
in the main blurring direction. This behavior does not arise in case of ACR, leading to better reconstructions. We hypothesize that this is due to the fact that a functional favoring high oscillation in a spatial direction is typically non-convex. In particular, if we consider the total variation in this spatial direction as a measure of oscillation, the negative total variation functional would emerge as an optimal functional to maximally favor high variation in the given spatial direction. This functional is concave, explaining why the convex structure of ACR prevents it from approximating this functional and hence from overly introducing a high amount of oscillation in the reconstruction.
% owing to convexity, which imposes restrictions on the model preventing such streaky artifacts from appearing. 
% One of the reasons for this behaviour is because of a rather small training dataset, which leads to the AR model overfitting to the training data. 
% This problem does not arise in case of ACR, which, thanks to lower model capacity, can successfully evade the peril of overfitting when the amount of training data is limited.
Thus, contrary to the conventional intuition, the restricted expressive power of a convex regularizer turns out to be advantageous, especially in a limited data scenario with a highly ill-conditioned forward operator. Similar to sparse-view CT, including the SFB term in ACR leads to a slightly better reconstruction performance and the SFB term as the standalone convex regularize performs worse than both TV and ACR with and without the SFB term. For limited-angle projection data, we found that LPD and the U-Net post-processor perform almost on par with each other. Since there is no data available for a  considerable angular region, including the acquisition physics into the network does not bring in any significant advantage.    
%%%%%%%%
\begin{table}[t]
\centering
\begin{tabular}{l| c| c| r}
%\hline
\textbf{methods} &  \textbf{PSNR (dB)}  & \textbf{SSIM} & \textbf{\# parameters} \\[1.2ex]
\hline
FBP & 17.1949  & 0.1852 & 1 \\[1.2ex]

TV & 25.6778  & 0.7934 & 1 \\[1.2ex]

LPD &  28.9480 & 0.8394 & 127\,370 \\[1.2ex]

FBP + U-Net &  29.1103 & 0.8067 & 14\,787\,777 \\[1.2ex]

AR & 23.6475 & 0.6257 & 133\,792 \\[1.2ex]

ACR (no SFB) &  26.4459 & 0.8184 & 34\,897 \\[1.2ex]

ACR (with SFB) &  26.7767 & 0.8266 & 42\,898\\[1.2ex]

SFB + $\rho_0\left\|\x\right\|_2^2$ &  19.1876 & 0.1436 & 8\,001 \\[1.2ex]
\hline
\end{tabular}
%} 
\vspace{0.1in}
\caption{\small{Average PSNR and SSIM over test data for different reconstruction methods for limited-angle CT.}}
\label{limited_ct_table} % is used to refer this table in the text 
\end{table}
%%%%%%%%
\begin{table}[t]
\centering
\begin{tabular}{c| c| c| c | c | c}
\textbf{methods} &  TV  & SFB & ACR & AR & AR2 \\[1.2ex]
\hline
\textbf{\# parameters} & 1 & 4\,706 & 142\,594 & 2\,562\,242 & 142\,594 \\
\end{tabular}
\vspace{0.1in}
\caption{Number of parameters in each of the deblurring methods.}
\label{table:number_of_parameters_deblurring}
\end{table}
%%%%%%%%
\begin{table}
    \centering
    \begin{tabular}{l | l r r r r r}
        \multicolumn{1}{c|}{} 
        &\multicolumn{1}{l}{\textbf{stats}} 
        & \multicolumn{1}{c}{TV} 
        & \multicolumn{1}{c}{SFB} 
        & \multicolumn{1}{c}{ACR} 
        & \multicolumn{1}{c}{AR} 
        & \multicolumn{1}{c}{AR2} \\
        \hline
\multirow{3}{*}{\textbf{PSNR (dB)}}
        &mean & $25.50$&   $26.05$&   $26.55$&   $26.35$&   $26.57$\\
        &median& $25.01$& $25.82$&  $26.12$&  $25.99$&  $26.21$\\
        & std. dev.& $2.08$&  $1.71$&  $2.05$&  $1.98$&  $1.88$\\
        \hline
        \multirow{3}{*}{\textbf{SSIM}}
        &mean& $0.80$&  $0.81$& $0.83$&   $0.82$&  $0.83$\\
        &median & $0.80$&  $0.81$&  $0.83$&  $0.82$&  $0.83$\\
        & std. dev.& $0.05$&  $0.04$&  $0.03$&  $0.03$&  $0.03$\\

\end{tabular}
    \\[2ex]
    \caption{Average test performance for deblurring on the STL-10 dataset for different methods in terms of PSNR and SSIM.}
    \label{table:deblurring_results_new}
\end{table}
%%%%%%%%
\begin{figure*}
    \centering
    \setlength\tabcolsep{1.pt}
    \begin{tabular}{r r r r r r}
        \multicolumn{1}{c}{ground-truth} 
        & \multicolumn{1}{c}{TV} 
        & \multicolumn{1}{c}{SFB} 
        & \multicolumn{1}{c}{ACR} 
        & \multicolumn{1}{c}{AR} 
        & \multicolumn{1}{c}{AR2} \\
        \\[-2ex]

        \includegraphics[trim={3.25cm 0.48cm 3.25cm 1.3cm},clip, width=.16\textwidth]{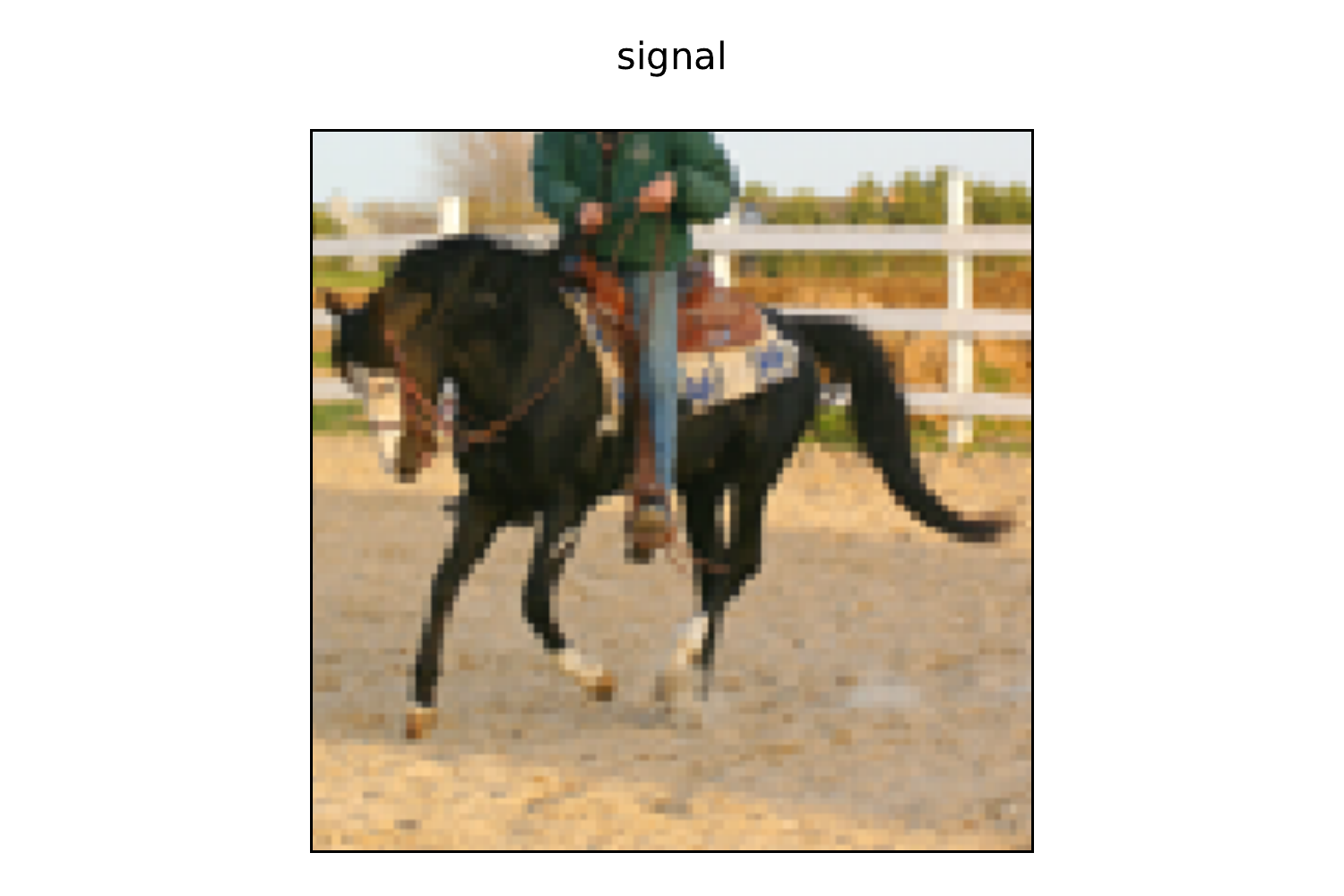} &        \includegraphics[trim={3.25cm 0.48cm 3.25cm 1.3cm},clip, width=.16\textwidth]{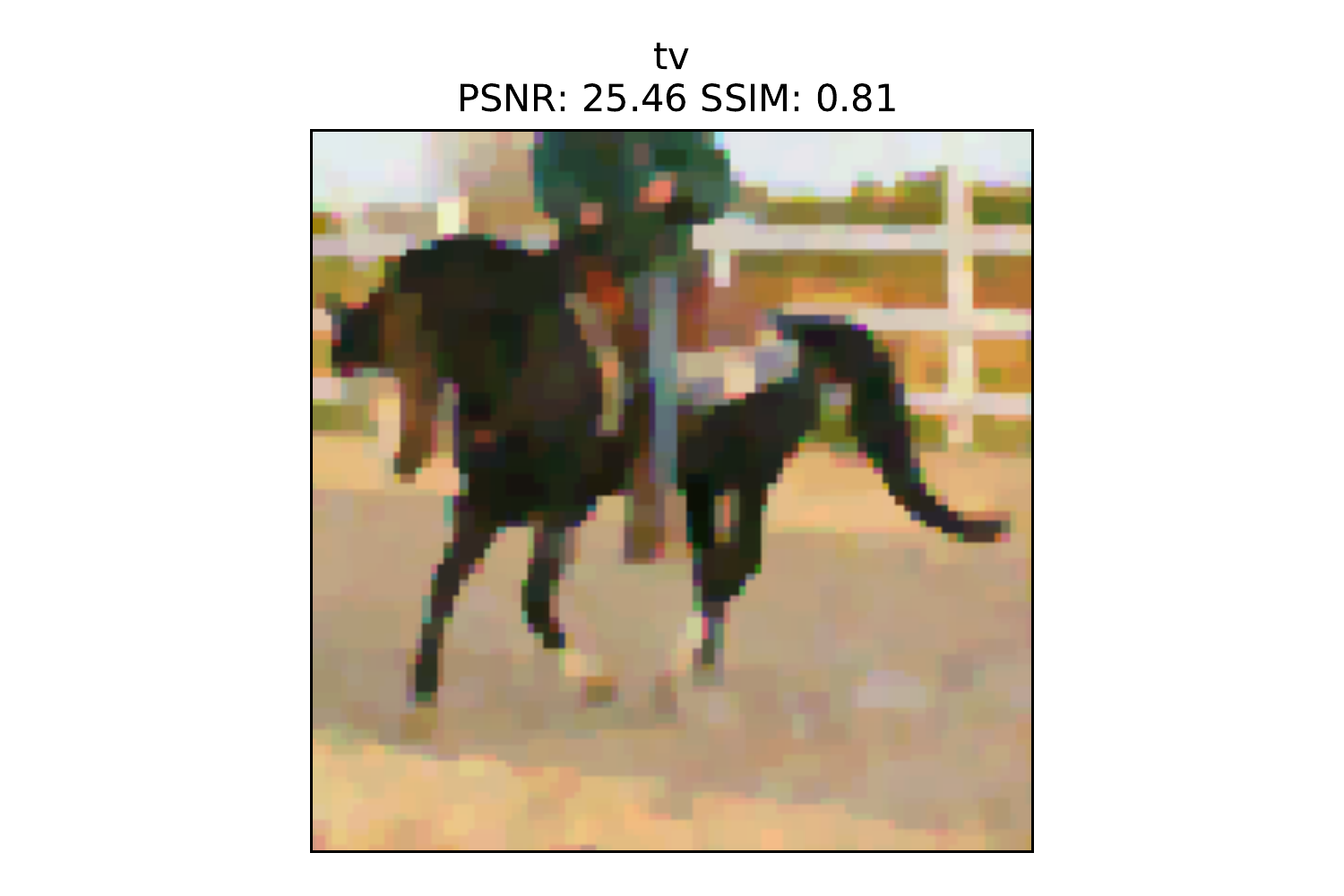} &        \includegraphics[trim={3.25cm 0.48cm 3.25cm 1.3cm},clip, width=.16\textwidth]{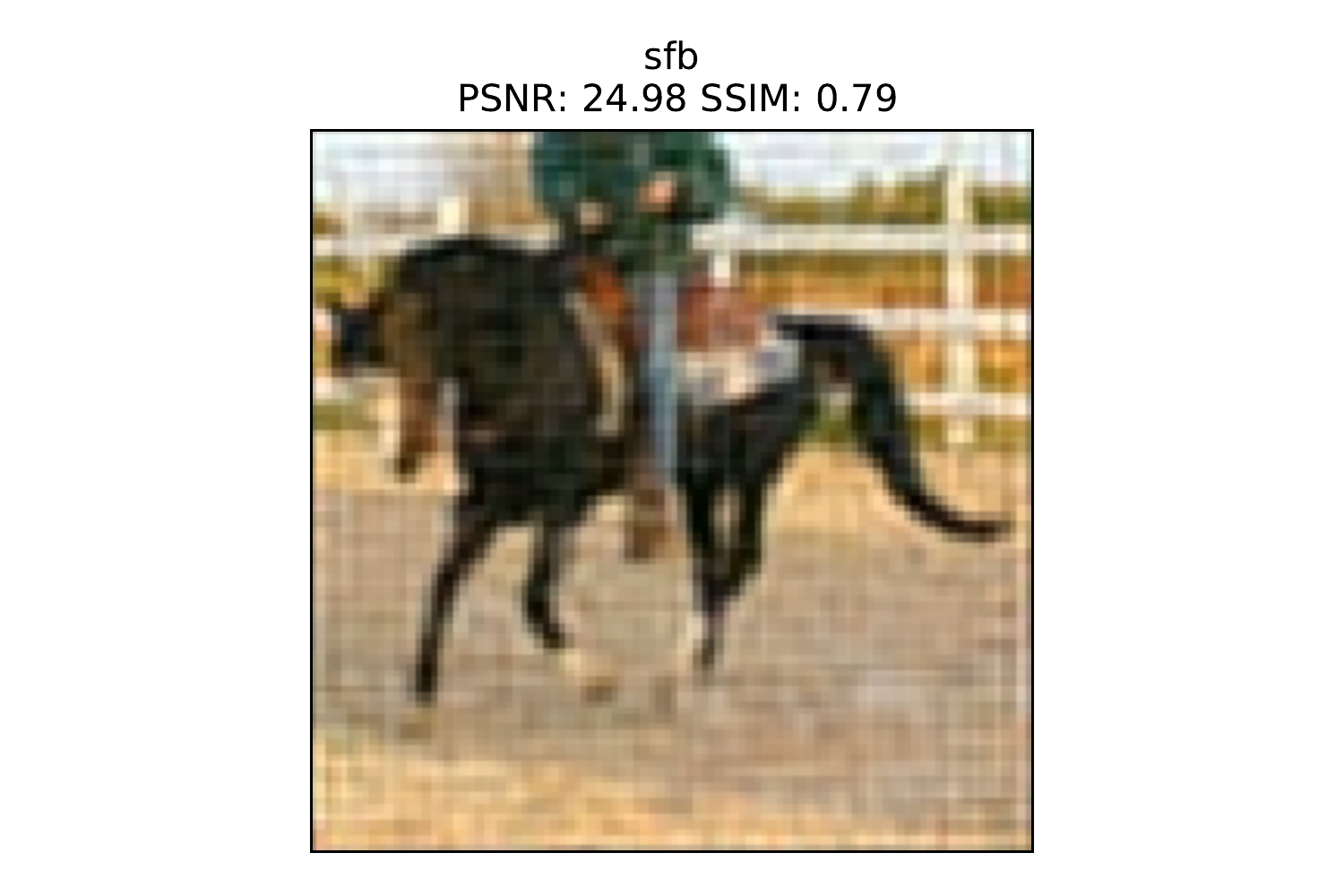} &        \includegraphics[trim={3.25cm 0.48cm 3.25cm 1.3cm},clip, width=.16\textwidth]{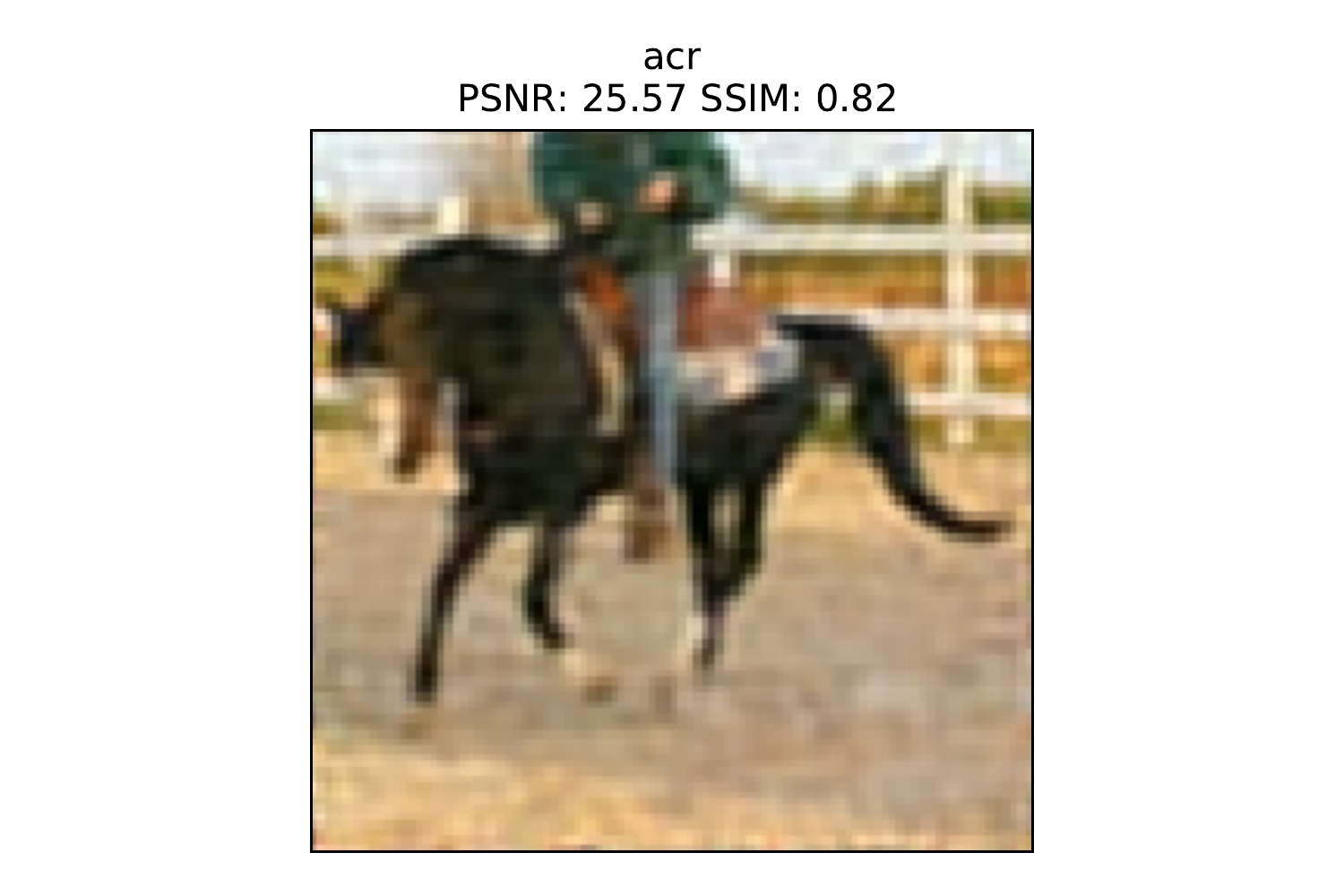} &        \includegraphics[trim={3.25cm 0.48cm 3.25cm 1.3cm},clip, width=.16\textwidth]{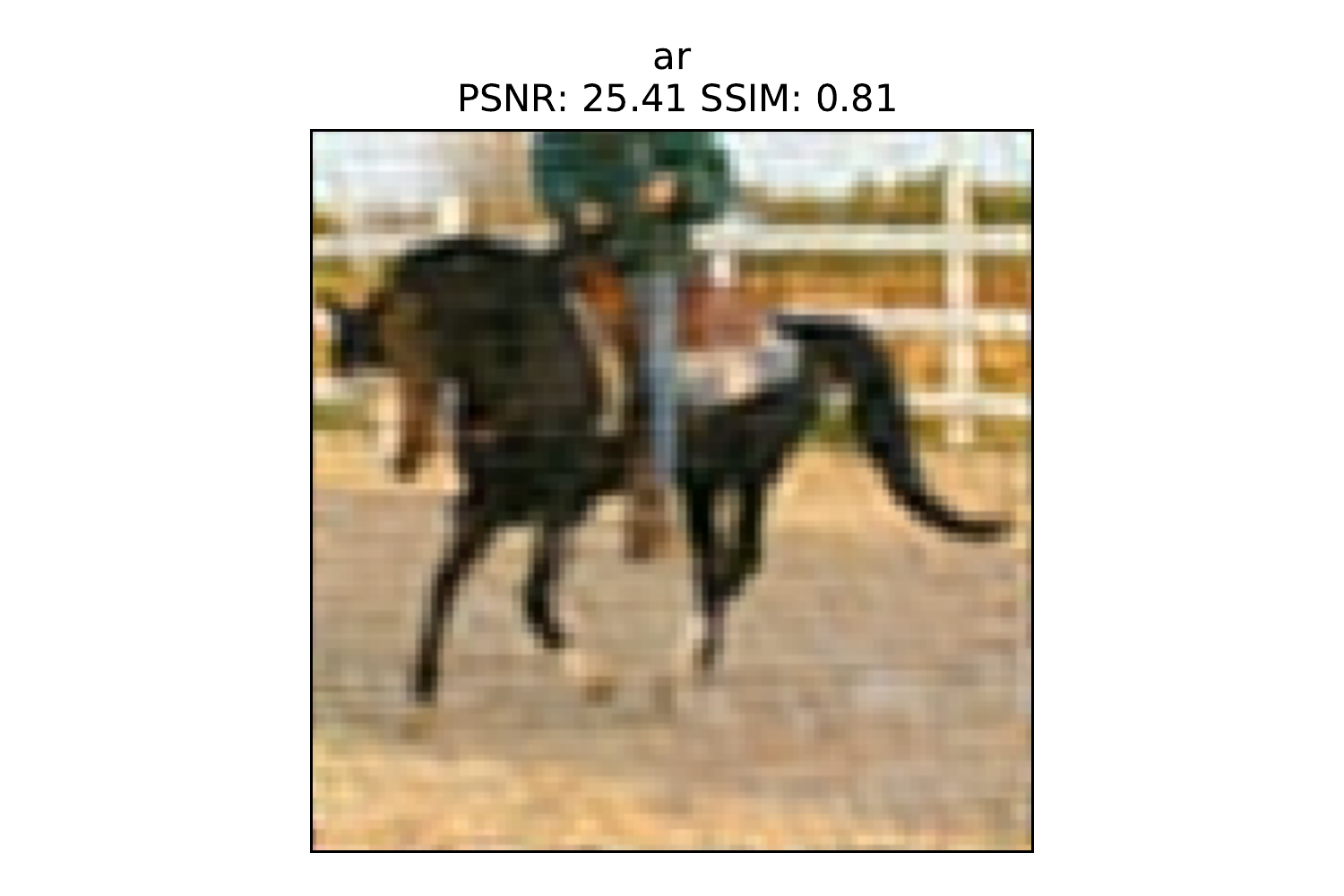} &        \includegraphics[trim={3.25cm 0.48cm 3.25cm 1.3cm},clip, width=.16\textwidth]{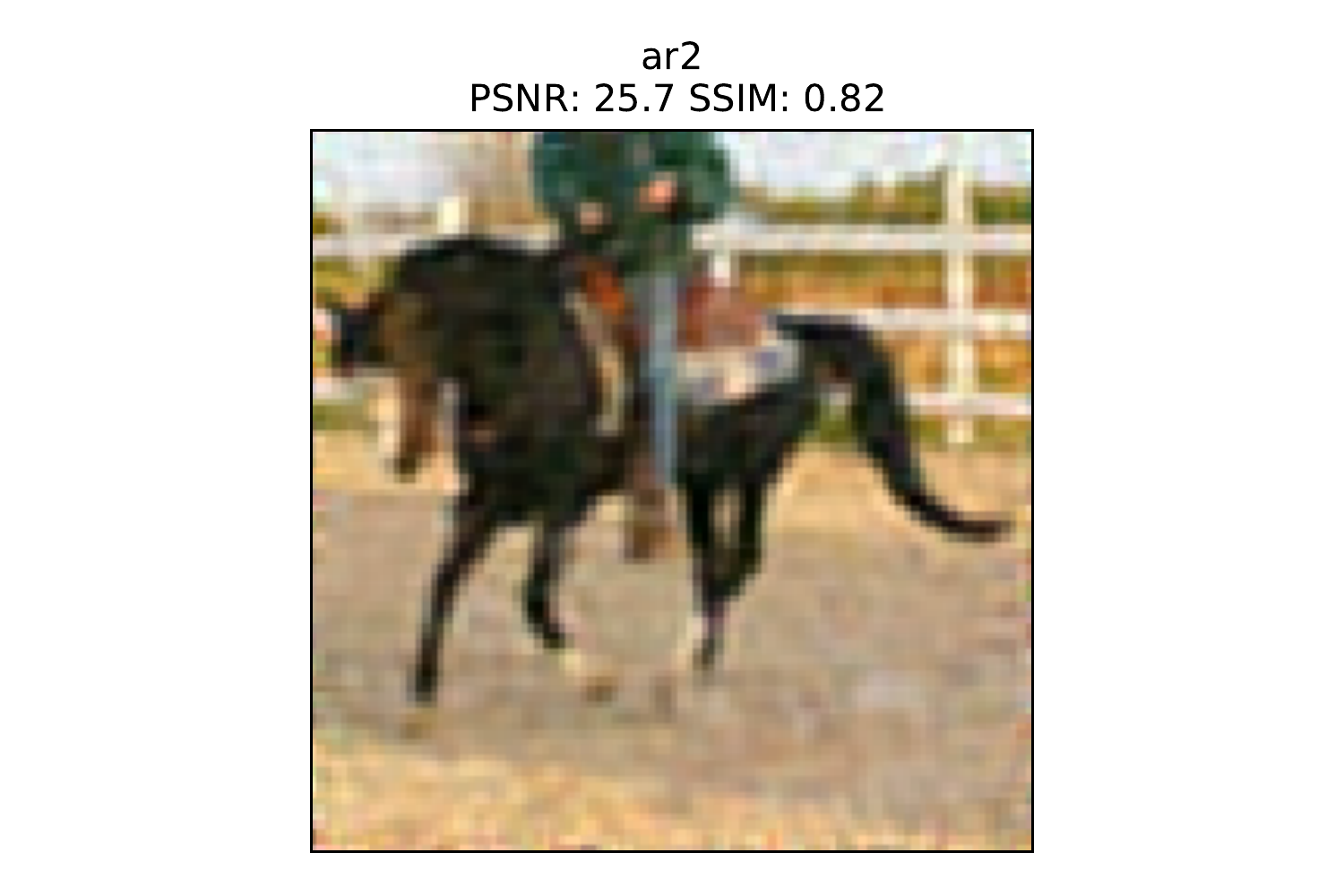} \\

        \includegraphics[trim={3.25cm 0.48cm 3.25cm 1.3cm},clip, width=.16\textwidth]{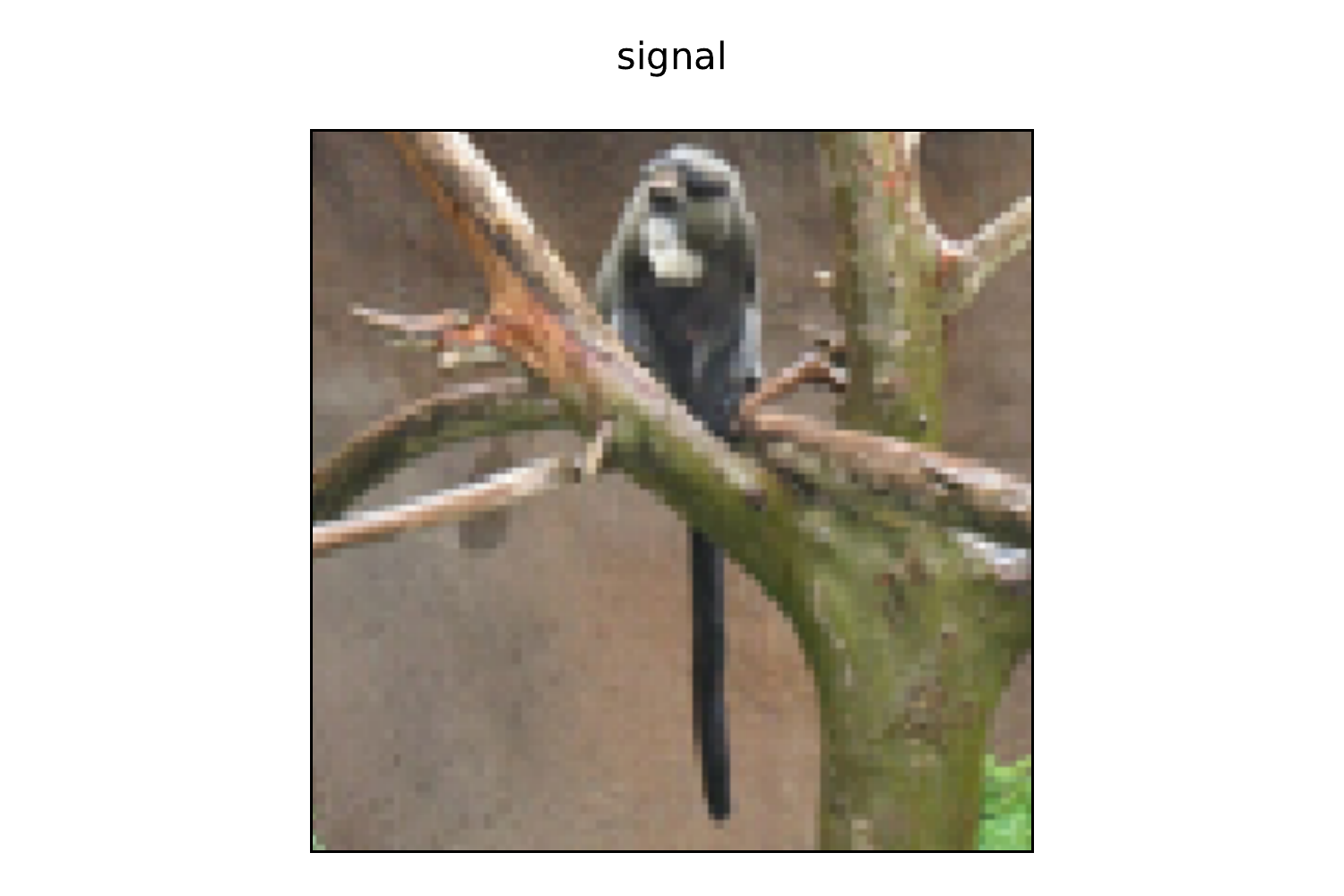} &        \includegraphics[trim={3.25cm 0.48cm 3.25cm 1.3cm},clip, width=.16\textwidth]{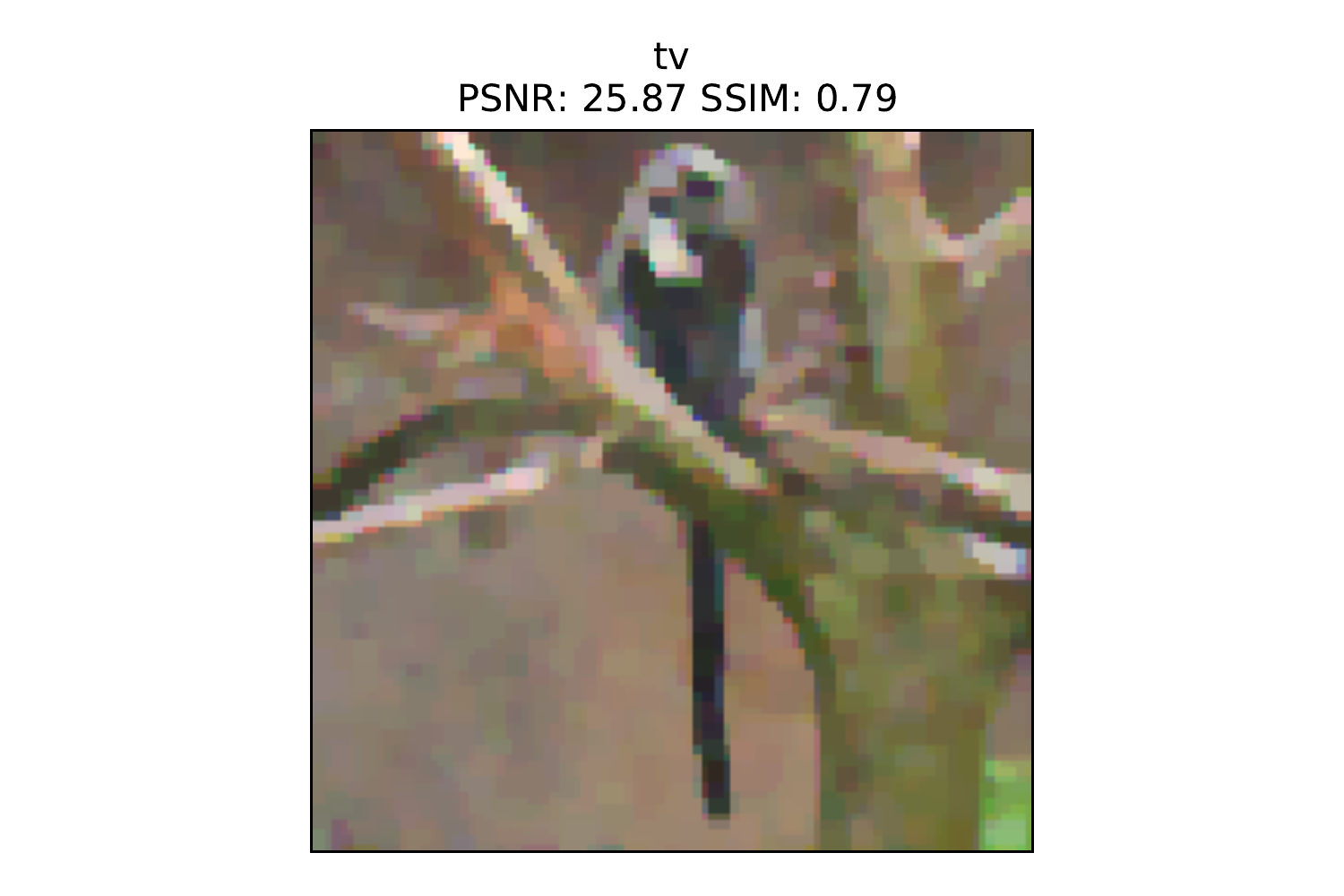} &        \includegraphics[trim={3.25cm 0.48cm 3.25cm 1.3cm},clip, width=.16\textwidth]{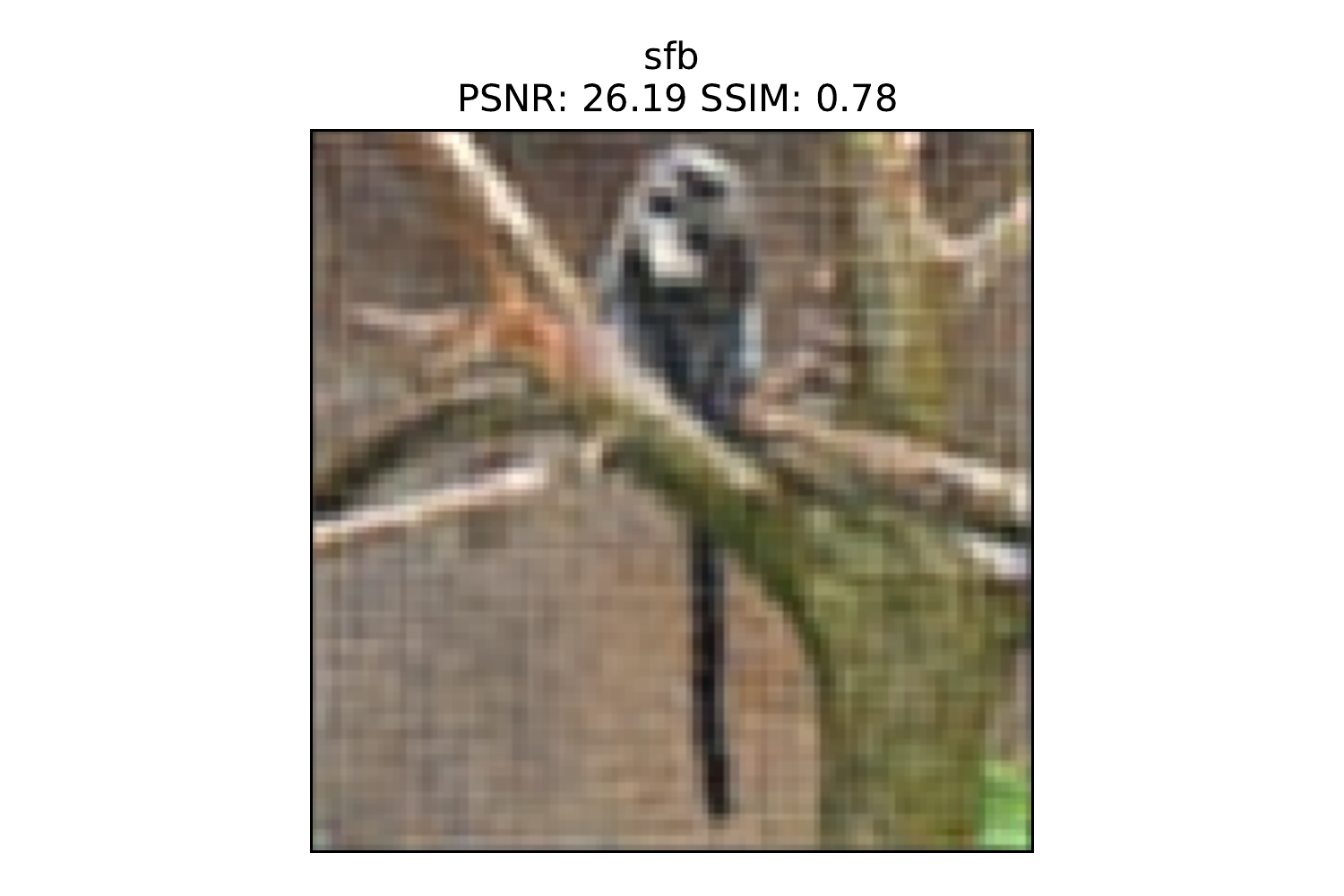} &        \includegraphics[trim={3.25cm 0.48cm 3.25cm 1.3cm},clip, width=.16\textwidth]{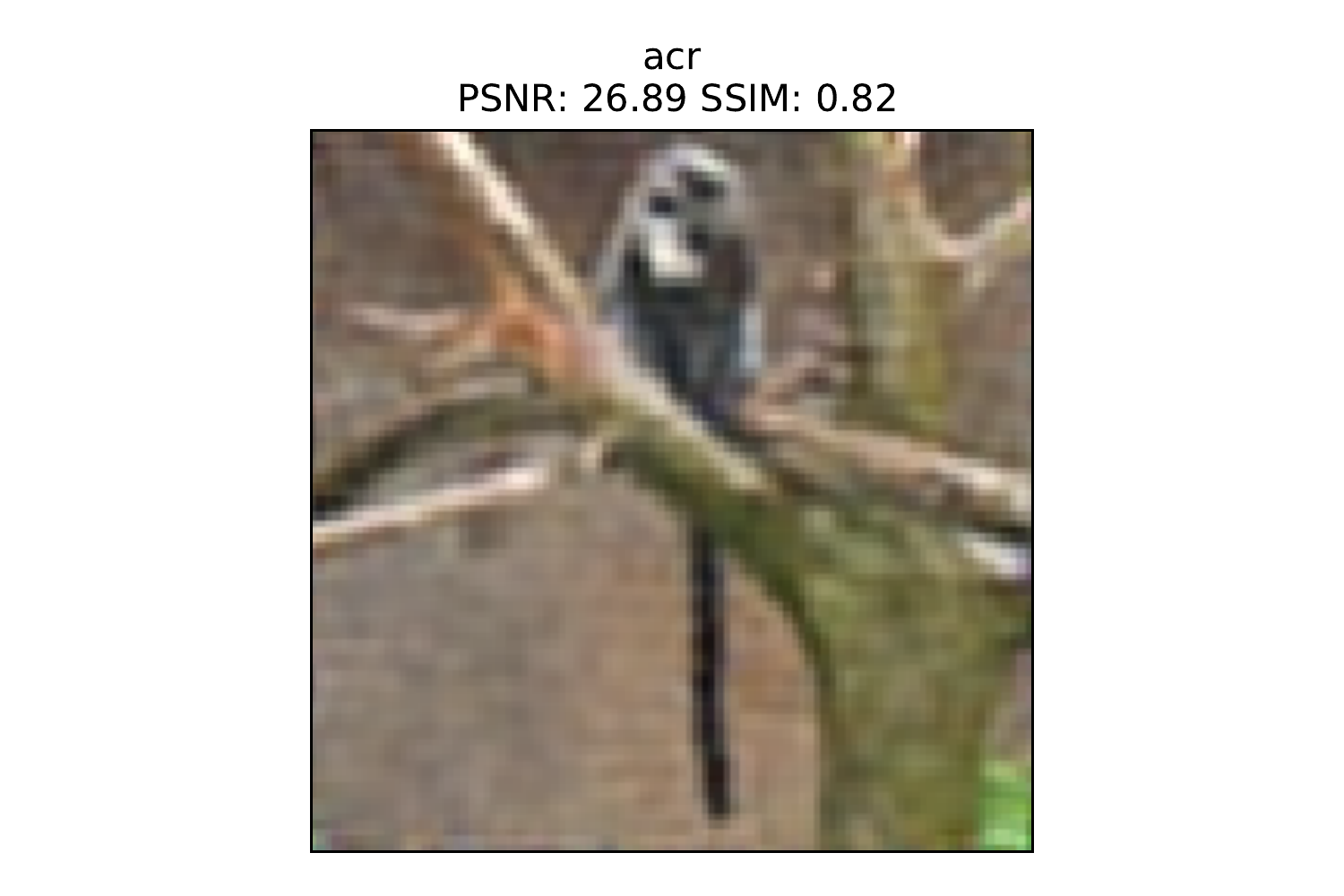} &        \includegraphics[trim={3.25cm 0.48cm 3.25cm 1.3cm},clip, width=.16\textwidth]{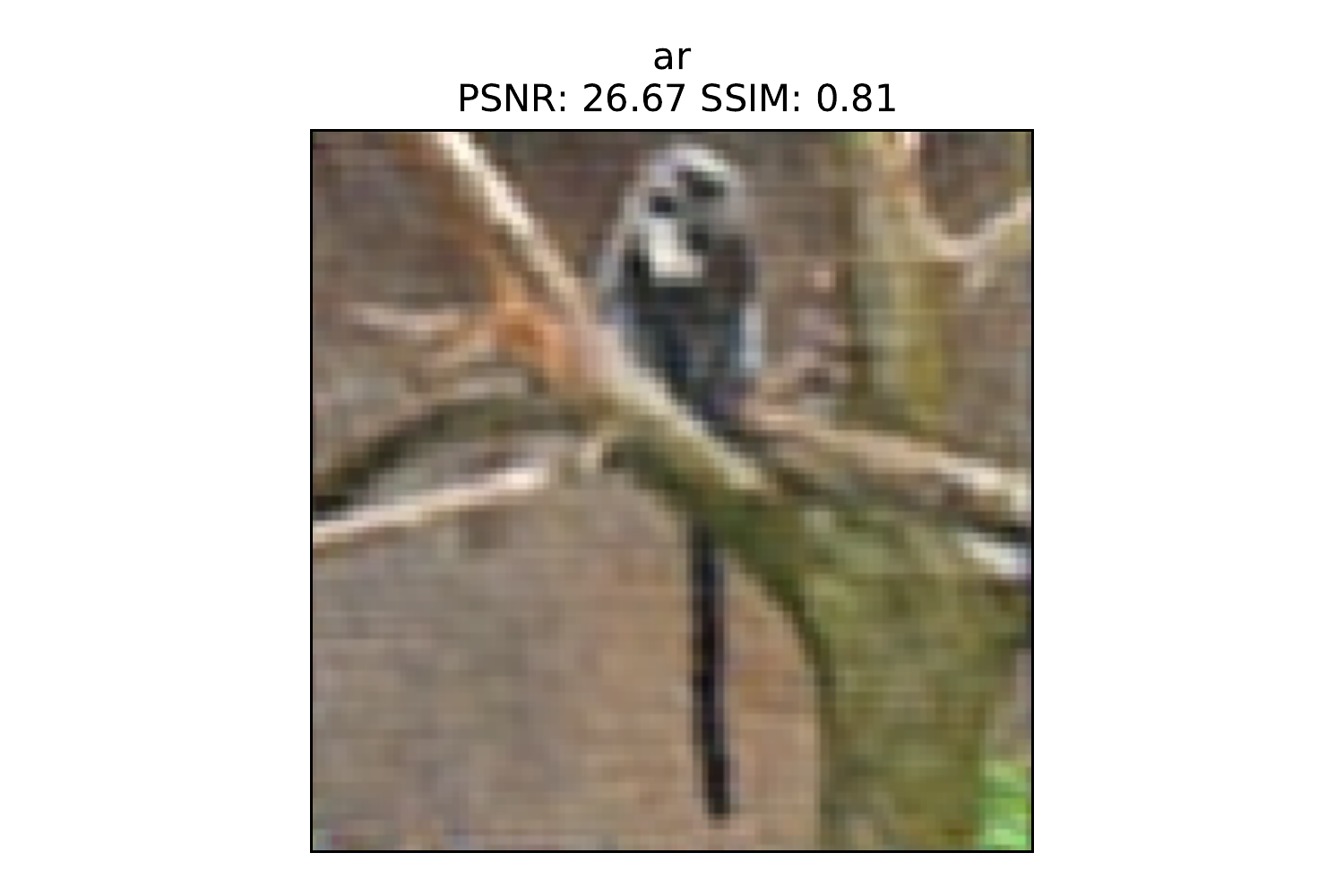} &        \includegraphics[trim={3.25cm 0.48cm 3.25cm 1.3cm},clip, width=.16\textwidth]{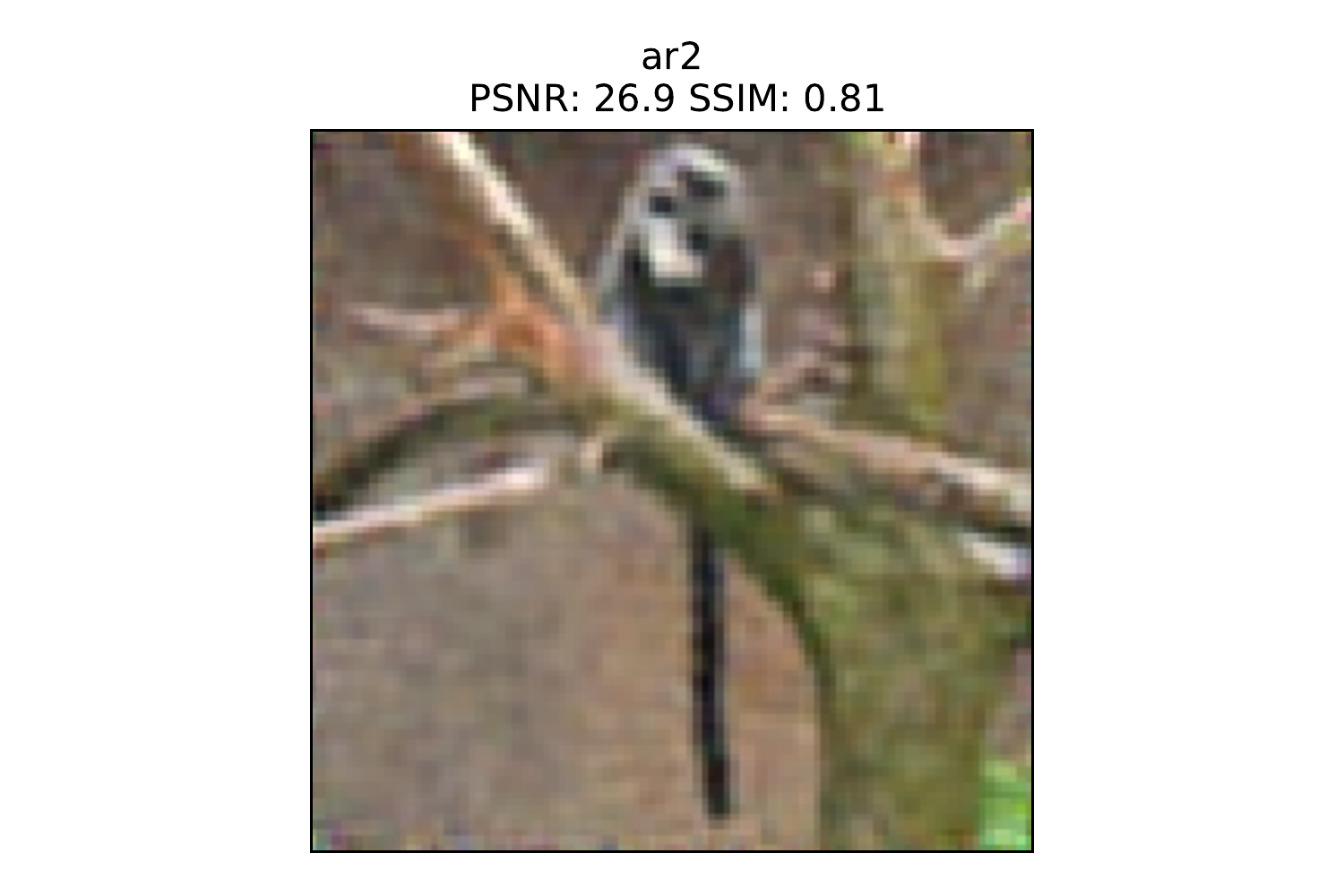} \\

        \includegraphics[trim={3.25cm 0.48cm 3.25cm 1.3cm},clip, width=.16\textwidth]{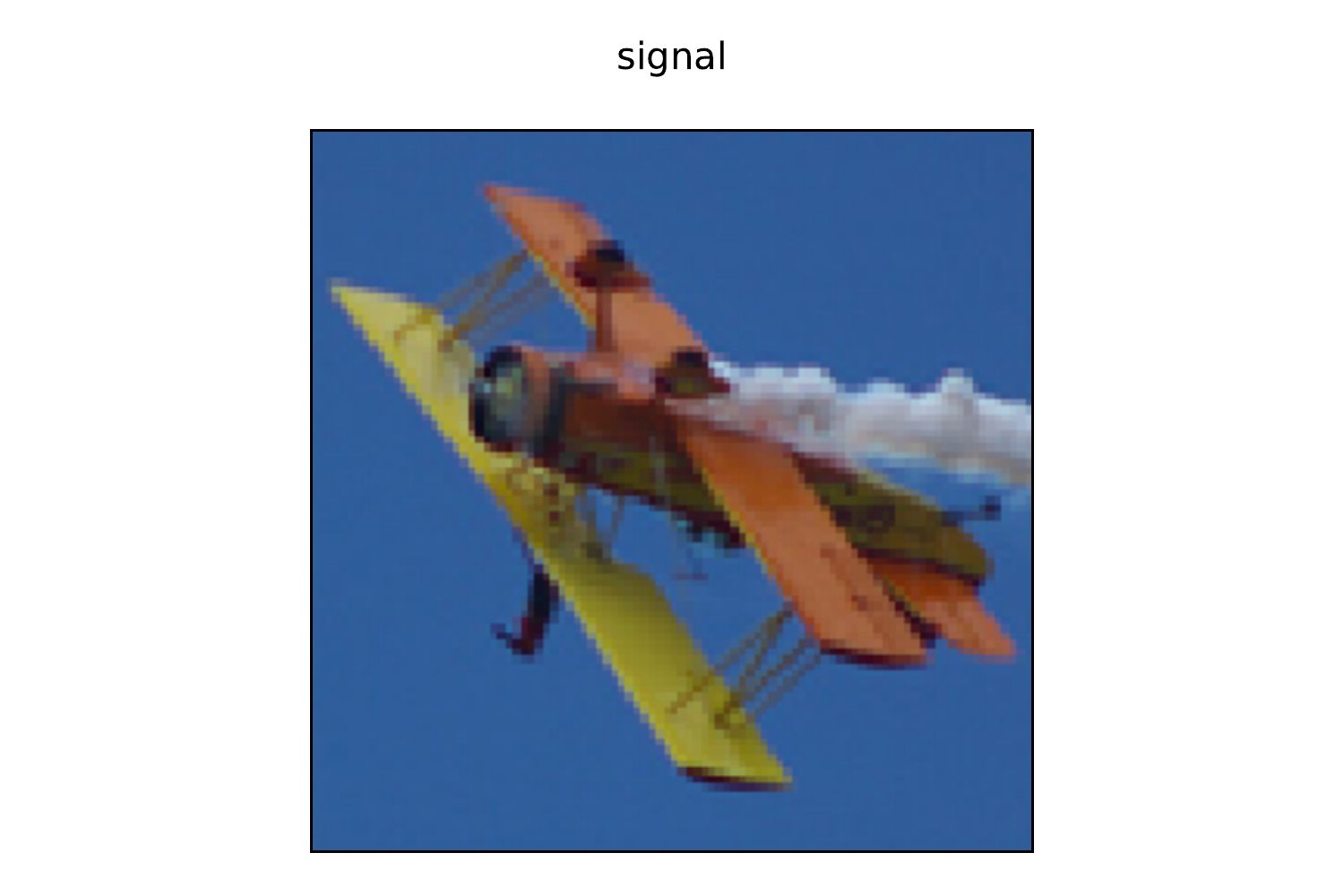} &        \includegraphics[trim={3.25cm 0.48cm 3.25cm 1.3cm},clip, width=.16\textwidth]{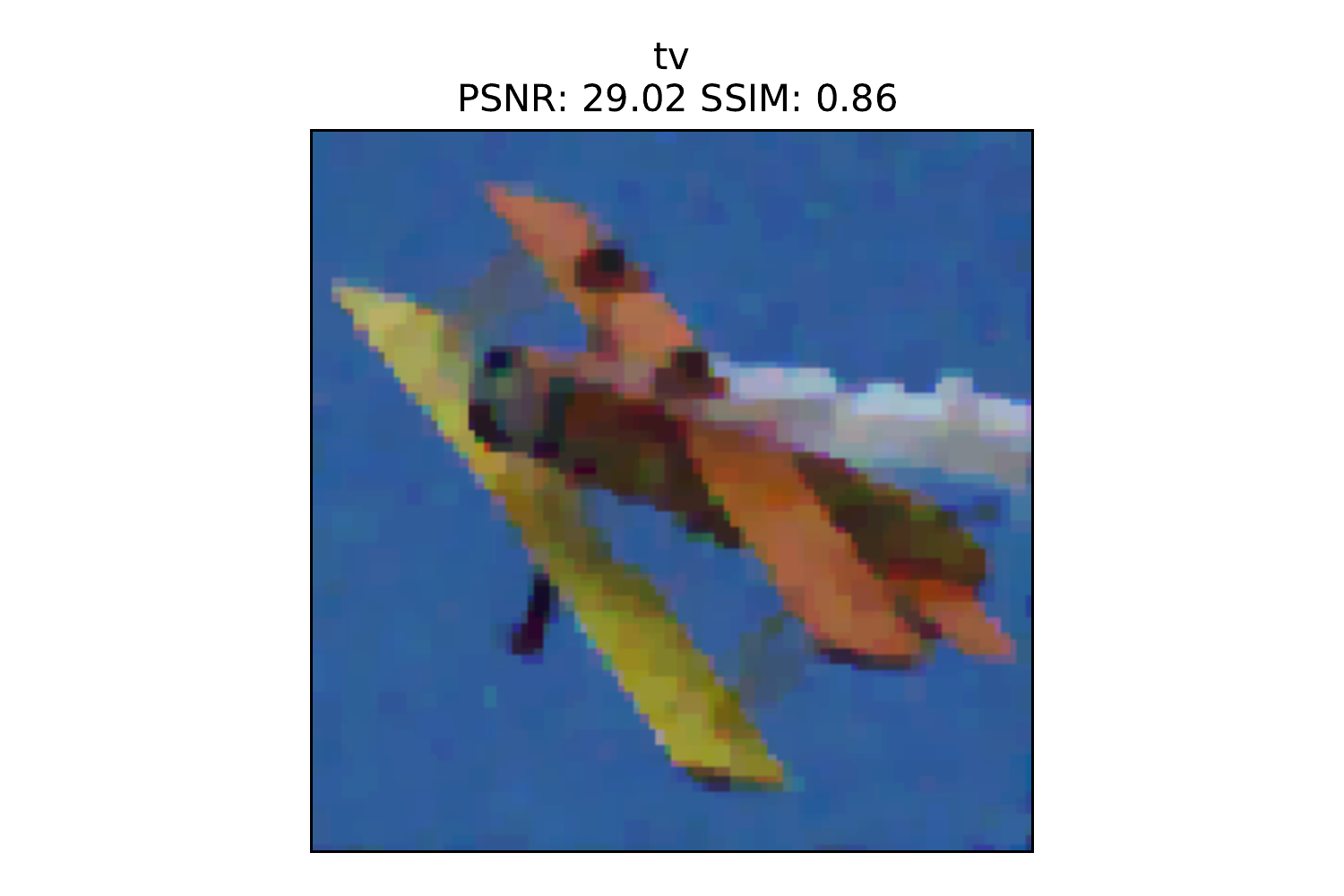} &        \includegraphics[trim={3.25cm 0.48cm 3.25cm 1.3cm},clip, width=.16\textwidth]{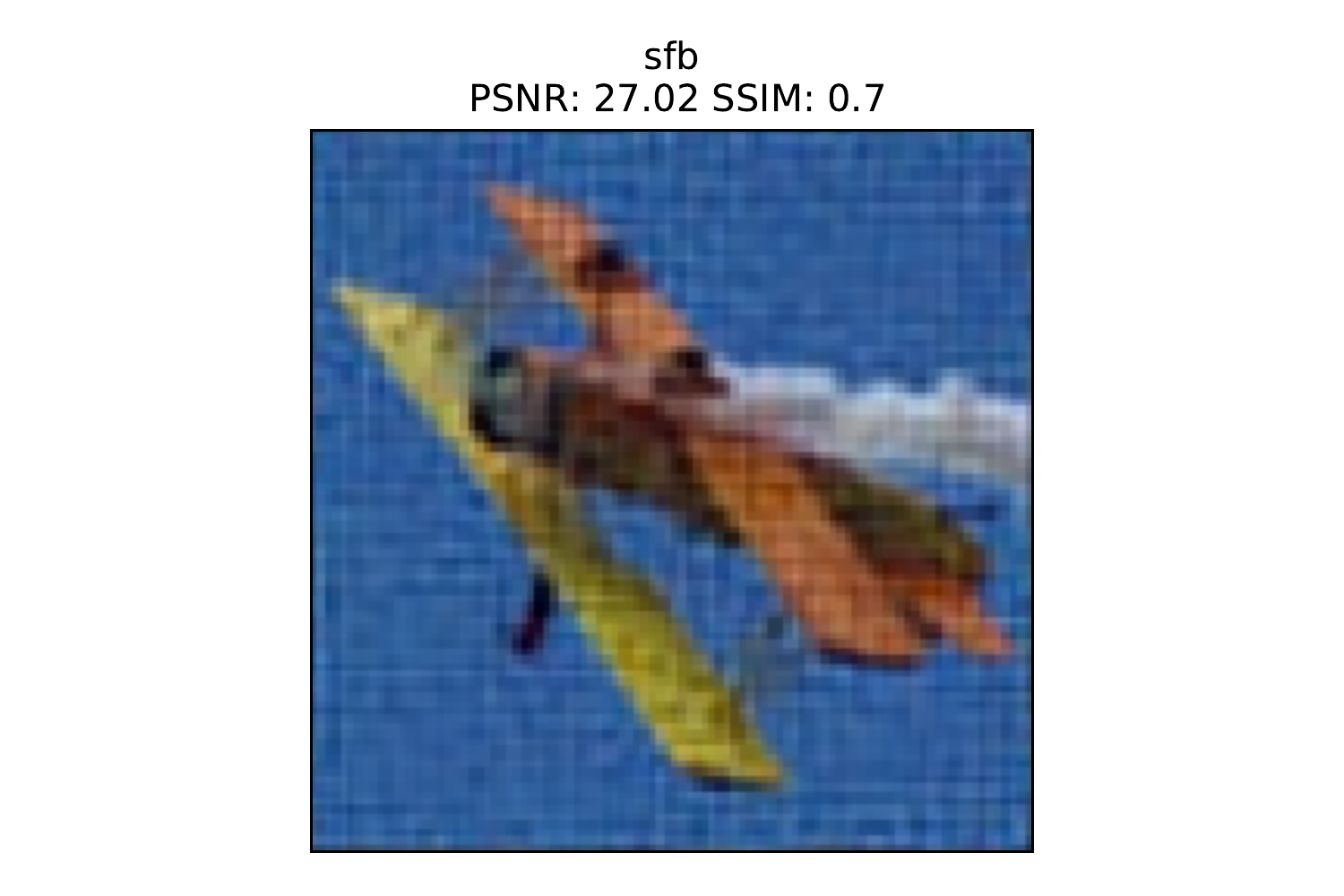} &        \includegraphics[trim={3.25cm 0.48cm 3.25cm 1.3cm},clip, width=.16\textwidth]{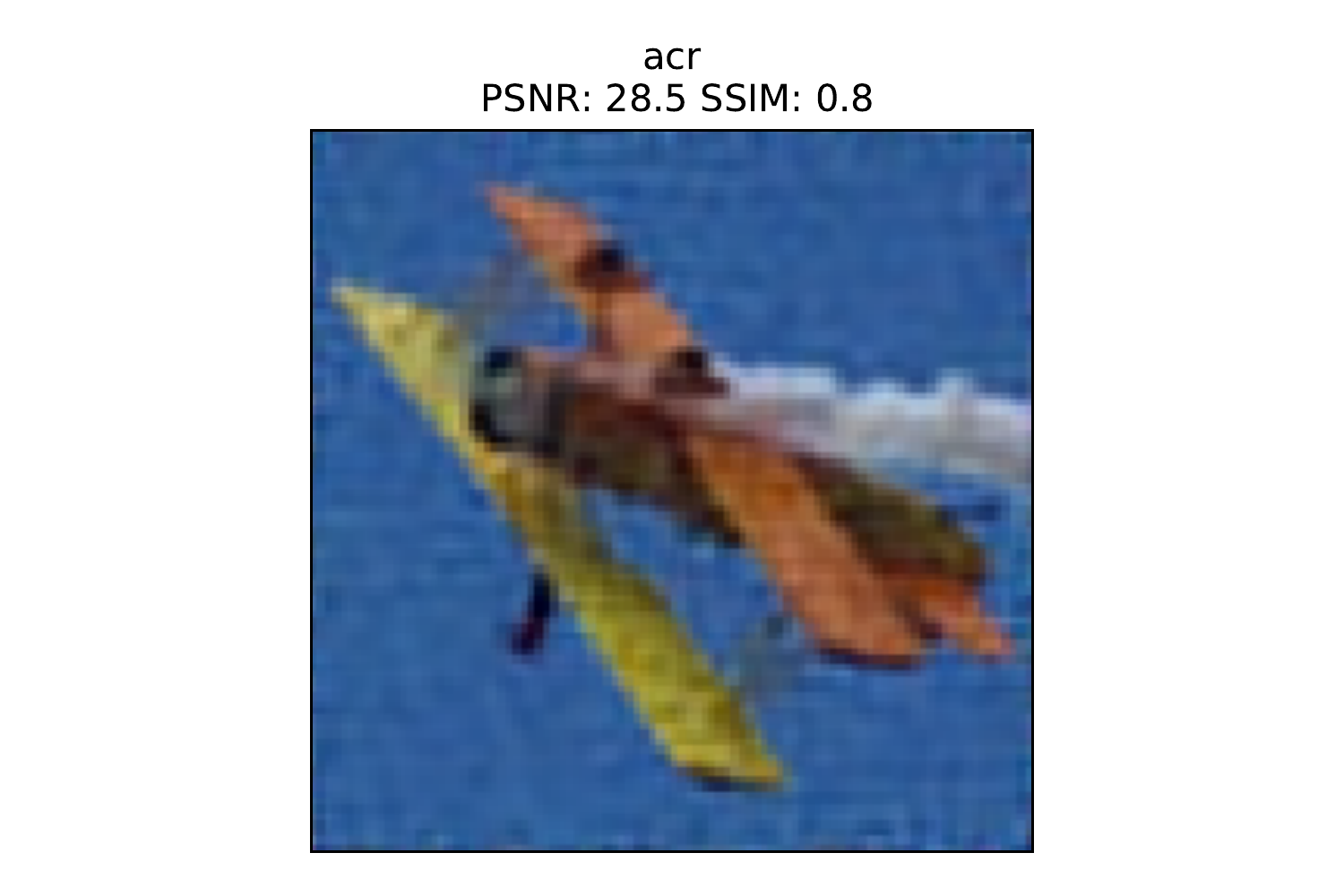} &        \includegraphics[trim={3.25cm 0.48cm 3.25cm 1.3cm},clip, width=.16\textwidth]{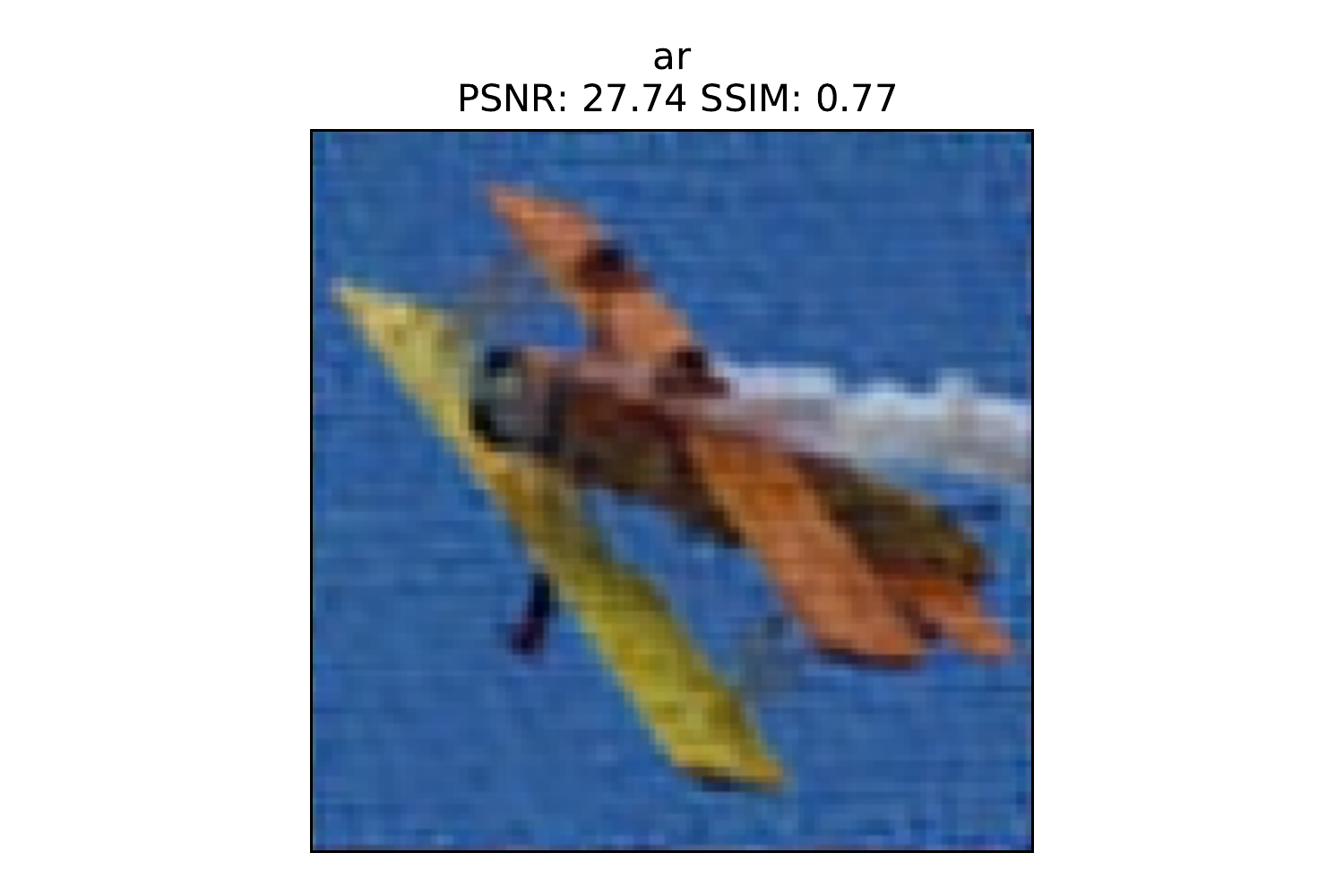} &        \includegraphics[trim={3.25cm 0.48cm 3.25cm 1.3cm},clip, width=.16\textwidth]{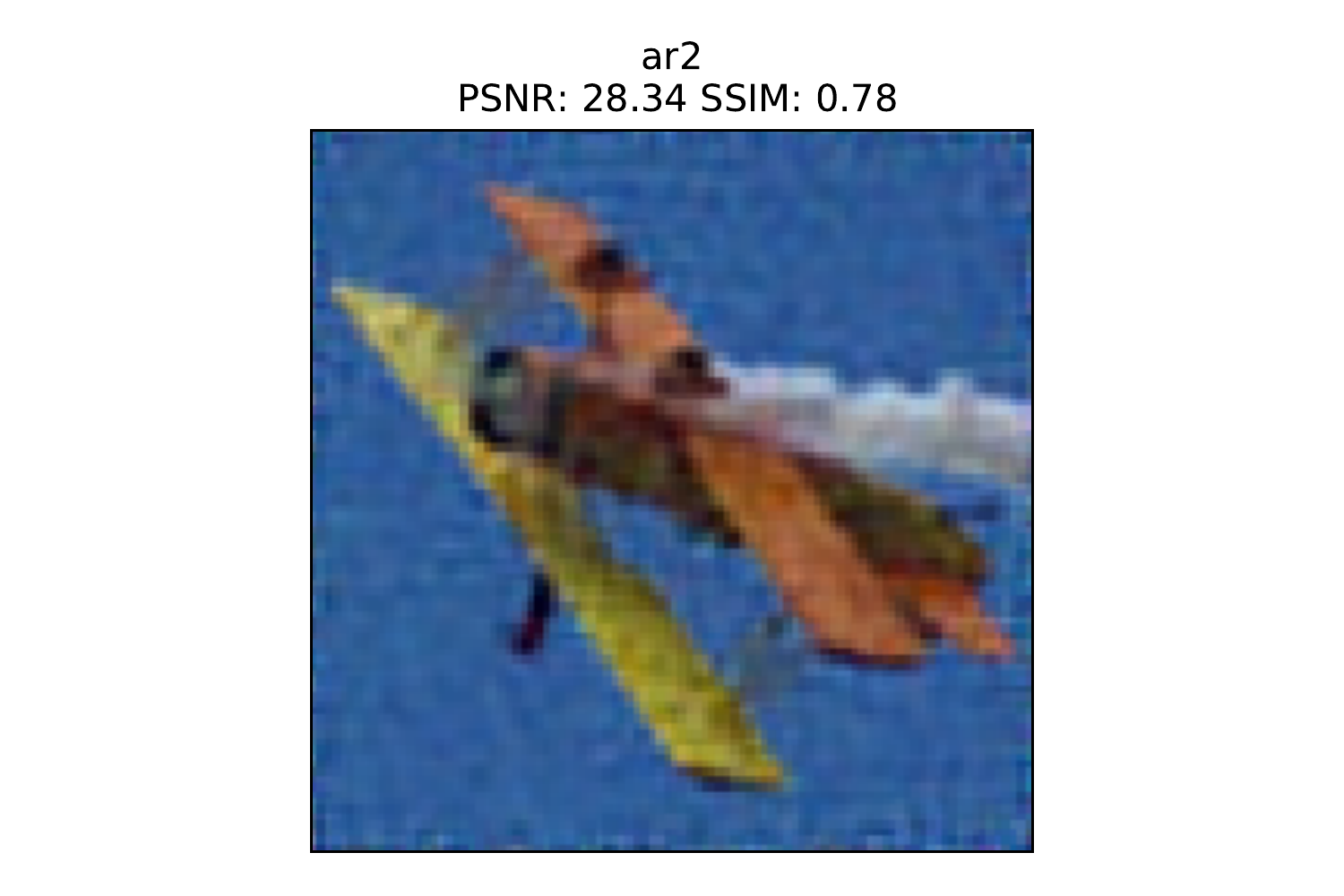} \\

        \includegraphics[trim={3.25cm 0.48cm 3.25cm 1.3cm},clip, width=.16\textwidth]{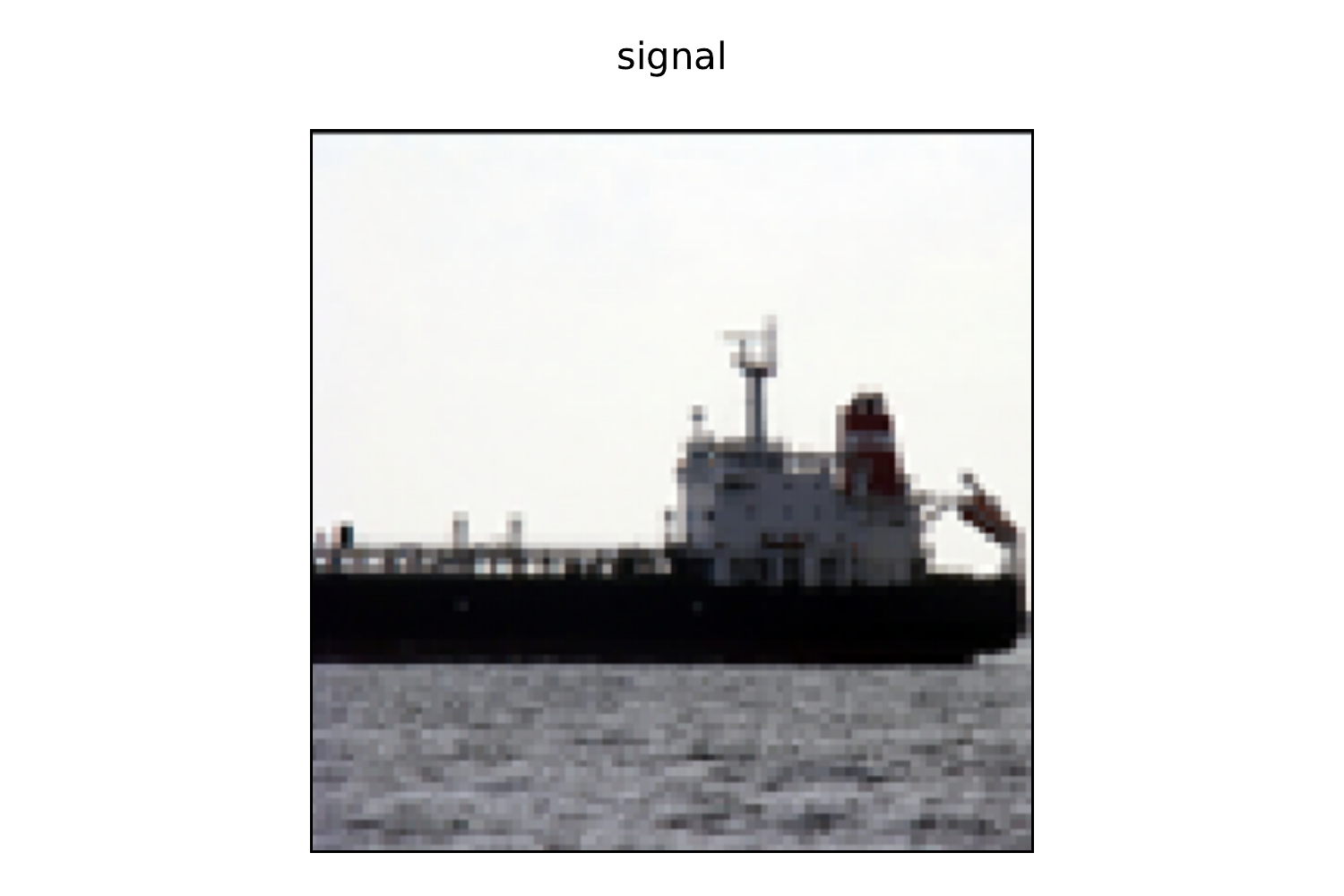} &        \includegraphics[trim={3.25cm 0.48cm 3.25cm 1.3cm},clip, width=.16\textwidth]{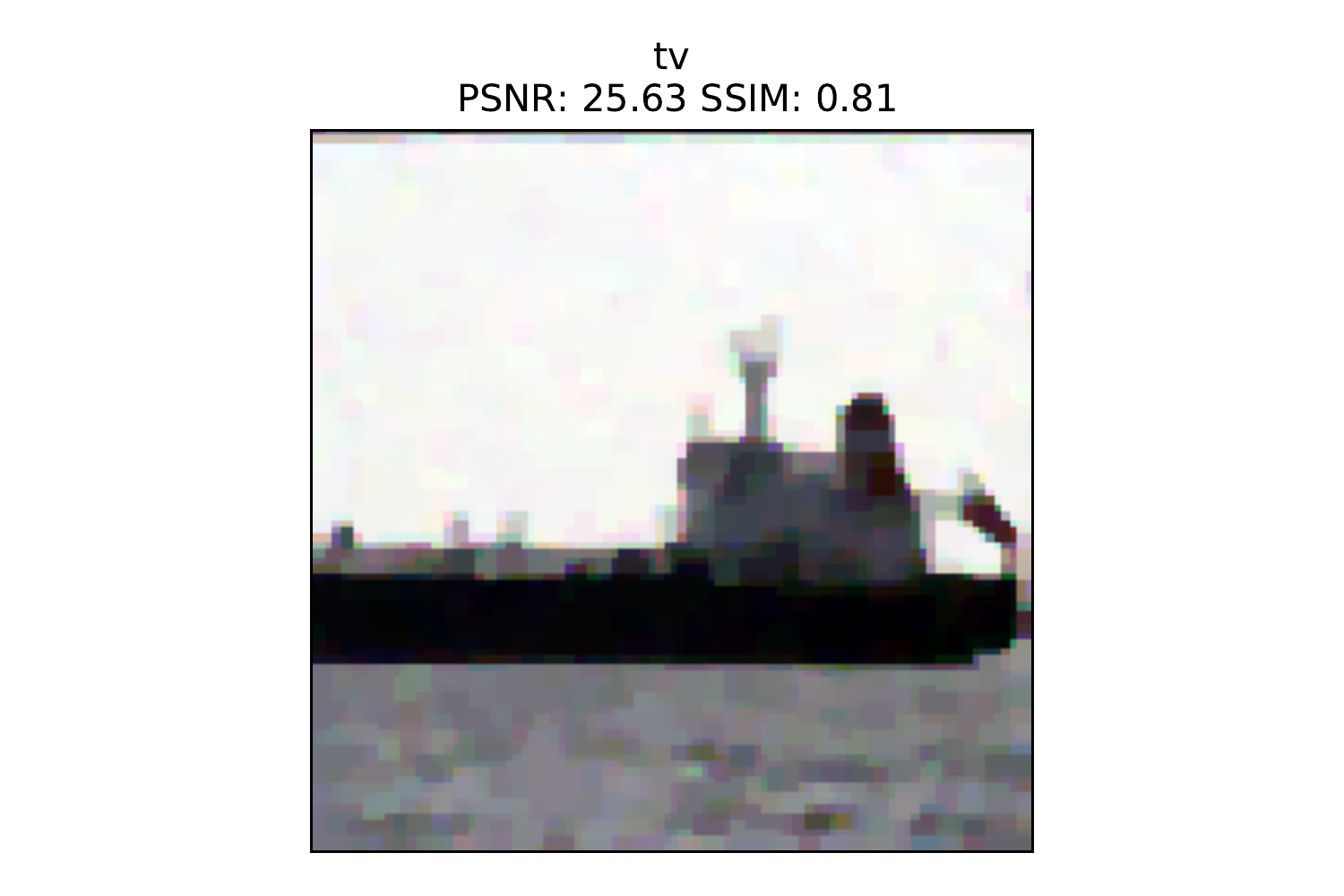} &        \includegraphics[trim={3.25cm 0.48cm 3.25cm 1.3cm},clip, width=.16\textwidth]{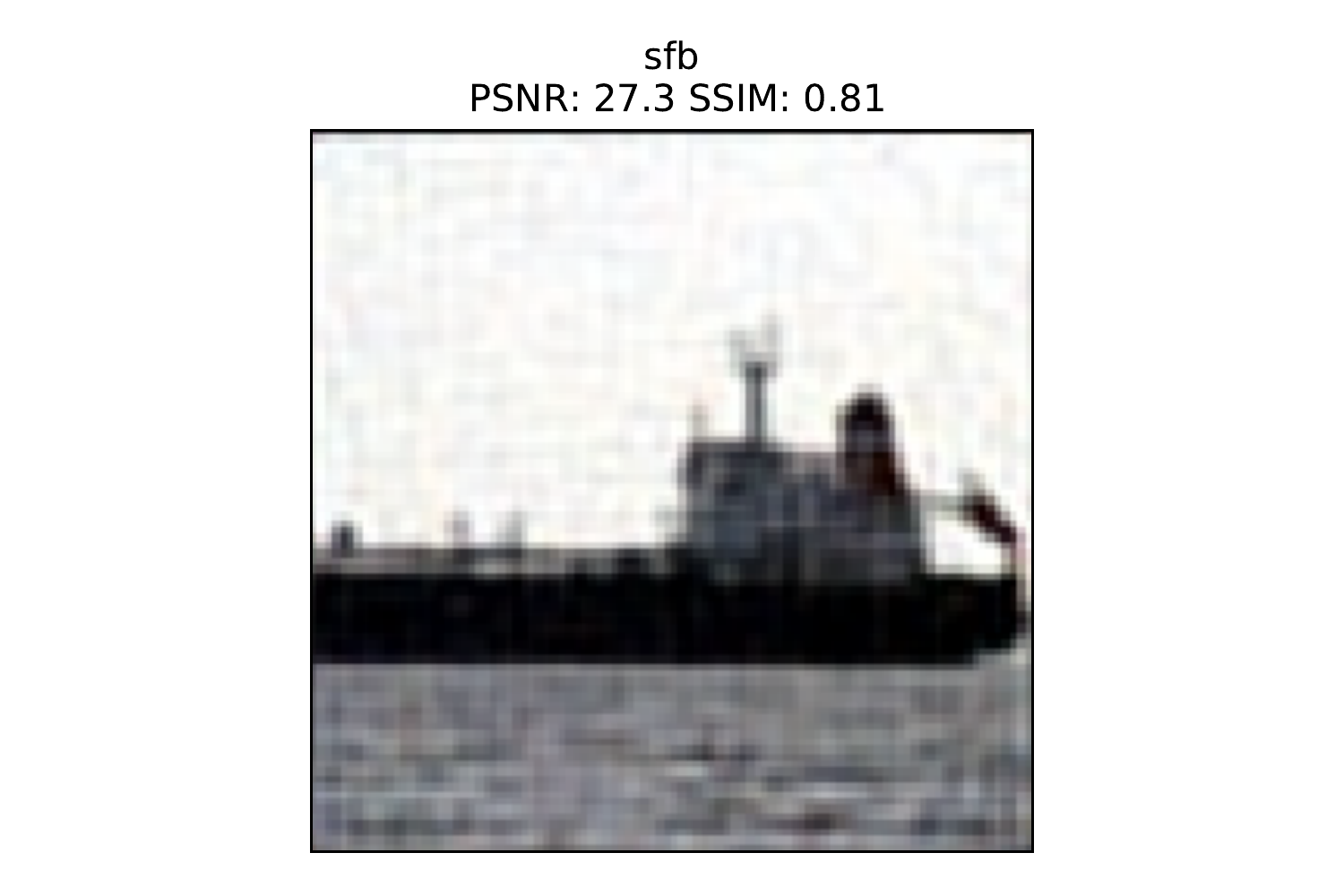} &        \includegraphics[trim={3.25cm 0.48cm 3.25cm 1.3cm},clip, width=.16\textwidth]{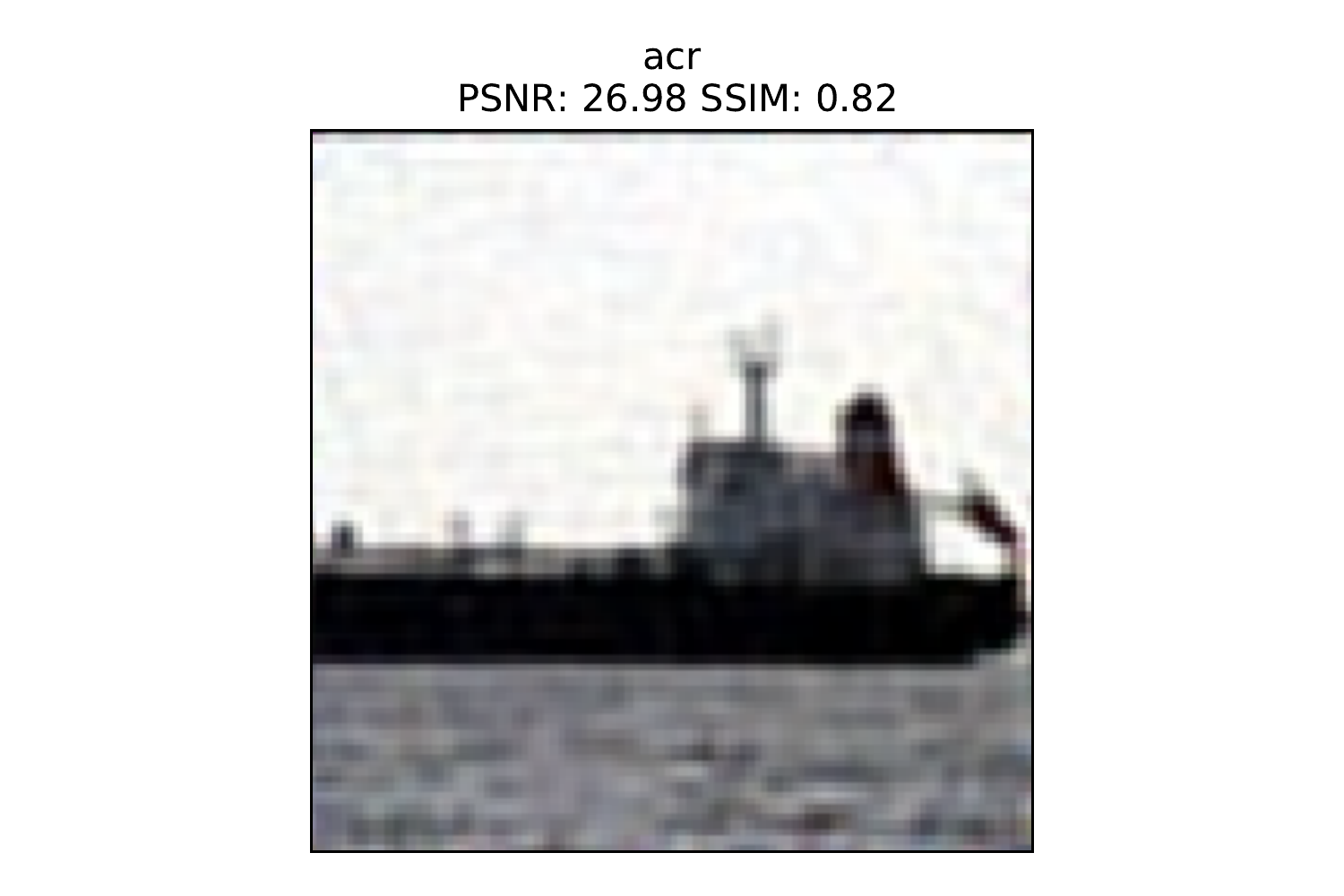} &        \includegraphics[trim={3.25cm 0.48cm 3.25cm 1.3cm},clip, width=.16\textwidth]{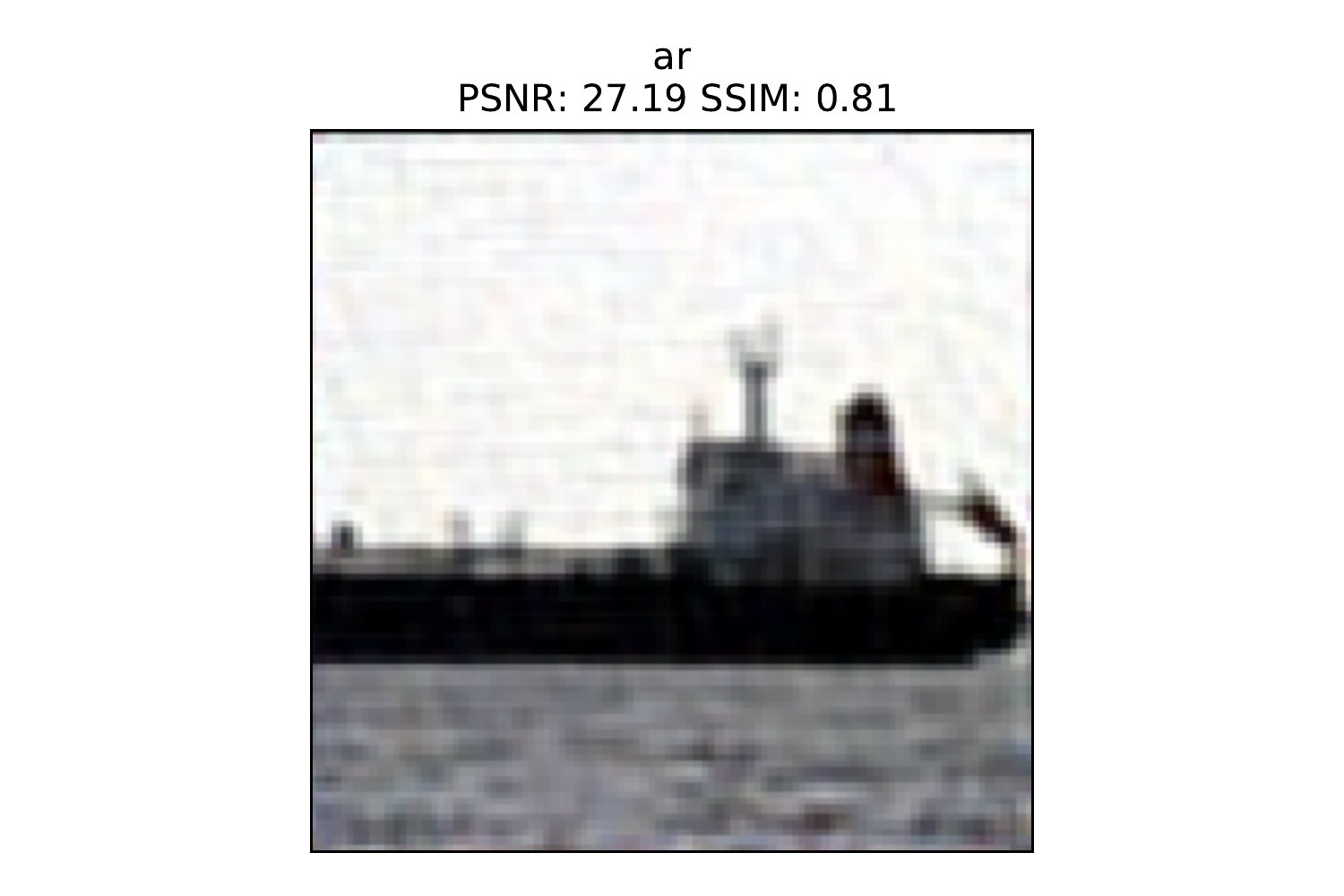} &        \includegraphics[trim={3.25cm 0.48cm 3.25cm 1.3cm},clip, width=.16\textwidth]{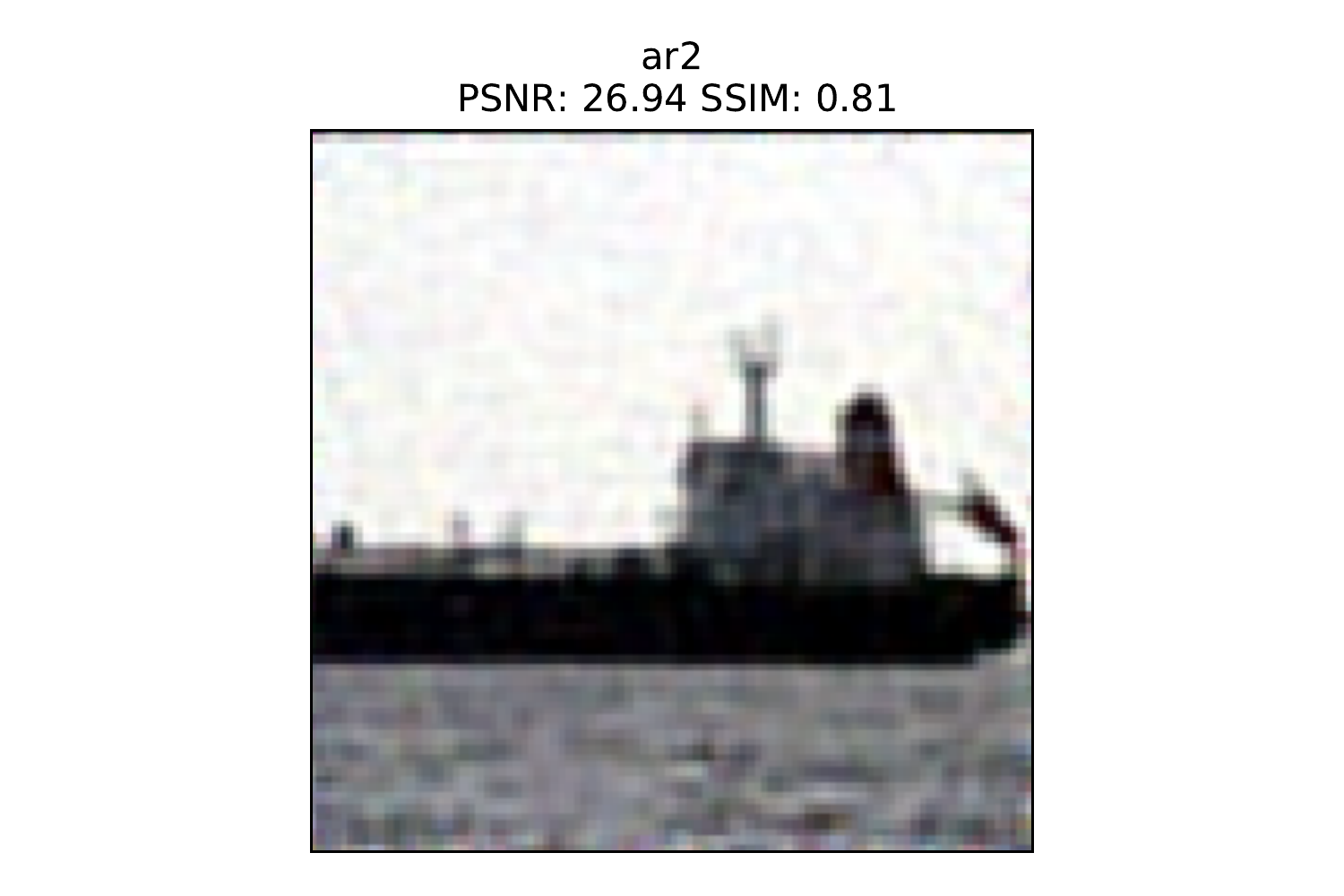} \\

        \includegraphics[trim={3.25cm 0.48cm 3.25cm 1.3cm},clip, width=.16\textwidth]{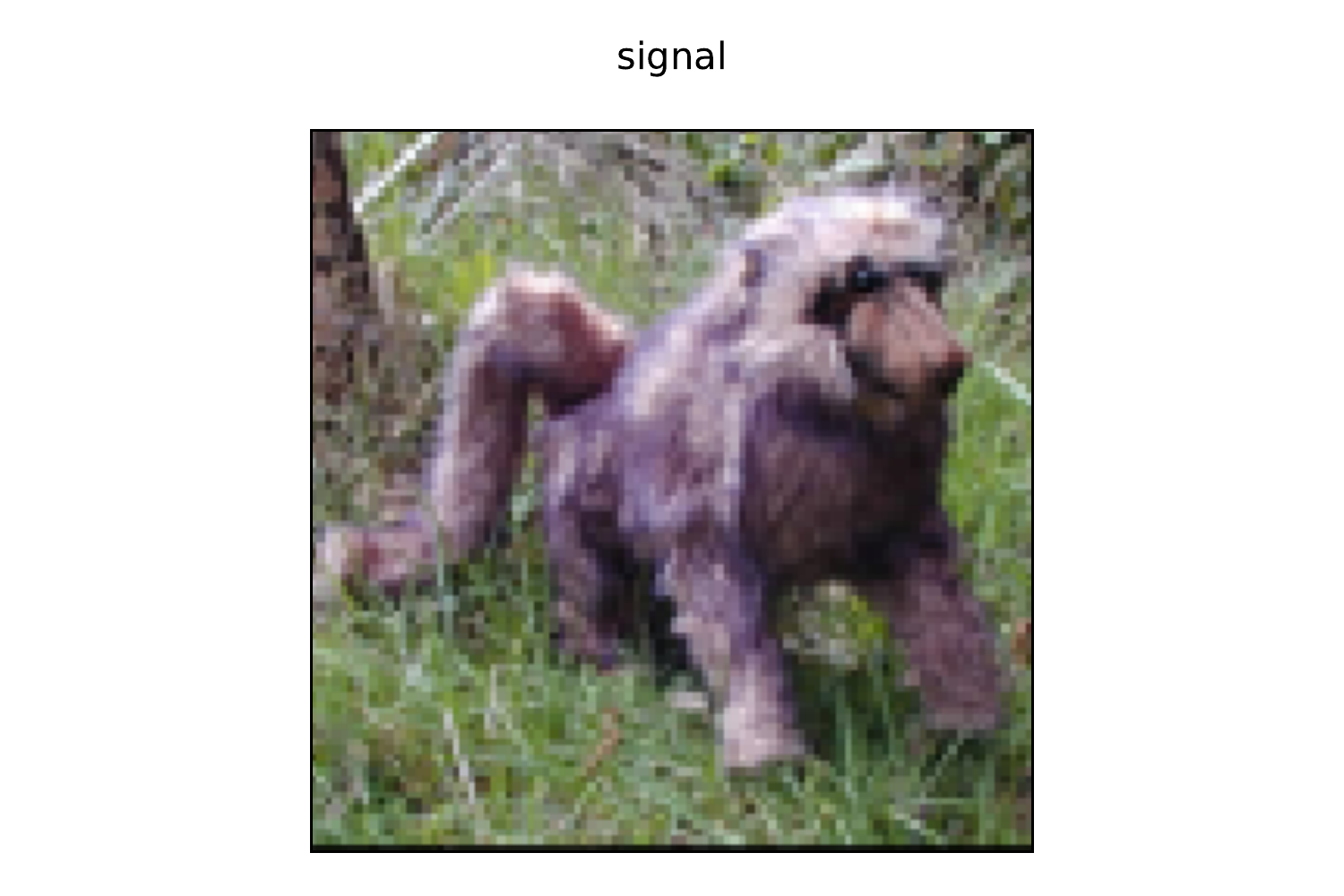} &        \includegraphics[trim={3.25cm 0.48cm 3.25cm 1.3cm},clip, width=.16\textwidth]{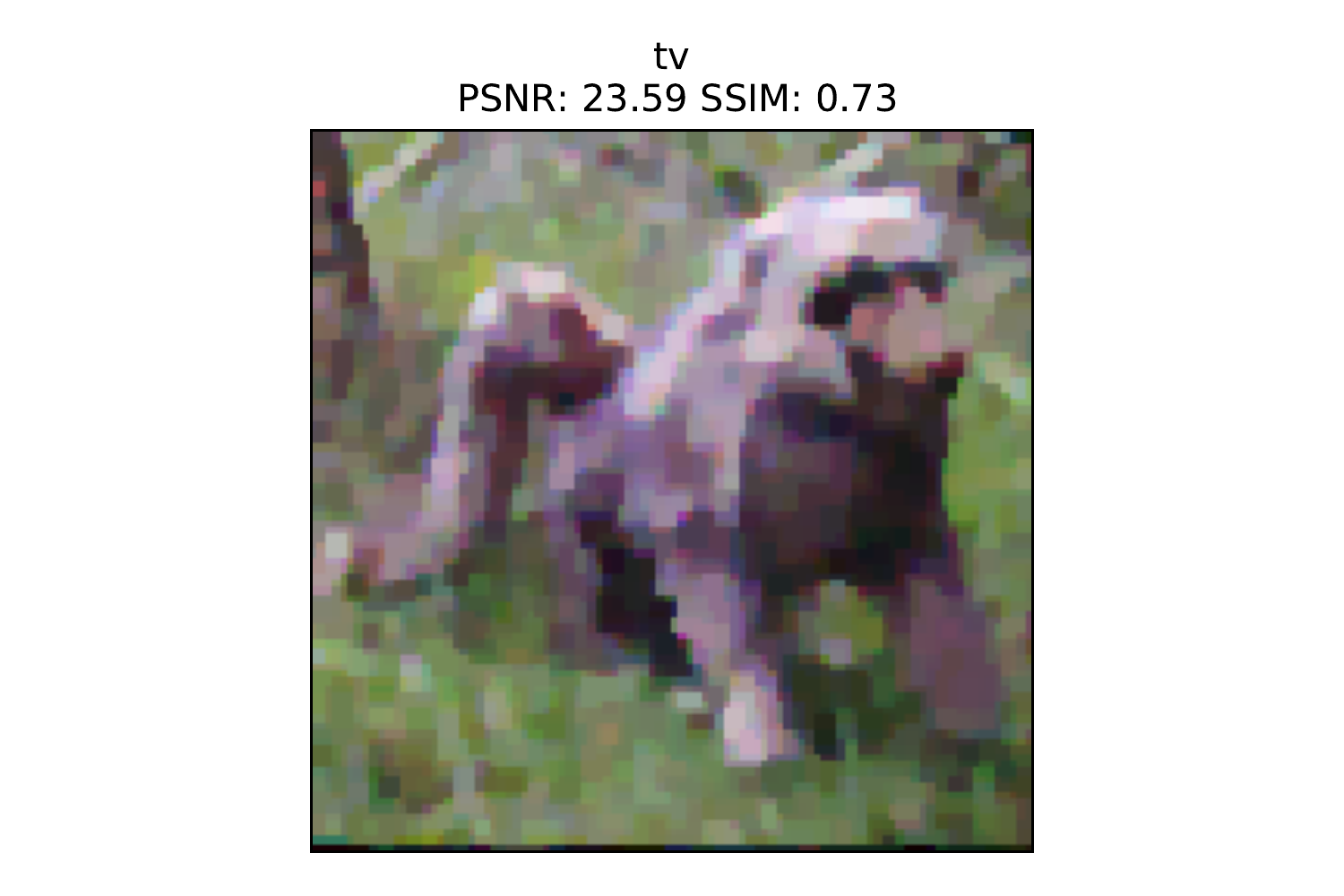} &        \includegraphics[trim={3.25cm 0.48cm 3.25cm 1.3cm},clip, width=.16\textwidth]{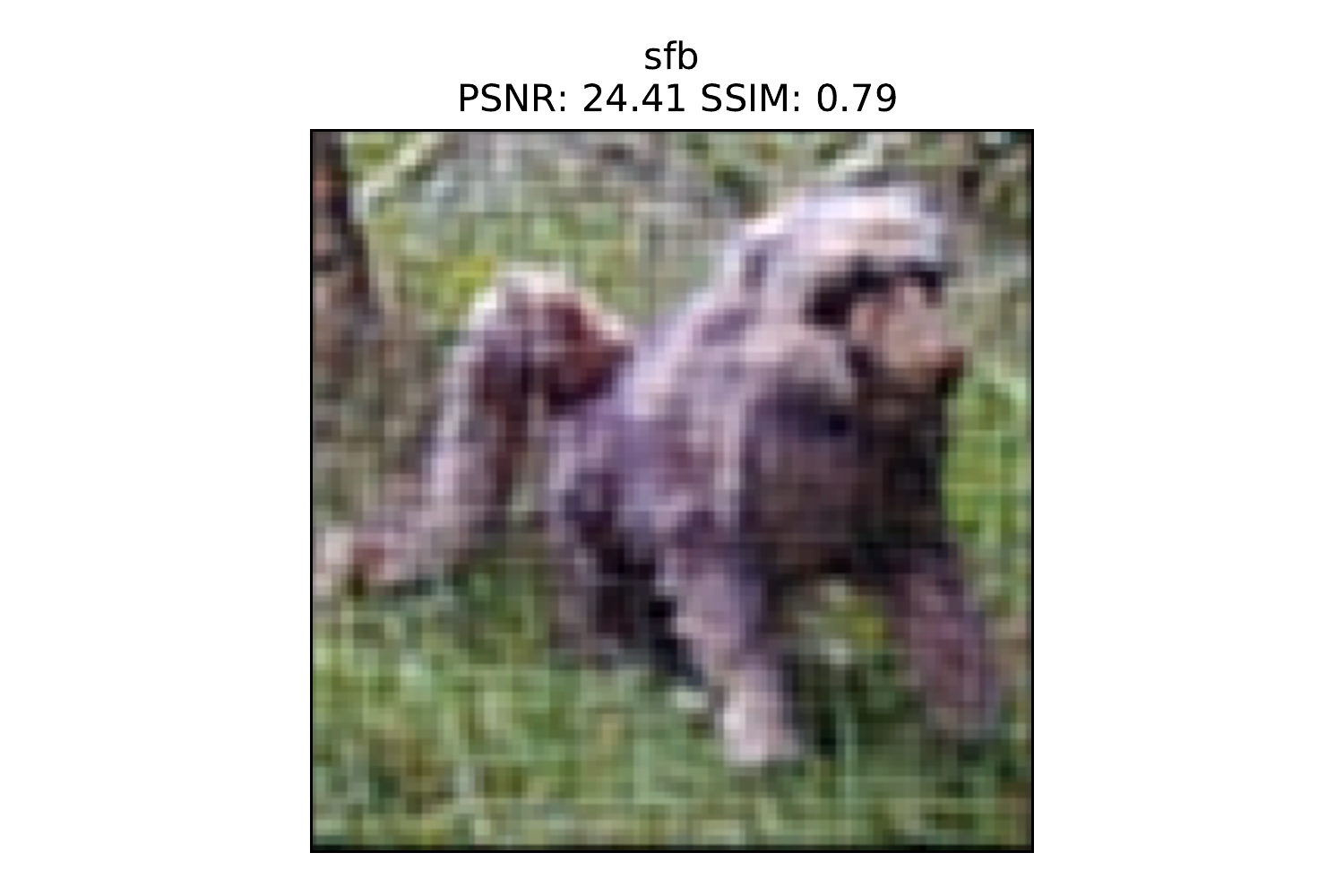} &        \includegraphics[trim={3.25cm 0.48cm 3.25cm 1.3cm},clip, width=.16\textwidth]{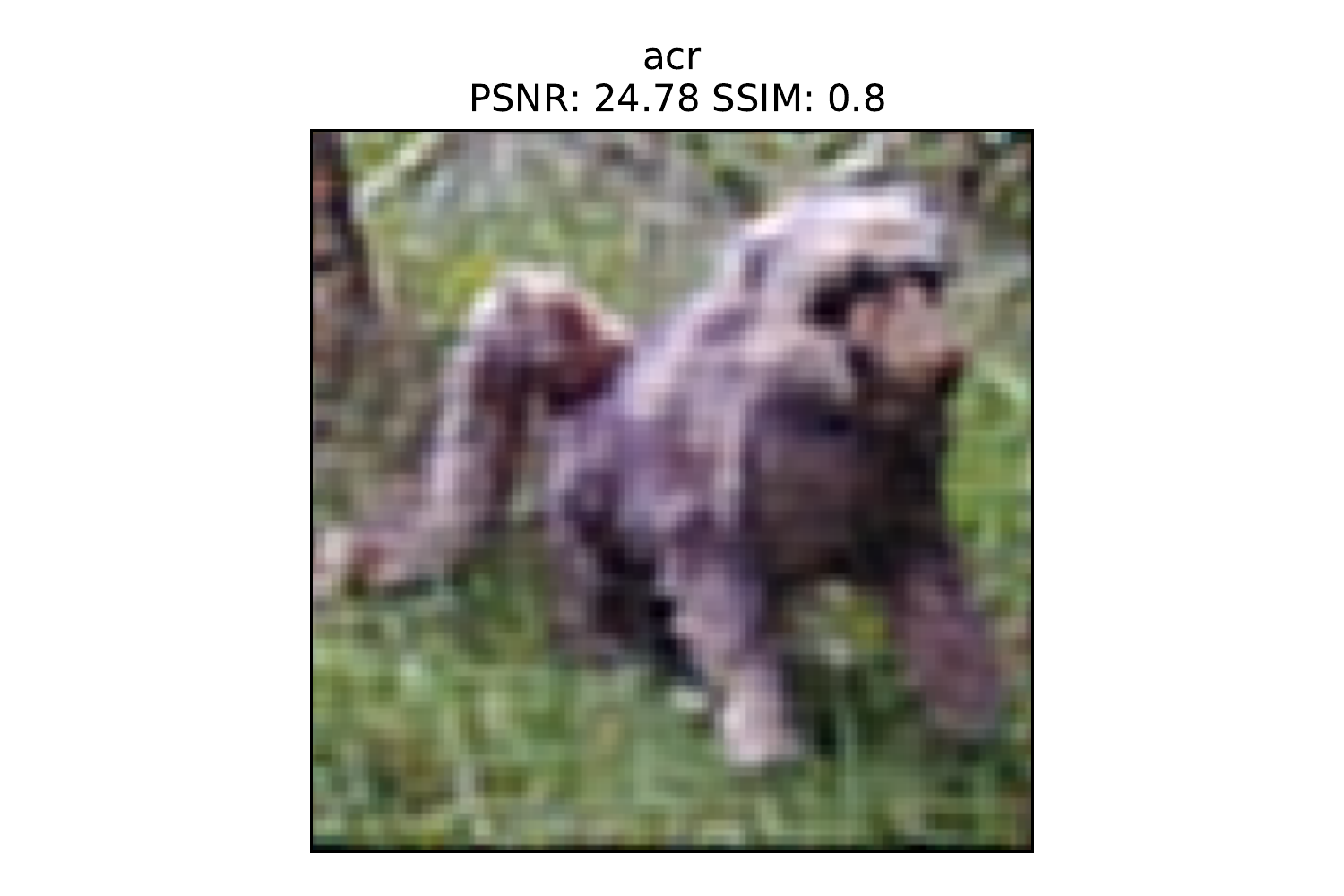} &        \includegraphics[trim={3.25cm 0.48cm 3.25cm 1.3cm},clip, width=.16\textwidth]{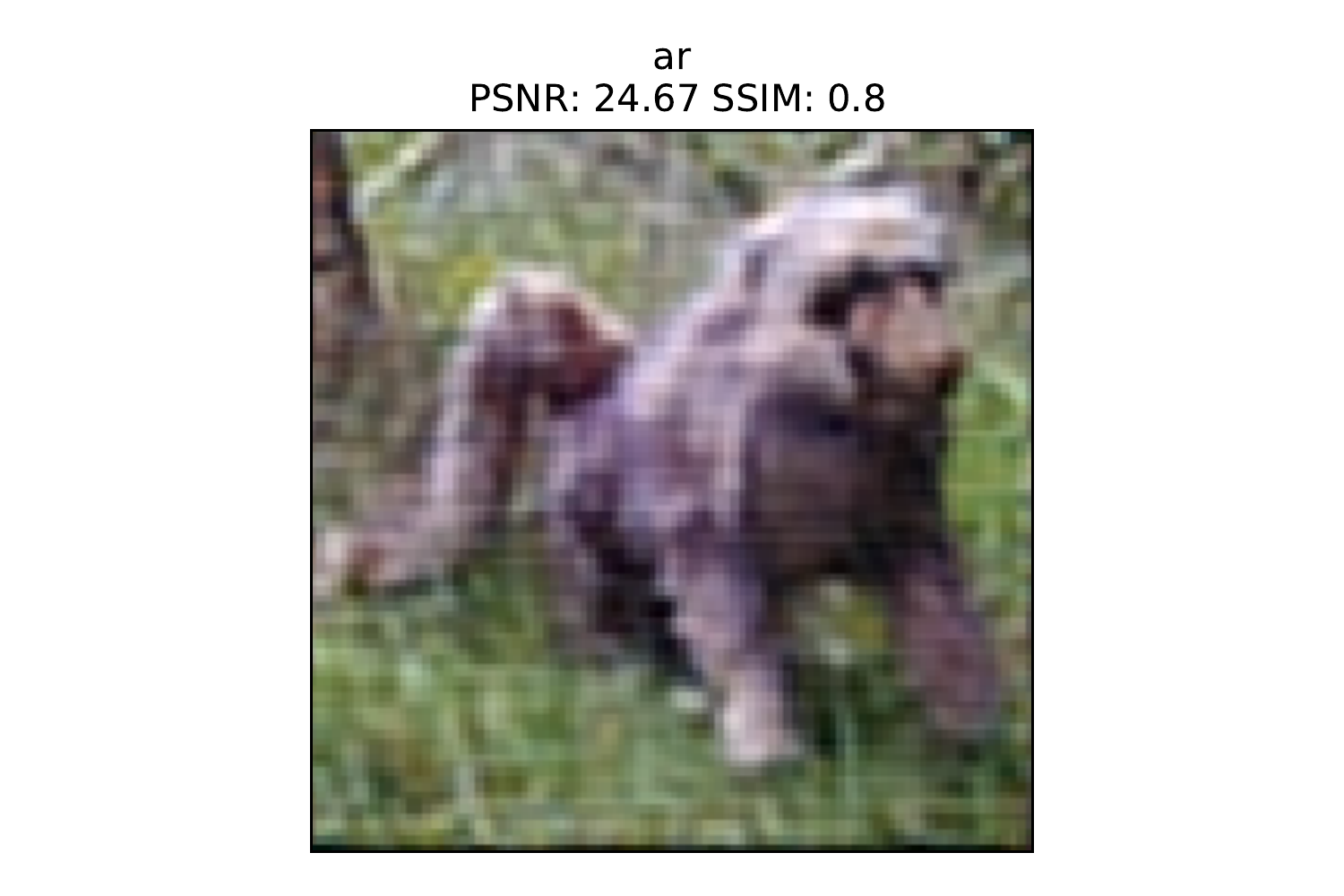} &        \includegraphics[trim={3.25cm 0.48cm 3.25cm 1.3cm},clip, width=.16\textwidth]{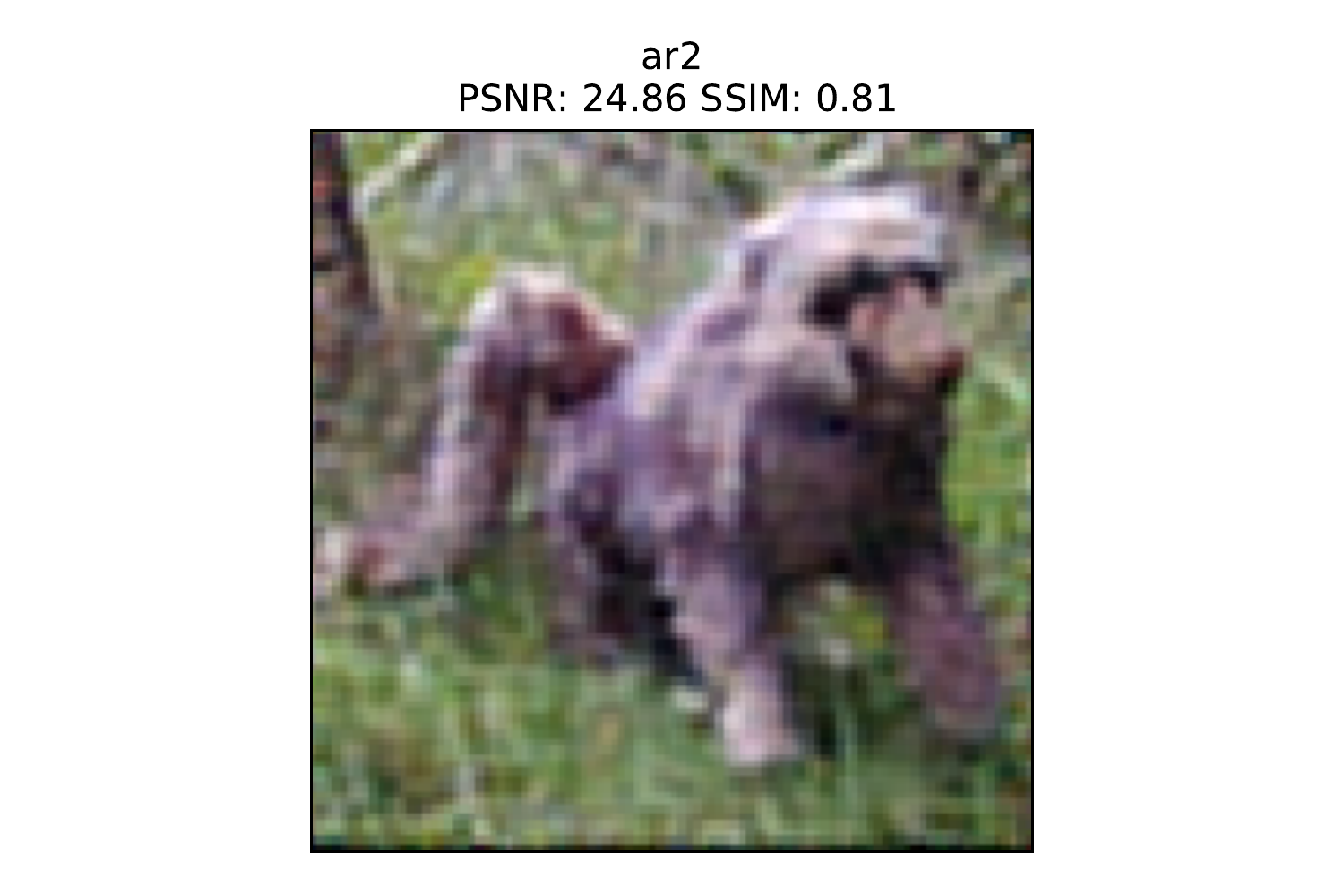} \\

        \includegraphics[trim={3.25cm 0.48cm 3.25cm 1.3cm},clip, width=.16\textwidth]{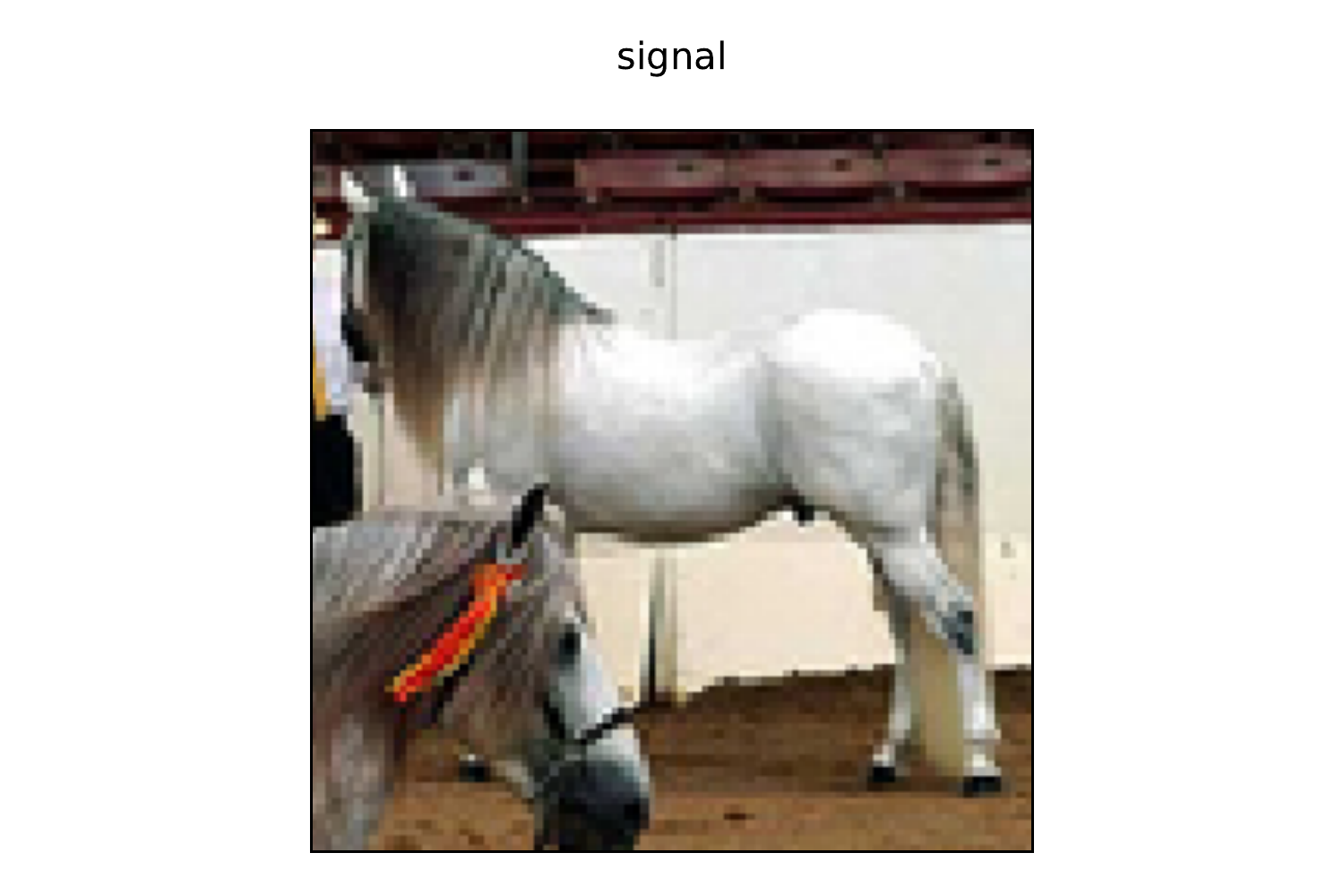} &        \includegraphics[trim={3.25cm 0.48cm 3.25cm 1.3cm},clip, width=.16\textwidth]{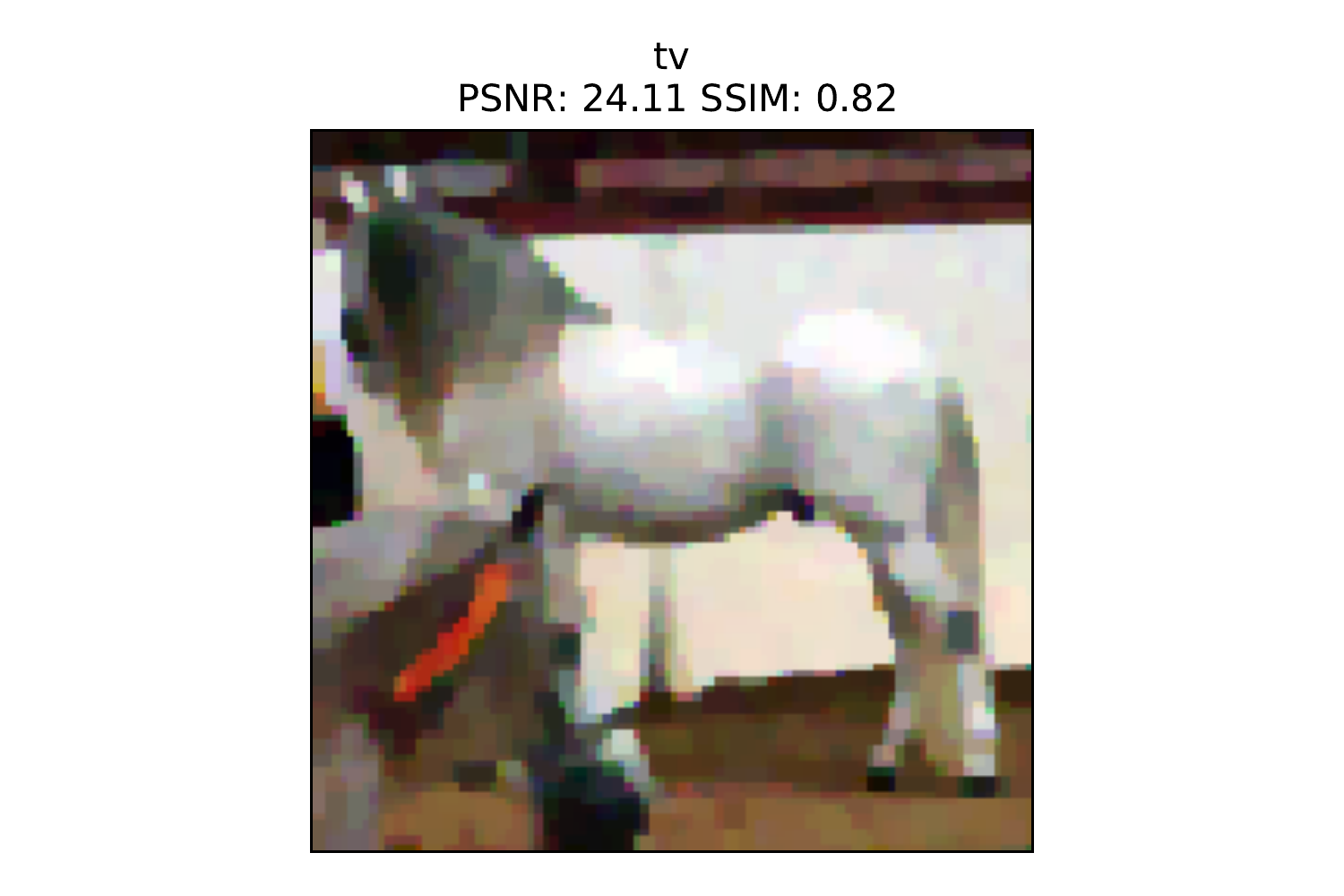} &        \includegraphics[trim={3.25cm 0.48cm 3.25cm 1.3cm},clip, width=.16\textwidth]{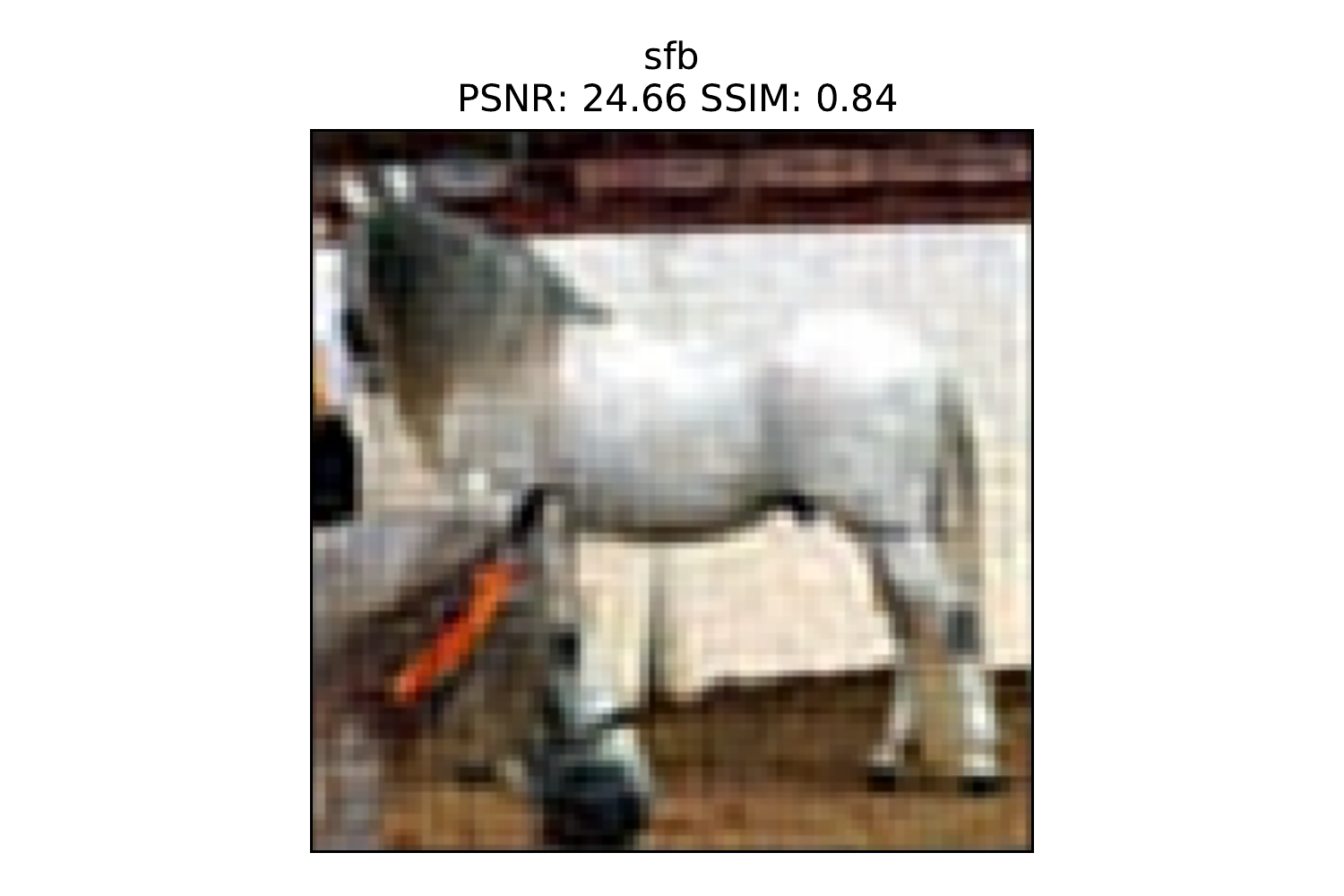} &        \includegraphics[trim={3.25cm 0.48cm 3.25cm 1.3cm},clip, width=.16\textwidth]{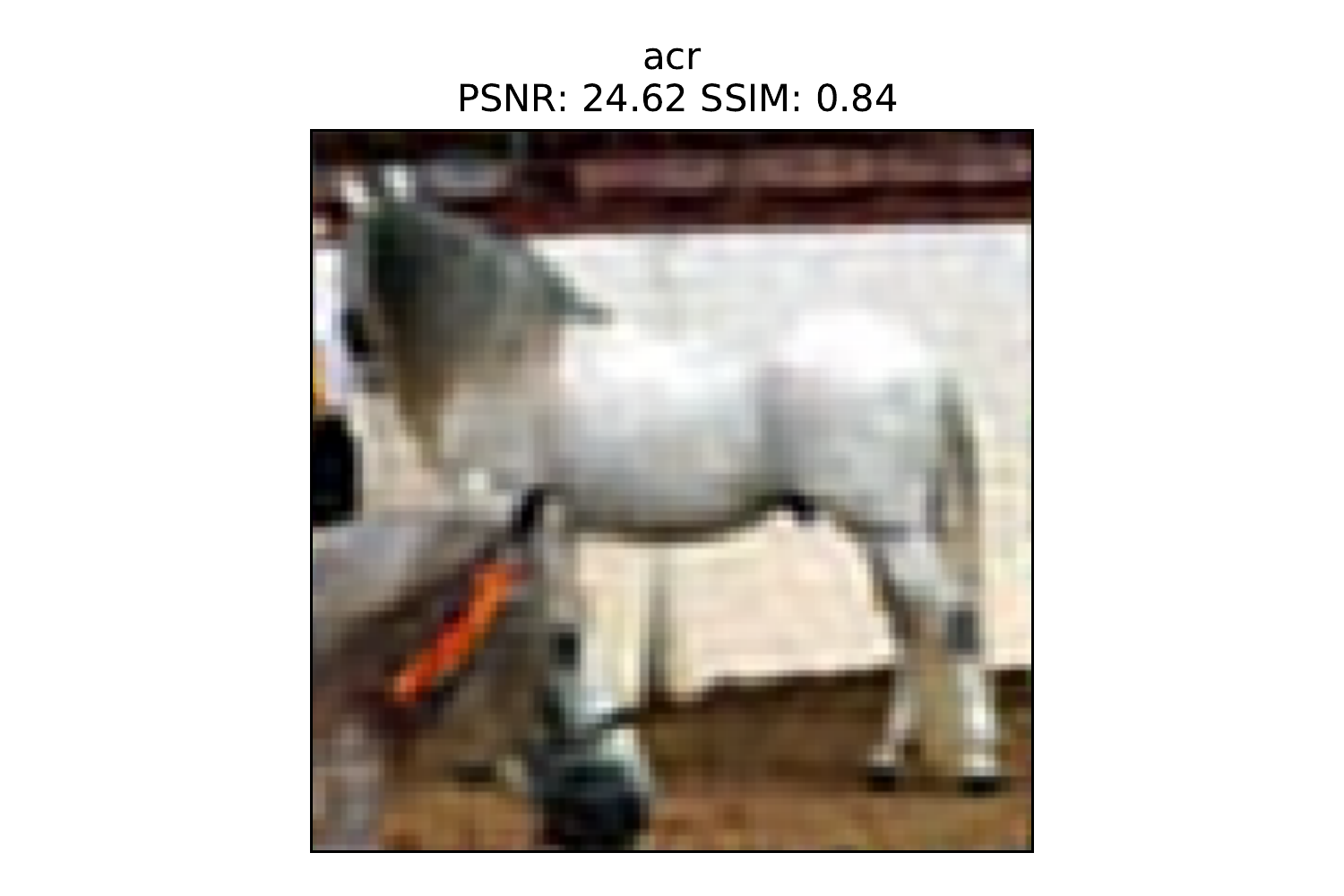} &        \includegraphics[trim={3.25cm 0.48cm 3.25cm 1.3cm},clip, width=.16\textwidth]{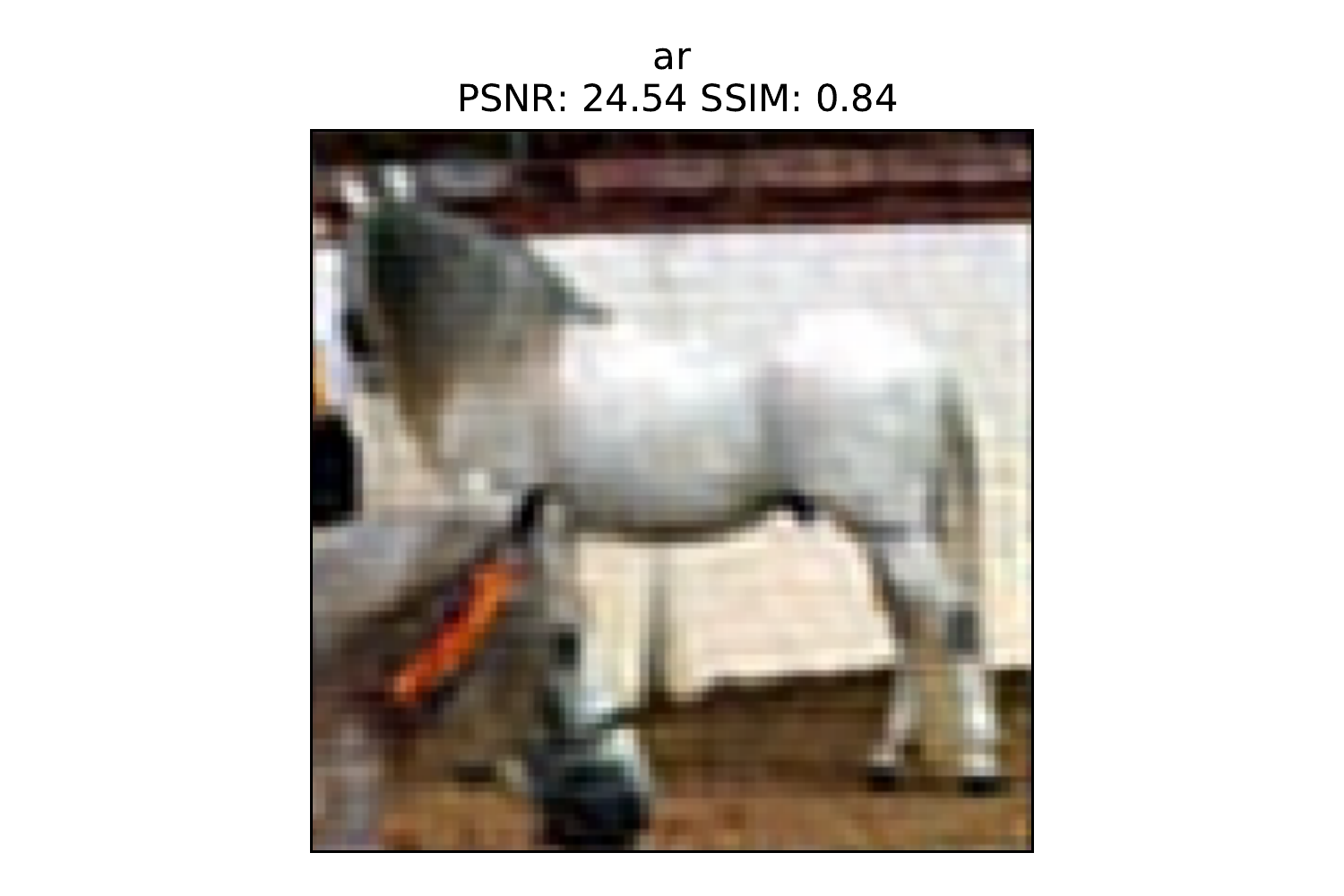} &        \includegraphics[trim={3.25cm 0.48cm 3.25cm 1.3cm},clip, width=.16\textwidth]{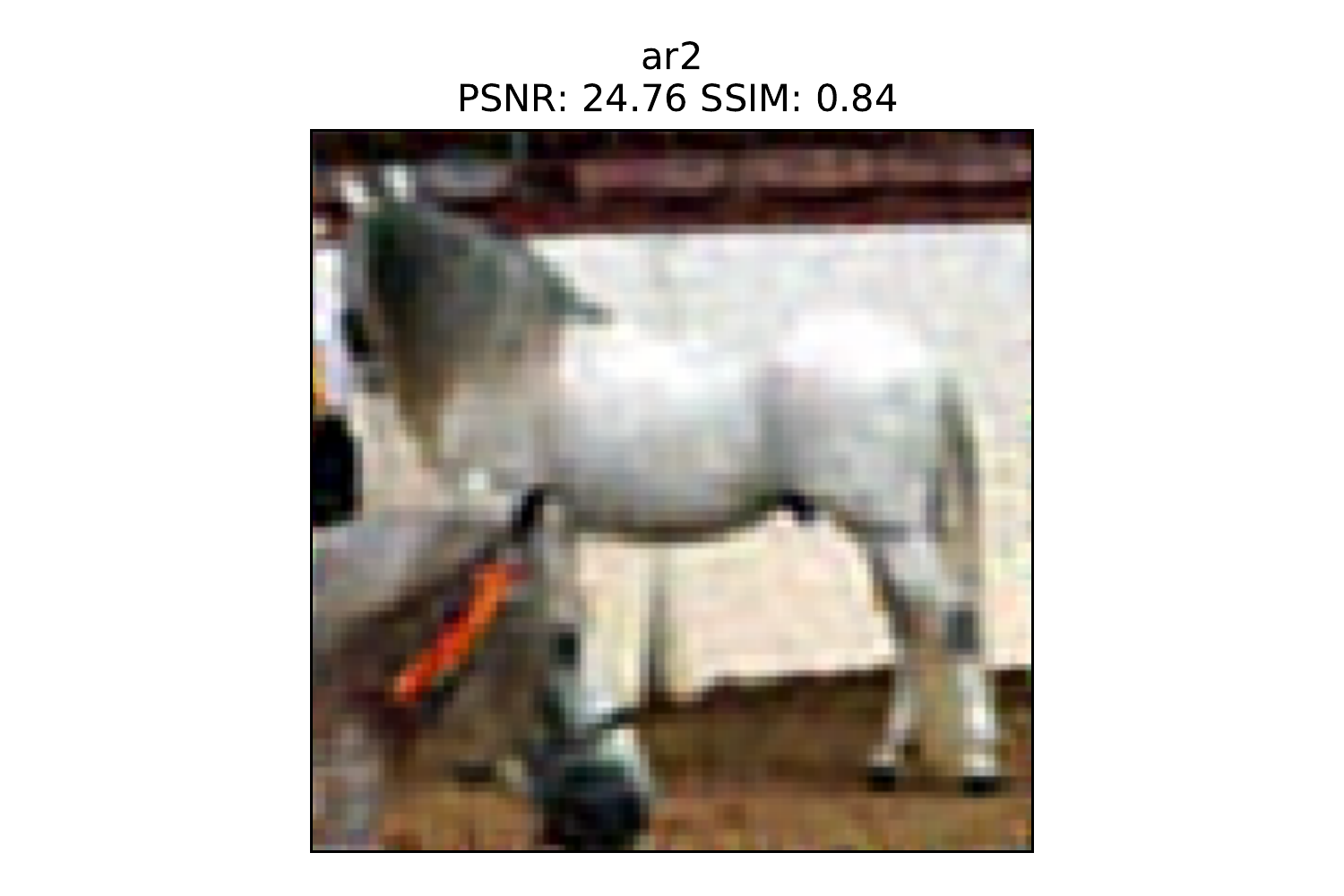}

        \\[1ex]
        % \multirow{3}{*}{\textbf{PSNR stats}:}
        % &\ \ \ \ mean: $25.50$& \ \ \ \ mean: $26.05$& \ \ \ \ mean: $26.55$& \ \ \ \ mean: $26.35$& \ \ \ \ mean: $26.57$\\
        % &\ \ median: $25.01$&\ \ median: $25.82$&\ \ median: $26.12$&\ \ median: $25.99$&\ \ median: $26.21$\\
        % & std.\ div.: $2.08$& std.\ div.: $1.71$& std.\ div.: $2.05$& std.\ div.: $1.98$& std.\ div.: $1.88$\\
        %  \hline
        % \multirow{3}{*}{\textbf{SSIM stats}:}
        % & \ \ \ mean: $0.80$& \ \ \ mean: $0.81$& \ \ \ mean: $0.83$& \ \ \ mean: $0.82$&\ \ \ mean: $0.83$\\
        % & \ median: $0.80$& \ median: $0.81$& \ median: $0.83$& \ median: $0.82$& \ median: $0.83$\\
        % & std.\ div.: $0.05$& std.\ div.: $0.04$& std.\ div.: $0.03$& std.\ div.: $0.03$& std.\ div.: $0.03$\\
    \end{tabular}
    \\[2ex]
    \caption{A visual comparison of different methods for deblurring some representative images in the STL10 dataset.}
    \label{table:deblurring_results}
\end{figure*}
\subsection{Image deblurring}
% We begin by describing the setup of the dataset. After that, we describe specific architectures we use and then move on to the training itself. Finally, we describe the setup we used to solve the variational problems before presenting and discussing the results.
We will now describe the experimental setup and numerical results for the image deblurring task conducted on the STL-10 dataset~\cite{coates2011analysis}. The deblurring dataset is created by computing the noisy measurements, $\y^\delta$, by first smoothing the ground truth images, $\x\in[0,1]^{3\times96\times96}$, from the STL-10 dataset, with a $3\times3$ averaging filter in each of the RGB channels and subsequently adding zero-centered Gaussian noise with a standard deviation of 0.05.
%All variants of AR and ACR are trained on pairs $(\x_i, \Op{A}^{\dagger} \y^\delta_i)$, where we approximated the action of $\Op{A}^{\dagger}$ on the measurement via an overfitting Landweber reconstruction, i.e., by stopping it past the Morozov's principle.\\
All variants of AR and ACR are trained on pairs $(\x_i, \Op{A}^{\dagger} \y^\delta_i)$, where the action of $\Op{A}^{\dagger}$ is approximated via an overfitting Landweber reconstruction. More specifically, we approximate $\Op{A}^{\dagger}\y^\delta$ by minimizing the data-fidelity error $\|\Op{A}\x-\y^\delta\|_2^2$ via gradient-descent and terminating the iterations when the Morozov's discrepancy principle is satisfied (i.e., $\|\Op{A}\x_k-\y^\delta\|_2$ falls below the noise level $\delta$, where $\x_k$ is the $k^{\text{th}}$ iterate of gradient-descent.)

\indent The architecture of the Lipschitz-convex component $\R'$ in ACR is a discretized version of the architecture described in Section~\ref{icnn_sec} and is built with $L=5$ layers. We set $\varphi_i$ to be the leaky-ReLU function with a negative slope of 0.2 for layers $i=0, \dots, L$. All $\B_i$ and $\W_i$ are given by convolutional layers with kernels of size $5\times5$ and $32$ output channels. The dimensional reduction to a single real-valued output in the last layer occurs via a global average-pooling. The negative weights in the filters of $\B_i$ are zero-clipped after each training step to preserve convexity. The convolutional layer in the SFB consists of a $7\times7$ kernel and has $32$ output channels. We compare the ACR and the SFB with the original AR and, due to its slightly sub-optimal performance, a second adversarial regularizer which we refer to as AR2. The AR2 regularizer has the same architecture as the ACR, but the $\B_i$'s are not restricted to be non-negative, thus allowing it to be non-convex. AR2 has significantly fewer parameters than AR and is less prone to overfitting as a consequence.\\
\indent All models are trained for 8 epochs with a batch size of 100 using the \textit{Adam} optimizer with a learning rate of $5\times 10^{-5}$ and $\beta=(0.9, 0.99)$. The number of trainable parameters in the networks are given in Table~\ref{table:number_of_parameters_deblurring}.\\
\indent Subsequent to training, we compute the reconstructions by minimizing the corresponding variational functional. The optimization is carried out by using gradient descent for 4096 steps with a step size of $10^4 / 27648\approx 0.36$ ($27648$ is the size of the image/measurement). We chose the best reconstruction out of all the $4096$ steps, which (with the PSNR maximizing penalty parameter) was, in all cases, approximately the last iteration. This, as before, confims that the ACR does not require early stopping.\\
\indent The results of the experiments are reported in Table~\ref{table:deblurring_results_new}, while the representative deblurred images are shown in Figure \ref{table:deblurring_results} for a visual comparison. One can see that all learned reconstruction methods outperform TV. However, while the ACR and the AR2 outperform TV by a significant margin, e.g., approximately 1 dB in terms the PSNR values, AR only outperforms it by a somewhat smaller margin, and the SFB outperforms TV only by half a dB. The results lead us to two conclusions for this deblurring setting: (i) The convexity of the ACR does not seem to be a significant constraint in terms of performance and (ii) one seems to benefit from using more powerful and expressive convex models than the SFB prior.
\section{Conclusions}
We proposed a novel data-driven strongly-convex regularization approach for inverse problems and established analytical convergence guarantee and stability estimate. Moreover, we showed the existence of a sub-gradient descent algorithm for minimizing the variational loss, leading to a reconstruction error that decays monotonically to zero with iterations. The proposed ACR approach brings together the power of data-driven inference and the provability of analytical convex regularization. Numerical performance evaluation on sparse-view CT suggests that ACR is superior to the classical TV reconstruction, but gets outperformed by the non-convex AR method, albeit with significantly fewer parameters in the regularizer network as compared to AR. The image deblurring experiment indicates that the ACR model is at least on par and sometimes better than AR in terms of reconstruction quality, thus showing a clear advantage both in terms of theoretical guarantees and numerical performance. We also noted that ACR could be parametrized more parsimoniously as compared to its non-convex counterpart without significantly affecting its performance, while avoiding overfitting in a limited-data scenario and thereby leading to visibly fewer artifacts in the reconstructed image. Further, we noted that, unlike AR, ACR did not require early stopping during training to avoid overfitting; or during reconstruction to prevent the gradient-descent updates from diverging. 
%\section*{Appendix A: Cram\'er-Rao Bound}
% \section*{Acknowledgments}
% \noindent The authors would like to thank Sebastian Lunz for his valuable inputs on  network parametrization in the infinite-dimensional setting and reproducing the performance of AR.

\ifCLASSOPTIONcaptionsoff
  \newpage
\fi

\bibliographystyle{IEEEtran}
\bibliography{bib}

\begin{thebibliography}{10}
\providecommand{\url}[1]{#1}
\csname url@samestyle\endcsname
\providecommand{\newblock}{\relax}
\providecommand{\bibinfo}[2]{#2}
\providecommand{\BIBentrySTDinterwordspacing}{\spaceskip=0pt\relax}
\providecommand{\BIBentryALTinterwordstretchfactor}{4}
\providecommand{\BIBentryALTinterwordspacing}{\spaceskip=\fontdimen2\font plus
\BIBentryALTinterwordstretchfactor\fontdimen3\font minus
  \fontdimen4\font\relax}
\providecommand{\BIBforeignlanguage}[2]{{%
\expandafter\ifx\csname l@#1\endcsname\relax
\typeout{** WARNING: IEEEtran.bst: No hyphenation pattern has been}%
\typeout{** loaded for the language `#1'. Using the pattern for}%
\typeout{** the default language instead.}%
\else
\language=\csname l@#1\endcsname
\fi
#2}}
\providecommand{\BIBdecl}{\relax}
\BIBdecl

\bibitem{data_driven_inv_prob}
S.~Arridge, P.~Maass, O.~\"Oktem, and C.-B. Sch\"onlieb, ``Solving inverse
  problems using data-driven models,'' \emph{Acta Numerica}, vol.~28, pp.
  1--174, 2019.

\bibitem{automap}
B.~Zhu, J.~Z. Liu, S.~F. Cauley, B.~R. Rosen, and M.~S. Rosen, ``Image
  reconstruction by domain-transform manifold learning,'' \emph{Nature}, vol.
  555, pp. 487--492, 2018.

\bibitem{postprocessing_cnn}
K.~H. {Jin}, M.~T. {McCann}, E.~{Froustey}, and M.~{Unser}, ``Deep
  convolutional neural network for inverse problems in imaging,'' \emph{IEEE
  Transactions on Image Processing}, vol.~26, no.~9, pp. 4509--4522, 2017.

\bibitem{jonas_learned_iterative}
J.~Adler and O.~\"Oktem, ``Solving ill-posed inverse problems using iterative
  deep neural networks,'' \emph{Inverse Problems}, vol.~33, no.~12, 2009.

\bibitem{lpd_tmi}
J.~Adler and O.~{\"O}ktem, ``Learned primal-dual reconstruction,'' \emph{IEEE
  transactions on medical imaging}, vol.~37, no.~6, pp. 1322--1332, 2018.

\bibitem{kobler2017variational}
E.~Kobler, T.~Klatzer, K.~Hammernik, and T.~Pock, ``Variational networks:
  connecting variational methods and deep learning,'' in \emph{German
  conference on pattern recognition}.\hskip 1em plus 0.5em minus 0.4em\relax
  Springer, 2017, pp. 281--293.

\bibitem{meinhardt2017learning}
T.~Meinhardt, M.~Moller, C.~Hazirbas, and D.~Cremers, ``Learning proximal
  operators: Using denoising networks for regularizing inverse imaging
  problems,'' in \emph{Proceedings of the IEEE International Conference on
  Computer Vision}, 2017, pp. 1781--1790.

\bibitem{elad_ksvd1}
M.~Aharon, M.~Elad, and A.~Bruckstein, ``K-svd: An algorithm for designing
  overcomplete dictionaries for sparse representation,'' \emph{IEEE
  Transactions on Signal Processing}, vol.~54, no.~11, pp. 4311--4322, 2006.

\bibitem{elad_ksvd1_analysis}
R.~Rubinsttein, T.~Peleg, and M.~Elad, ``Analysis k-svd: A dictionary-learning
  algorithm for the analysis sparse model,'' \emph{IEEE Transactions on Signal
  Processing}, vol.~61, no.~3, pp. 661--677, 2013.

\bibitem{sparse_t}
S.~Ravishankar and Y.~Bresler, ``Learning sparsifying transforms,'' \emph{IEEE
  Transactions on Signal Processing}, vol.~61, no.~5, pp. 1072--1086, 2012.

\bibitem{mlcsc_elad}
J.~Sulam, V.~Papyan, Y.~Romano, and M.~Elad, ``Multilayer convolutional sparse
  modeling: pursuit and dictionary learning,'' \emph{IEEE Transactions on
  Signal Processing}, vol.~5, no.~15, pp. 4090--4104, 2018.

\bibitem{romano2017RED}
Y.~Romano, M.~Elad, and P.~Milanfar, ``The little engine that could:
  Regularization by denoising (red),'' \emph{SIAM Journal on Imaging Sciences},
  vol.~10, no.~4, pp. 1804--1844, 2017.

\bibitem{chan2016plug}
S.~H. Chan, X.~Wang, and O.~A. Elgendy, ``Plug-and-play admm for image
  restoration: Fixed-point convergence and applications,'' \emph{IEEE
  Transactions on Computational Imaging}, vol.~3, no.~1, pp. 84--98, 2016.

\bibitem{red_schniter}
E.~T. Reehorst and P.~Schniter, ``Regularization by denoising: clarifications
  and new interpretations,'' \emph{IEEE Transactions on Computational Imaging},
  vol.~5, no.~1, pp. 52--67, 2019.

\bibitem{rare_deep_prior}
J.~Liu, Y.~Sun, C.~Eldeniz, W.~Gan, A.~H., and K.~U. S., ``Rare: image
  reconstruction using deep priors learned without groundtruth,'' \emph{IEEE J.
  Selected Topics in Signal Processing}, vol.~14, no.~6, pp. 1088--1099, 2020.

\bibitem{online_pnp_tci}
Y.~Sun, B.~Wohlberg, and K.~U. S., ``An online plug-and-play algorithm for
  regularized image reconstruction,'' \emph{IEEE Transactions on Computational
  Imaging}, vol.~5, no.~3, pp. 395--408, 2019.

\bibitem{ar_nips}
S.~Lunz, O.~{\"O}ktem, and C.-B. Sch{\"o}nlieb, ``Adversarial regularizers in
  inverse problems,'' in \emph{Advances in Neural Information Processing
  Systems}, 2018, pp. 8507--8516.

\bibitem{nett_paper}
H.~Li, J.~Schwab, S.~Antholzer, and M.~Haltmeier, ``{NETT}: Solving inverse
  problems with deep neural networks,'' \emph{arXiv preprint
  arXiv:1803.00092v3}, Dec. 2019.

\bibitem{ulyanov2018deepImagePrior}
D.~Ulyanov, A.~Vedaldi, and V.~Lempitsky, ``Deep image prior,'' in
  \emph{Proceedings of the IEEE Conference on Computer Vision and Pattern
  Recognition}, 2018, pp. 9446--9454.

\bibitem{kobler2020total}
E.~Kobler, A.~Effland, K.~Kunisch, and T.~Pock, ``Total deep variation for
  linear inverse problems,'' in \emph{Proceedings of the IEEE Conference on
  Computer Vision and Pattern Recognition}, 2020, pp. 7549--7558.

\bibitem{peng2019auto}
P.~Peng, S.~Jalali, and X.~Yuan, ``Auto-encoders for compressed sensing,''
  2019.

\bibitem{dittmer2019regularization}
S.~Dittmer, T.~Kluth, P.~Maass, and D.~O. Baguer, ``Regularization by
  architecture: A deep prior approach for inverse problems,'' \emph{Journal of
  Mathematical Imaging and Vision}, pp. 1--15, 2019.

\bibitem{amos2017input}
B.~Amos, L.~Xu, and J.~Z. Kolter, ``Input convex neural networks,'' in
  \emph{International Conference on Machine Learning}, 2017, pp. 146--155.

\bibitem{wgan_main}
M.~Arjovsky, S.~Chintala, and L.~Bottou, ``Wasserstein gan,'' \emph{arXiv
  preprint arXiv:1701.07875v3}, Dec. 2017.

\bibitem{scherzer2009variational}
O.~Scherzer, M.~Grasmair, H.~Grossauer, M.~Haltmeier, and F.~Lenzen,
  \emph{Variational methods in imaging}.\hskip 1em plus 0.5em minus 0.4em\relax
  Springer, 2009.

\bibitem{boyd2004convex}
S.~Boyd and L.~Vandenberghe, \emph{Convex optimization}.\hskip 1em plus 0.5em
  minus 0.4em\relax Cambridge university press, 2004.

\bibitem{wgan_gp}
I.~Gulrajani1, F.~Ahmed, M.~Arjovsky, V.~Dumoulin, and A.~Courville, ``Improved
  training of wasserstein gans,'' \emph{arXiv preprint arXiv:1704.00028v3},
  Dec. 2017.

\bibitem{parikh2014proximal}
N.~Parikh and S.~Boyd, ``Proximal algorithms,'' \emph{Foundations and Trends in
  optimization}, vol.~1, no.~3, pp. 127--239, 2014.

\bibitem{nesterov_tutorial_cvx}
Y.~Nesterov, ``Primal-dual subgradient methods for convex problems,''
  \emph{Mathematical programming}, vol. 120, no.~1, pp. 221--259, 2009.

\bibitem{subgrad_convergence}
\BIBentryALTinterwordspacing
S.~Boyd, \emph{Subgradient methods}, May 2014 (accessed July 27, 2020).
  [Online]. Available:
  \url{https://stanford.edu/class/ee364b/lectures/subgrad_method_notes.pdf}
\BIBentrySTDinterwordspacing

\bibitem{mayo_ct_challenge}
C.~McCollough, ``Tfg-207a-04: Overview of the low dose ct grand challenge,''
  \emph{Medical Physics}, vol.~43, no.~6, pp. 3759--3760, 2014.

\bibitem{odl}
J.~Adler, H.~Kohr, and O.~\"Oktem, ``Operator discretization library (odl),''
  \emph{Software available from https://github.com/odlgroup/odl}, 2017.

\bibitem{coates2011analysis}
A.~Coates, A.~Ng, and H.~Lee, ``An analysis of single-layer networks in
  unsupervised feature learning,'' in \emph{Proceedings of the fourteenth
  international conference on artificial intelligence and statistics}, 2011,
  pp. 215--223.

\bibitem{ssim_paper_2004}
Z.~Wang, A.~C. Bovik, H.~R. Sheikh, and E.~P. Simoncelli, ``Image quality
  assessment: From error visibility to structural similarity,'' \emph{IEEE
  Transactions on Image Processing}, vol.~13, no.~4, pp. 600--612, 2004.

\bibitem{sfb_bresler}
L.~Pfister and Y.~Bresler, ``Learning filter-bank sparsifying transforms,''
  \emph{arXiv preprint arXiv:1803.01980v1}, 2018.

\bibitem{conv_transform_learning_angshul}
J.~Maggu, E.~Chouzenoux, G.~Chierchia, and A.~Majumdar, ``Convolutional
  transform learning,'' in \emph{International Conference on Neural Information
  Processing}, 2018, pp. 391--398.

\bibitem{paszke2017automatic}
A.~Paszke, S.~Gross, S.~Chintala, G.~Chanan, E.~Yang, Z.~DeVito, Z.~Lin,
  A.~Desmaison, L.~Antiga, and A.~Lerer, ``Automatic differentiation in
  pytorch,'' 2017.

\bibitem{kingma2014adam}
D.~P. Kingma and J.~Ba, ``Adam: A method for stochastic optimization,''
  \emph{arXiv preprint arXiv:1412.6980}, 2014.

\bibitem{icnn_opt_transport}
Y.~Chen, Y.~Shi, and B.~Zhang, ``Optimal control via neural networks: A convex
  approach,'' \emph{arXiv preprint arXiv:1805.11835v5}, 2019.

\end{thebibliography}

\end{document}